%% file: llm_opt_con_survey.tex
\documentclass[journal]{IEEEtai}

\usepackage[colorlinks,urlcolor=blue,linkcolor=blue,citecolor=blue]{hyperref}

\usepackage{color,array}

\usepackage{graphicx}

\usepackage{amsmath}
\usepackage{booktabs}
\usepackage{algorithm}
\usepackage{algpseudocode}  
\usepackage{graphicx}
\usepackage{hyperref}
\usepackage{color,array}
\usepackage{pifont}  
\newcommand{\cmark}{\ding{51}}  
\newcommand{\xmark}{\ding{55}}  
\usepackage[T1]{fontenc}
\usepackage{lmodern}
\usepackage{listings}
\usepackage{xcolor}

\usepackage{amsmath}
\usepackage{booktabs}
\usepackage{algorithm}
\usepackage{algpseudocode}  
\usepackage{graphicx}
\usepackage{hyperref}
\usepackage{color,array}
\usepackage{pifont}  
\usepackage[T1]{fontenc}
\usepackage{lmodern}
\usepackage{listings}
\usepackage{xcolor} 
\usepackage{comment}

\usepackage{cite}
\usepackage{amsmath,amssymb,amsfonts}

\usepackage{threeparttable}

\usepackage{tikz}
\definecolor{hidden-red}{RGB}{205, 44, 36}
\definecolor{hidden-blue}{RGB}{194,232,247}
\definecolor{hidden-orange}{RGB}{243,202,120}
\definecolor{hidden-green}{RGB}{34,139,34}
\definecolor{hidden-pink}{RGB}{255,245,247}
\definecolor{hidden-black}{RGB}{20,68,106}

\newcommand{\highlight}[1]{\textcolor{black}{#1}}
\newcommand\encircle[2][]{\tikz[overlay]\node[fill=blue!20,inner sep=2pt, anchor=text, rectangle, rounded corners=1.5mm,#1] {#2};\phantom{#2}}

\definecolor{hidden-red}{RGB}{205, 44, 36}
\definecolor{hidden-blue}{RGB}{194,232,247}
\definecolor{hidden-orange}{RGB}{243,202,120}
\definecolor{hidden-green}{RGB}{34,139,34}
\definecolor{hidden-pink}{RGB}{255,245,247}
\definecolor{hidden-black}{RGB}{20,68,106}

\definecolor{myGreen}{RGB}{127,210,85}
\definecolor{myOrange}{RGB}{242,154,66}
\definecolor{myYellow}{RGB}{247,223,65}
\definecolor{myRed}{RGB}{232,80,43}
\definecolor{myViolet}{RGB}{162,57,102}
\definecolor{myBlue}{HTML}{4686f3}
\definecolor{myYellowv2}{HTML}{E6C802}
\definecolor{myOrangev2}{HTML}{ED8E55}
\definecolor{MyGreenv2}{HTML}{009B55}
\definecolor{MyRedv2}{HTML}{c22f2f}
\definecolor{DarkRed}{RGB}{130,25,0}
\definecolor{PurpleRed}{RGB}{204,0,102}
\definecolor{DarkGreen}{RGB}{30,130,30}
\definecolor{DarkBlue}{RGB}{0,0,250}
\definecolor{DarkYellow}{RGB}{255,128,0}

\usepackage{listings}
\usepackage{xcolor} 

\begin{document}

\title{A Survey on Mathematical Reasoning and Optimization with Large Language Models}

\author{Ali Forootani$^{*}$, \IEEEmembership{Senior Member, IEEE}
\thanks{Ali Forootani is with Helmholtz Centre for Environmental Research - UFZ, Permoserstraße 15, 04318 Leipzig, Germany, \texttt{email: ali.forootani@ufz.de, aliforootani@ieee.org}.}
}



\markboth{Journal of IEEE Transactions on Artificial Intelligence, Vol. 00, No. 0, Month 2020}
{First A. Author \MakeLowercase{\textit{et al.}}: Bare Demo of IEEEtai.cls for IEEE Journals of IEEE Transactions on Artificial Intelligence}

\maketitle

\begin{abstract}
Mathematical reasoning and optimization are fundamental to artificial intelligence and computational problem-solving. Recent advancements in large language models have significantly enhanced AI-driven mathematical reasoning, theorem proving, and optimization techniques. This survey explores the evolution of mathematical problem-solving in AI, from early statistical learning approaches to the latest deep learning-based and transformer-driven methodologies. We review the capabilities of pre-trained language models  and Large Language Models in handling arithmetic operations, complex mathematical reasoning, theorem proving, and structured symbolic computation.  

A key focus is on how Large Language Models integrate with optimization and control frameworks, including mixed-integer programming, linear quadratic control, and multi-agent optimization strategies. We discuss how Large Language Models assist in problem formulation, constraint generation, and heuristic search, thereby bridging the gap between theoretical mathematical reasoning and practical optimization applications. Additionally, we analyze various enhancement techniques, including Chain-of-Thought  reasoning, instruction tuning, and tool-augmented approaches that improve Large Language Model's problem-solving capabilities.  

Despite these advancements, Large Language Models face challenges in numerical precision, logical consistency, and verification of mathematical proofs. We highlight emerging trends in hybrid neural-symbolic reasoning, structured prompt engineering, and multi-step self-correction mechanisms aimed at improving Large Language Model performance in mathematical and optimization tasks. Future research should focus on refining interpretability, integrating domain-specific solvers, and ensuring the robustness of AI-driven decision-making frameworks. This survey provides a comprehensive review of the current landscape and future directions of mathematical reasoning and optimization with Large Language Models, offering insights into their potential applications across engineering, finance, and scientific research.  
\end{abstract}




\section{Introduction}\label{introduction}

Mathematics serves as a cornerstone of human cognition, underpinning a vast array of disciplines, including the natural sciences, engineering, medicine, finance, computer science, and the social sciences. In the field of natural language processing (NLP), the development of computational models capable of autonomously solving mathematical word problems has been an area of significant research interest since the early 1960s \cite{feigenbaum1963computers,bobrow1964natural,briars1984integrated,fletcher1985understanding}. This pursuit is not merely about solving arithmetic and algebraic expressions but also about advancing general reasoning mechanisms, which are fundamental to achieving artificial general intelligence (AGI) \cite{wang2017deep}. By tackling the complexities of mathematical reasoning, researchers aim to push the boundaries of AI systems' logical deduction capabilities and their ability to understand and manipulate symbolic knowledge.

Early approaches to solving mathematical word problems primarily relied on statistical learning techniques \cite{hosseini-etal-2014-learning,kushman-etal-2014-learning,zhou2015learn,mitra2016learning}. Over time, more sophisticated methodologies emerged, incorporating semantic parsing and structured representations \cite{kushman-etal-2014-learning,roy2015reasoning,roy2015solving,shi2015automatically,koncel2015parsing}. These innovations laid the foundation for modern deep learning-based mathematical problem-solving, where neural networks have become a powerful tool for automating complex mathematical reasoning tasks \cite{wang2017deep,couperus2023large}. 

In recent years, the field has witnessed unprecedented advancements, largely driven by the rise of powerful pre-trained language models (PLMs) and large language models (LLMs) \cite{wei2022chain,matzakos2023learning}. These models, including BERT \cite{devlinBERTPretrainingDeep2019}, RoBERTa \cite{liuRoBERTaRobustlyOptimized2019}, BART \cite{lewis2020bart}, GPT-1 \cite{radford_gpt-1_2018}, and GPT-2 \cite{radford_gpt-2_2019}, have demonstrated remarkable capabilities in learning from large-scale textual corpora, thereby acquiring general linguistic and numerical reasoning skills. 

To enhance the mathematical reasoning abilities of PLMs, researchers have explored various techniques, including pre-training and fine-tuning on specialized mathematical datasets \cite{clarkRegentsScienceExams2021,gevaInjectingNumericalReasoning2020,fengInjectingNumericalReasoning2022}. Several models, such as GenBERT\cite{gevaInjectingNumericalReasoning2020}, NF-NSM\cite{fengInjectingNumericalReasoning2022}, MathBERT\cite{pengMathBERTPretrainedModel2021}, and LISA \cite{jiang2021lisa}, integrate numerical data and mathematical expressions into their training objectives, enhancing their ability to perform symbolic and numerical reasoning. Moreover, architectural innovations such as specialized loss functions and task-specific modules have been proposed to further refine PLM-based mathematical comprehension \cite{zhu2021tat,zhao2022multihiertt,jieLearningReasonDeductively2022,liSeekingPatternsNot2022}. 

The emergence of large-scale LLMs, exemplified by GPT-4 \cite{openai2023gpt4}, has significantly transformed mathematical problem-solving. These models have demonstrated remarkable generalization across various NLP tasks, benefiting from key techniques such as in-context learning \cite{min2022metaicl,brown2020language,chen2022meta} and instruction learning \cite{yang2023gpt,liu_goat_2023}. Notably, research by Wang et al. \cite{wang2022self} suggests that LLMs with over 100 billion parameters—such as GPT-3 (175 billion) \cite{brown2020language} and PaLM (540 billion) \cite{chowdhery2022palm}—can effectively solve complex mathematical problems through chain-of-thought (CoT) reasoning when provided with well-structured demonstrations \cite{wei2022chain}. 

Advancements in CoT methodologies have further improved the reasoning capabilities of LLMs, leading to the development of structured reasoning techniques, programmatic reasoning frameworks, and dynamic CoT models \cite{chen2022program,lu2022dynamic,zhang2023automatic,fu2022complexity}. Additionally, tool-augmented models that integrate external symbolic manipulation systems have been proposed to enhance mathematical problem-solving performance \cite{gao2023pal,drori2022neural,he2023solving}. Such approaches highlight the potential of combining neural and symbolic reasoning to achieve more robust AI systems capable of higher-order mathematical cognition. 

Several surveys have reviewed different aspects of mathematical reasoning in LLMs. Notably:  

\begin{itemize}  
    \item \cite{lu-etal-2023-survey} provides a comprehensive survey of deep learning methodologies for mathematical reasoning (DL4Math).  
    \item \cite{qiao-etal-2023-reasoning} explores general reasoning mechanisms in language models, covering arithmetic, commonsense, logical, symbolic, and multi-modal reasoning (LM4Reasoning).
    \item \cite{chuCoTReasoningSurvey2023} presents an analytical review of chain-of-thought (CoT) reasoning methodologies.
    \item \cite{qiu2020pre} and \cite{zhao2023survey} offer surveys on pre-trained language models (PLMs) and LLMs, respectively.  
    \item \cite{yan2024survey} reviews multi-modal LLMs for mathematical reasoning (MMLM4Math).  
\end{itemize}  

Overall, the fusion of deep learning techniques, structured reasoning approaches, and tool-augmented models continues to drive rapid progress in AI-driven mathematical problem-solving. While existing methods have demonstrated impressive performance, fundamental challenges remain, particularly in areas such as formal proof generation, theorem proving, and real-world mathematical applications. As research continues to push the boundaries of AI's mathematical reasoning capabilities, future advancements will likely be characterized by improved interpretability, robustness, and the seamless integration of symbolic and neural reasoning paradigms.

\subsection{Bridging Mathematical Reasoning and Optimization with LLMs}

Mathematical reasoning forms the foundation of logical inference, problem-solving, and numerical computation in artificial intelligence \cite{liu2023mathematical}. Large language models (LLMs) have demonstrated remarkable capabilities in handling mathematical reasoning tasks, including theorem proving, algebraic manipulation, and structured numerical problem-solving. Through advancements such as Chain-of-Thought (CoT) prompting, programmatic reasoning, and tool-augmented inference, LLMs have shown substantial progress in solving complex mathematical problems \cite{imani2023mathprompter}. However, the transition from theoretical reasoning to practical applications, such as optimization and control, requires an additional layer of adaptability where models can not only deduce but also generate and refine solutions iteratively.

Optimization and control problems often demand a structured approach where solutions must satisfy strict mathematical constraints while achieving predefined objectives. In scenarios such as mixed-integer programming (MIP) and linear programming (LP), traditional solvers rely on well-established algorithms, including branch-and-bound methods and gradient-based optimization. While LLMs may not replace these solvers outright, they can significantly enhance the modeling process by formulating problem statements, generating initial constraints, and suggesting heuristic improvements \cite{wang2024leveraging}. By leveraging their extensive pretraining in symbolic mathematics and structured problem-solving, LLMs can aid domain experts in defining well-posed optimization problems and verifying computational solutions.

A crucial advantage of LLMs in this domain lies in their ability to process natural language descriptions of complex optimization scenarios and convert them into structured mathematical formulations \cite{ahmed2403lm4opt}. For instance, in the context of battery energy storage system (BESS) optimization, an LLM can help translate high-level policy constraints—such as cost minimization, peak shaving, or energy arbitrage—into a solvable mathematical model. Similarly, in control applications like linear quadratic regulation (LQR), LLMs can assist in deriving optimal control laws by structuring the necessary state-space representations and computing feedback gains iteratively.

By integrating mathematical reasoning with optimization and control strategies, LLMs present an opportunity to bridge the gap between abstract theoretical models and real-world engineering applications \cite{Xuhan2025}. Their ability to automate certain aspects of model formulation, constraint generation, and heuristic search makes them valuable tools for decision-making in complex systems. Future research should focus on refining their numerical precision, enhancing symbolic computation capabilities, and improving their ability to interface seamlessly with existing optimization solvers. This synergy between mathematical reasoning and optimization will pave the way for more intelligent, interpretable, and efficient decision-support systems in engineering and scientific domains.

This paper is organized as follows: In Section~\ref{Tasks}, we introduce the types of mathematical tasks that motivate the use of language models. Section~\ref{PLMs} presents pre-trained language model (PLM)-based approaches for mathematical reasoning, with a particular focus on Non-Autoregressive Language Models discussed in Section~\ref{Non-Autoregression LMs}. Section~\ref{LLMs_reasoning} explores large language model (LLM)-based methods tailored for mathematical reasoning. In Section~\ref{llm_comparisons}, we compare various mathematical LLMs, followed by a performance analysis in Section~\ref{llm_performance_analysis}. Section~\ref{challenges_methods} investigates the application of LLMs to time series analysis, outlining the associated challenges, methodologies, and future directions. A technical comparison of commonly used LLMs is provided in Section~\ref{comparisons_famous_llms}. Section~\ref{applications} discusses the role of LLMs in optimization and control. Section~\ref{Challenges and Further Directions} highlights open challenges and outlines potential research directions. Finally, conclusions are drawn in Section~\ref{conclusion}.

\section{Mathematical Tasks}
\label{Tasks}

In this section, we categorize mathematical tasks into two primary domains: \textit{mathematical calculation} and \textit{mathematical reasoning}.

\subsection{Mathematical Calculation}
\label{Mathematical Calculation}

The emergence of LMs has significantly influenced computational research, particularly in performing arithmetic operations. Early LMs exhibited only rudimentary arithmetic capabilities, encoding numbers as mere textual representations. However, as these models evolved, they demonstrated the ability to develop arithmetic skills through fine-tuning on specialized datasets, even without explicitly designed numerical representations \cite{lai2024executing}.

\subsubsection{Arithmetic Representation}
\label{Arithmetic Representation}

Initially, numerical values were often omitted, oversimplified, or treated as ordinary text tokens, leading to suboptimal performance on mathematical tasks. For instance, BERT performs five times worse when the expected answer is a numerical value rather than a textual span in the DROP benchmark \cite{dua2019drop}. This limitation highlighted the necessity of refining how LMs process and represent numbers.

Recent research has proposed several techniques to enhance numerical representation within LMs:

\begin{itemize}
    \item GenBERT \cite{gevaInjectingNumericalReasoning2020} tokenizes numbers at the digit level and fine-tunes on arithmetic word problems.
    \item In \cite{zhang2020language} a technique is introduced that converts numbers into scientific notation (e.g., 314.1 as 3141[EXP]2) to facilitate better numerical understanding.
    \item Studies such as \cite{spithourakis2018numeracy} and \cite{wallace2019nlp} integrate digit embeddings into a single embedding that represents entire numbers more effectively.
    \item In \cite{berg-kirkpatrickEmpiricalInvestigationContextualized2020} a method is proposed to leverage digit-RNN and exponent embeddings, emphasizing the exponent while disregarding the mantissa for improved numerical processing.
    \item GOAT \cite{liu_goat_2023} introduced Consistent Tokenization to enhance the numerical representation consistency within LMs.
\end{itemize}

These techniques have collectively improved the way LMs interpret, store, and manipulate numerical values, significantly boosting their effectiveness in arithmetic reasoning.

\subsubsection{Arithmetic Calculation}
\label{Arithmetic Calculation}

The ability of LMs to perform arithmetic computations has been a widely studied topic. Early research primarily focused on fundamental operations such as addition and subtraction \cite{nogueira2021investigating, wang2021exploring}, whereas more recent studies have expanded into complex operations like multi-digit multiplication \cite{muffo_evaluating_nodate}.

Key findings in arithmetic calculation include:

\begin{itemize}
    \item In \cite{yuan2023well} arithmetic capabilities across various LMs, including GPT-4 \cite{openai2023gpt4}, Galactica \cite{taylor_galactica_2022}, and LLaMA \cite{touvron2023llama} hase been evaluated.
    \item In \cite{zhouTeachingAlgorithmicReasoning2022} has been demonstrated that specialized prompt engineering can significantly improve addition accuracy, though multiplication remains challenging for numbers exceeding seven digits.
    \item In \cite{jelassiLengthGeneralizationArithmetic2023} length generalization has been investigated in arithmetic using relative position embeddings and training set priming.
    \item ScratchpadGPT \cite{nye_show_2021} leveraged Chain-of-Thought (CoT) prompting to enhance addition accuracy.
    \item In GOAT \cite{liu_goat_2023} supervised instruction tuning has been applied to improve performance on large-integer arithmetic tasks.
    \item MathGLM \cite{yang2023gpt} decomposed complex arithmetic expressions into simpler subproblems, enhancing model precision.
\end{itemize}

These advancements illustrate the growing sophistication of LMs in arithmetic computations and their potential applications in computational mathematics.

\subsection{Mathematical Reasoning}
\label{Mathematical Reasoning}

Mathematical reasoning is a critical capability in artificial intelligence, enabling models to understand, explain, and solve complex mathematical problems. Research in this domain has focused on two primary areas: math problem-solving and theorem proving.

\subsubsection{Math Problem Solving}
\label{Math Problem Solving}

Math problem-solving requires LMs to apply logical reasoning and computational models to solve mathematical problems at various levels of complexity. Generally, mathematical problems fall into two categories:

\begin{itemize}
    \item Contextualized problems: These involve understanding a given scenario, translating it into a mathematical problem, and generating a step-by-step solution.
    \item Direct mathematical problems: These require direct computation of a solution without the need for additional contextual interpretation.
\end{itemize}

Several notable advancements in this domain include:

\begin{itemize}
    \item MathPrompter \cite{imani2023mathprompter} utilizes GPT-3 DaVinci for solving mathematical word problems (MWPs).
    \item In \cite{yuan2023scaling} Rejection Sampling Fine-Tuning (RFT) has been proposed to enhance mathematical reasoning capabilities in LMs.
    \item MetaMath \cite{yu2023metamath} generates synthetic mathematical problems to facilitate self-improvement.
    \item WizardMath \cite{luo2023wizardmath} applies evolutionary instruction tuning to refine LM's mathematical reasoning skills.
    \item MathAttack \cite{zhou2023mathattack} evaluates the robustness of LMs against adversarial examples in mathematical problem-solving.
    \item LLEMMA \cite{azerbayev8llemma} is an open-source LM designed specifically for mathematical reasoning.
\end{itemize}

These models illustrate the ongoing effort to refine LM's ability to reason mathematically and solve increasingly complex problems.

\subsubsection{Theorem Proving}
\label{Theorem Proving}

Theorem proving (TP) is a challenging AI task that involves verifying mathematical statements using formal logic and structured reasoning. Recent research has investigated the ability of LMs to engage in theorem proving by leveraging deep learning techniques and formal proof verification systems.

Key advancements in this area include:

\begin{itemize}
    \item GPT-f \cite{polu2020generative} generates proof strategies using deep learning-based methods.
    \item In \cite{jiang2022draft} incorporating informal proofs into TP models to enhance theorem reasoning has been proposed.
    \item NaturalProofs \cite{welleck2021naturalproofs} compiled a comprehensive corpus for evaluating theorem-proving models.
    \item DeepMath \cite{irving2016deepmath} and INT \cite{wu2020int} assess the generalization capabilities of LMs in theorem proving.
    \item CoqGym \cite{yang2019learning} provides a dataset of human-written proofs to aid LM training in formal theorem proving.
    \item ASTactic \cite{yang2019learning} autonomously generates proof strategies using deep learning.
    \item LLEMMA \cite{azerbayev8llemma} explores auto-formalization and tactic prediction for mathematical proof generation.
\end{itemize}

Despite promising advancements, a significant challenge remains: hallucination in neural text generation \cite{maynez-etal-2020-faithfulness, ji2023survey}. Errors in logical reasoning can accumulate across proof steps, leading to incorrect conclusions. This highlights the urgent need for robust verification mechanisms to ensure mathematical rigor in AI-generated proofs.


Mathematical reasoning and arithmetic problem-solving represent crucial steps in improving LMs' structured reasoning capabilities. While current research continues to push the boundaries of AI-driven mathematical understanding, future advancements will likely focus on interpretability, error reduction, and the seamless integration of symbolic and neural reasoning approaches.


\section{Pre-trained Language Models (PLMs)-Based Approaches for Mathematical Reasoning} \label{PLMs}  

PLMs, built on the Transformer architecture \cite{vaswani2017attention}, have profoundly influenced Natural Language Processing (NLP) through the pre-training and fine-tuning paradigm. PLMs have been extensively utilized in text generation \cite{noorbakhsh2021pretrained} and can be broadly categorized into two major types: Autoregressive LMs (ALMs), such as GPT-1 \cite{radford_gpt-1_2018} and T5 \cite{rael_exploring_nodate}, and Non-Autoregressive LMs (NALMs), including BERT \cite{kenton2019bert} and RoBERTa\cite{liuRoBERTaRobustlyOptimized2019}. These architectures have laid the groundwork for advanced mathematical reasoning models, offering new capabilities in symbolic manipulation, theorem proving, and arithmetic problem-solving.


\subsection{Autoregressive Language Models} \label{Autoregression LMs}

Autoregressive Language Models (ALMs) generate text in a sequential manner, where each token is predicted based on the preceding tokens. These models are primarily categorized into two types: causal decoders, such as GPT-1 \cite{radford_gpt-1_2018} and GPT-2 \cite{radford_gpt-2_2019}, and encoder-decoder architectures, exemplified by T5 \cite{rael_exploring_nodate}. These architectures have significantly contributed to advancements in mathematical reasoning within Pretrained Language Models (PLMs).

Several research efforts have focused on adapting autoregressive models to improve mathematical reasoning capabilities:

\begin{itemize}
    \item GPT-f \cite{polu2020generative} explores automated theorem proving, demonstrating that large-scale language models can generate formal proofs, thereby showcasing their potential for logical deduction.
    \item The Expression-Pointer Transformer (EPT) \cite{kim_point_2020} addresses expression fragmentation issues, significantly enhancing equation-solving abilities in symbolic mathematical tasks.
    \item Generate \& Rank \cite{shen_generate_2021} introduces a ranking-based post-processing mechanism that refines the quality of generated solutions for mathematical word problems (MWPs).
    \item THOR \cite{jiang2022thor} integrates theorem proving within language models, leveraging logical reasoning to improve deductive capabilities.
    \item HyperTree Proof Search (HTPS) \cite{lample_hypertree_nodate} optimizes proof search strategies by incorporating prior results, surpassing GPT-f in theorem-proving efficiency and accuracy.
    \item Galactica \cite{taylor_galactica_2022} enhances performance on mathematical reasoning benchmarks, outperforming conventional PLMs on datasets such as MMLU \cite{hendryckstest2021} and MATH \cite{hendrycks2021measuring} through improved pretraining and fine-tuning techniques.
\end{itemize}

The performance of autoregressive PLMs in mathematical reasoning has been further refined through the development of specialized datasets and fine-tuning methodologies:

\begin{itemize}
    \item MATH-PLM \cite{hendrycks2021measuring} underscores the challenge of scaling PLMs for mathematical reasoning and provides benchmark datasets that serve as rigorous evaluation tools.
    \item LISA \cite{jiang2021lisa} constructs theorem-proving models by leveraging formal proof structures, thereby strengthening structured logical reasoning within PLMs.
    \item PACT \cite{han2021proof} introduces self-supervised learning techniques to enhance proof-term generation, improving the interpretability and accuracy of mathematical proofs.
    \item Minerva \cite{lewkowycz2022solving} fine-tunes PLMs on high-quality technical mathematical content, achieving state-of-the-art (SOTA) performance on various mathematical reasoning tasks.
    \item LIME \cite{wu_lime_nodate} proposes a novel pretraining strategy that induces mathematical reasoning biases, leading to superior performance compared to conventional transformer-based models.
\end{itemize}

The success of autoregressive LMs in solving mathematical problems can largely be attributed to their capacity to model complex sequential dependencies. This makes them particularly well-suited for theorem proving, symbolic manipulation, and structured logical inference tasks.


\subsection{Non-Autoregressive Language Models}
\label{Non-Autoregression LMs}  

Non-Autoregressive LMs (NALMs) differ from their autoregressive counterparts by generating multiple tokens simultaneously rather than sequentially. This parallelism significantly reduces dependency on previously generated tokens, leading to improved efficiency. Models such as BERT \cite{kenton2019bert} and RoBERTa \cite{liuRoBERTaRobustlyOptimized2019} employ masked token prediction, allowing them to learn contextual representations more effectively, making them well-suited for NLP tasks that require robust comprehension rather than generative capabilities.

In mathematical reasoning, fine-tuned variants of BERT have demonstrated remarkable performance:

\begin{itemize}
    \item Aristo \cite{clarkRegentsScienceExams2021} adapts BERT for solving scientific exam questions, demonstrating its ability to process structured reasoning problems.
    \item GenBERT \cite{gevaInjectingNumericalReasoning2020} and NF-NSM \cite{fengInjectingNumericalReasoning2022} enhance numerical reasoning by incorporating numerical data into pre-training objectives.
    \item MWP-BERT \cite{liangMWPBERTNumeracyaugmentedPretraining2022} improves numerical representation in mathematical word problems through numeracy-augmented pretraining.
    \item MathBERT \cite{pengMathBERTPretrainedModel2021} jointly trains on both textual and mathematical formula representations, improving mathematical comprehension.
    \item TAGOP \cite{zhu2021tat}, MT2Net \cite{zhao2022multihiertt}, and DeductReasoner \cite{jieLearningReasonDeductively2022} leverage PLMs for advanced mathematical operations and logical reasoning.
    \item BERT-TD \cite{liSeekingPatternsNot2022} applies contrastive learning to improve mathematical word problem pattern recognition.
\end{itemize}

Although NALMs are not inherently designed for generative tasks, they excel in mathematical comprehension due to their bidirectional encoding capabilities. By leveraging masked language modeling and contextual token embeddings, these models can capture intricate mathematical structures and improve equation parsing accuracy.

\subsection{Discussion and Future Directions}

The adoption of PLM-based approaches in mathematical reasoning has yielded significant progress, particularly in theorem proving, arithmetic calculations, and mathematical problem-solving. However, several challenges remain:

\begin{itemize}
    \item Numerical Precision: PLMs often struggle with exact arithmetic computations due to floating-point representation errors.
    \item Long-Range Dependencies: Complex mathematical proofs require deep reasoning chains, which current models struggle to retain effectively.
    \item Hallucination in Proof Generation: The tendency of LMs to generate incorrect but plausible-looking proofs remains a significant limitation.
    \item Hybrid Neural-Symbolic Methods: Integrating PLMs with symbolic solvers could enhance the rigor and correctness of mathematical reasoning.
\end{itemize}

Future research directions should focus on improving interpretability, reducing hallucinations, and integrating structured reasoning modules to ensure mathematical rigor. As PLMs continue to evolve, their role in automating theorem proving and mathematical modeling is expected to expand, pushing the boundaries of artificial intelligence in formal reasoning.

\section{LLM-Based Approaches for Mathematical Reasoning}
\label{LLMs_reasoning}

LLMs have revolutionized text processing and generation, extending their capabilities to mathematical reasoning \cite{touvron2023llama,openai2023gpt4}. Since mathematics follows structured linguistic and symbolic rules, LLMs capable of understanding these rules serve as powerful tools for problem-solving in various mathematical domains \cite{azerbayev8llemma,luo2023wizardmath}. Existing research in LLM-based mathematical reasoning can be broadly categorized into four key areas: instruction learning, tool-based methods, fundamental Chain-of-Thought (CoT) techniques, and advanced CoT methodologies.

\subsection{Instruction Learning}
\label{Instruction Learning}

Instruction learning improves mathematical reasoning in LLMs through structured task formulation, supervised fine-tuning, and in-context learning.

Instruction Building involves generating structured explanations and guidance to enhance model capabilities. Auto-explanation \cite{nakamoto2023enhancing} enables LLMs to construct self-explanatory mathematical datasets, improving problem comprehension. RefAug \cite{zhang2024learn} leverages GPT-4o to augment mathematical reasoning data, enriching training samples. Evol-Instruct \cite{xu_wizardlm_2023} introduces an iterative refinement mechanism where LLMs generate and refine their own instructional prompts, further enhanced by WizardMath's \cite{luo2023wizardmath} RLEIF optimization method. EURUS \cite{yuan2024advancing} integrates CoT \cite{wei2022chain} with external tools, optimizing reasoning paths using Direct Preference Optimization (DPO) \cite{rafailov2024direct} and Noise Contrastive Alignment (NCA) \cite{chen2024noise}.

Instruction Tuning fine-tunes LLMs to align outputs with human intent \cite{zhang_instruction_2023}. MathGLM \cite{yang2023gpt} and GOAT \cite{liu_goat_2023} improve arithmetic operations through structured tokenization, while Calculon \cite{muffo_evaluating_nodate} employs digit decomposition for numerical precision. PaLM 2-L-Math \cite{liu_improving_2023} demonstrates that combining solution re-ranking with majority voting enhances performance in mathematical reasoning. LLEMMA \cite{azerbayev8llemma} pre-trains on mathematical corpora and integrates a Python interpreter to execute symbolic computations. More recent LLMs such as Qwen2.5-MATH \cite{yang2024qwen2} and InternLM-Math \cite{ying2024internlmmath} employ iterative fine-tuning and reward modeling to refine mathematical understanding and improve accuracy.

In-Context Learning (ICL) enhances mathematical reasoning by presenting task examples during inference without modifying model parameters \cite{brown2020language}. ScratchpadGPT \cite{nye_show_2021} improves multi-step calculations by generating intermediate reasoning steps. Codex-math\cite{drori2022neural} fine-tunes LLMs on code-based mathematical tasks for automated problem-solving. LAMBADA \cite{kazemi2022lambada} introduces backward reasoning, guiding LLMs in reverse problem-solving, while LogicSolver \cite{yang2022logicsolver} embeds algebraic logic into in-context prompts to enhance interpretability.

\subsection{Tool-Based Methods}
\label{Tool-based Methods}

LLMs can augment their mathematical reasoning abilities by integrating symbolic solvers and programming tools \cite{schick2023toolformer,parisi2022talm}. 

Single-tool methods focus on integrating LMs with a specific symbolic reasoning framework. SymbLLM \cite{he2023solving} enhances mathematical word problem (MWP) solving by leveraging symbolic solvers, while LeanReasoner \cite{jiang2024leanreasoner} incorporates Lean theorem proving to ensure logical consistency. PAL \cite{gao2023pal} improves problem-solving by generating Python programs as intermediate reasoning steps.

Multi-tool approaches extend LLM's capabilities by enabling interactions with multiple external tools. Toolformer \cite{schick2023toolformer} equips LLMs with APIs for calculators, search engines, and symbolic computation systems. ART \cite{paranjape2023art} expands on this by integrating automated retrieval-based tool selection, dynamically choosing the most appropriate tool for each problem.

\subsection{Fundamental Chain-of-Thought (CoT) Methods}
\label{Fundamental CoT Methods}

Chain-of-Thought (CoT) reasoning enhances LLM's performance in mathematical tasks by decomposing problems into step-by-step logical sequences \cite{wei2022chain}. 

Auto-CoT \cite{zhang2023automatic} automatically clusters representative problems to generate effective reasoning examples, reducing reliance on manual prompt engineering. Rejection Sampling Fine-Tuning (RFT) \cite{yuan2023scaling} refines solution paths by discarding incorrect reasoning chains through rejection sampling. PromptPG-CoT \cite{lu2022dynamic} optimizes CoT prompts using reinforcement learning (RL), improving generalization across diverse mathematical tasks.

\subsection{Advanced Chain-of-Thought (CoT) Methods}
\label{Advanced CoT Methods}

Verify-based methods enhance mathematical reasoning by incorporating self-verification mechanisms. VerifyCoT \cite{ling2023deductive} ensures deductive consistency, while SSC-CoT \cite{zhao2024stepwise} validates intermediate steps to prevent cascading reasoning errors.

Ensemble-based approaches improve accuracy by aggregating multiple reasoning trajectories. Self-Consistency \cite{wang2022self} selects the most frequent solution among multiple sampled responses, while MCR \cite{yoran2023answering} uses majority voting across distinct reasoning paths.

\textbf{Planning-based methods} organize reasoning hierarchically for structured problem-solving. Tree-of-Thought (ToT) \cite{yao2023tree} structures problem-solving as a tree-like process, exploring multiple reasoning pathways. Graph-of-Thought (GoT) \cite{yao2023beyond} extends this concept to graph-based reasoning, capturing complex interdependencies between logical steps.

Self-correction methods introduce iterative feedback loops to refine responses. STaR \cite{zelikman2022star} and CoRe \cite{zhu2022solving} employ Monte Carlo Tree Search (MCTS) to iteratively improve generated solutions. OmegaPRM \cite{luo2024improve} applies automated supervision to enhance Gemini's mathematical reasoning capabilities.

Socratic Teaching Methods \cite{nelson1980socratic} refine mathematical understanding through structured questioning. Socratic Models (SMs) \cite{zeng2022socratic} integrate dialogue-based learning, enabling LLMs to explore mathematical concepts interactively.

\subsection{Future Directions}
\label{Future Directions}

While LLM-based mathematical reasoning has made remarkable progress, several challenges remain:

\begin{itemize}
    \item Robust Verification: Ensuring LLMs generate provably correct mathematical solutions remains an open problem \cite{zhou2024don}.
    \item Handling Complex Proofs: Current models struggle with multi-step logical deductions in formal theorem proving \cite{dong2024formal}.
    \item Hybrid Symbolic-Neural Methods: Combining symbolic solvers with LLMs could enhance mathematical accuracy and reliability \cite{li2025neuro}.
\end{itemize}

Future research should focus on improving reasoning consistency \cite{wu2024enhancing}, integrating interactive problem-solving frameworks \cite{collins2024evaluating}, and enhancing mathematical comprehension through hybrid neural-symbolic approaches \cite{li2025neuro}.


\input{tabs/method_compare}

\input{tabs/method_results}

\input{tabs/tab_dataset}


\section{Comparison of Mathematical Large Language Models}\label{llm_comparisons}

Mathematical LLMs have evolved significantly as seen in benchmarks like MATH \cite{MATH} and geometry reasoning \cite{trinh2024solving}, leveraging various architectural designs, training methodologies, reasoning strategies, and tool integrations to enhance mathematical problem-solving capabilities \cite{patil2025advancing}.


However, open-source models lag behind proprietary ones like GPT-4 \cite{gpt4} and Gemini-Ultra \cite{gemini}. To bridge this gap, Deepseek \cite{deepseek-llm} introduced a domain-specific model (spmath) trained on the DeepSeekMath Corpus, a high-quality dataset of 120B math tokens curated from Common Crawl using a fastText-based classifier \cite{joulin2016fasttext}.

This model, initialized with DeepSeek-Coder-Base-v1.5 7B \cite{deepseek-coder}, outperforms Minerva 540B \cite{minerva} on MATH \cite{MATH} and exhibits strong multilingual performance on CMATH \cite{wei2023cmath} and AGIEval \cite{agieval}. Pre-training on code enhances mathematical reasoning, supporting the idea that coding skills improve logic-based problem-solving. spmath-Instruct 7B is fine-tuned using chain-of-thought \cite{cot}, program-of-thought \cite{pot, pal}, and tool-integrated reasoning \cite{tora}, achieving results comparable to 70B open-source instruction-tuned models. Additionally, the Deepseek introduced Group Relative Policy Optimization (GRPO), a RL method that eliminates the critic model, significantly reducing training costs while improving performance on GSM8K (82.9\% $\rightarrow$ 88.2\%) and MATH (46.8\% $\rightarrow$ 51.7\%). A unified paradigm is presented in \cite{deepseek-llm} to connect various RL-based fine-tuning methods, such as Rejection Sampling Fine-Tuning (RFT)\cite{yuan2023scaling}, Direct Preference Optimization (DPO) \cite{dpo}, and Proximal Policy Optimization (PPO) \cite{schulman2017proximal}, with extensive experiments exploring online vs. offline training, outcome vs. process supervision, and single-turn vs. iterative RL. Deepseek also constructs the DeepSeekMath Corpus using an iterative fastText-based classifier, filtering 35.5M math-related web pages while preventing contamination from evaluation datasets like GSM8K \cite{gsm8k} and MATH \cite{MATH}. Evaluations show spmath-Base 7B surpasses Mistral 7B \cite{mistral} and Llemma 34B \cite{llemma} in mathematical reasoning, excelling in theorem proving on miniF2F \cite{minif2f} and performing well in general reasoning (MMLU \cite{mmlu}, BBH \cite{bbh}) and coding (HumanEval \cite{codex}, MBPP \cite{mbpp}). After supervised fine-tuning with 776K mathematical instructions, spmath-Instruct 7B outperforms all 7B models and rivals 70B instruction-tuned models such as InternLM2-Math 20B, WizardMath 7B \cite{wizardmath}, MetaMath 70B \cite{metamath}, and MAmmoTH 70B \cite{MathInstruct}.


Table \ref{table: comparison of mathematical LMs} provides a comparative overview of multiple mathematical LLMs, highlighting key aspects such as training methods, reasoning techniques, RL applications, multi-modal capabilities, and accessibility. This section explores the observed trends, major advancements, and existing limitations in the domain of mathematical LLMs.

\subsection{Chain-of-Thought (CoT) Reasoning}
One of the most critical advancements in mathematical reasoning for LLMs is the incorporation of Chain-of-Thought (CoT) prompting \cite{wei2022chain}. CoT enables models to break down complex problems into intermediate steps, improving their ability to solve mathematical and logical problems. Foundational CoT (F-CoT) is a common feature among models such as MATH-PLM \cite{hendrycks2021measuring}, ControlMath \cite{chen2024controlmath}, and EURUS \cite{yuan2024advancing}, which helps them generate structured multi-step solutions. More advanced forms of CoT reasoning (A-CoT) are implemented in models like ToT \cite{yao2023tree}, Self-check \cite{miao2023selfcheck}, and Reflexion \cite{shinn2023reflexion}, allowing them to refine their solutions iteratively. Some recent models, such as QwQ \cite{qwq-32b-preview}, o1 \cite{openai_o1_system_card}, and LLaVA-o1 \cite{xu2024llava}, employ Long CoT, which extends reasoning over a longer context for more complex mathematical problems.

\subsection{Integration of External Tools}
Tool integration is another key differentiator among mathematical LLMs. Some models, such as Toolformer \cite{schick2023toolformer} and LeanReasoner\cite{jiang2024leanreasoner}, incorporate external symbolic solvers and calculators to enhance their accuracy. Other models, such as Codex-math \cite{drori2022neural} and MuMath-Code \cite{yin2024mumath}, rely on program-based reasoning, allowing them to execute mathematical scripts and verify calculations dynamically. These tool-enhanced approaches generally outperform models that rely purely on neural network inference without external computation.

\subsection{Training Paradigms: Pre-Training, Fine-Tuning, and Prompt Engineering}
Mathematical LLMs adopt different training strategies, primarily categorized into Pre-Training (Pre-T), Fine-Tuning (FT), and Prompt Engineering. Pre-trained models, such as Minerva \cite{lewkowycz2022solving}, GPT-4o \cite{hurst2024gpt}, and Gemini \cite{team2023gemini}, are trained on large mathematical datasets before being adapted for specific problem-solving tasks. Fine-tuned models, such as WizardMath \cite{luo2023wizardmath} and InternLM-Math \cite{ying2024internlmmath}, undergo domain-specific adaptations to improve their mathematical reasoning skills. Prompt-based models, such as PAL \cite{gao2023pal} and PoT \cite{chen2022program}, rely on structured prompt design to guide their reasoning, making them more flexible for zero-shot and few-shot learning.

\subsection{RL for Mathematical Reasoning}
RL techniques, particularly Proximal Policy Optimization (PPO) and Direct Preference Optimization (DPO), have been applied to mathematical LLMs to refine their solution generation process. Models like WizardMath \cite{luo2023wizardmath} and InternLM-Math \cite{ying2024internlmmath} leverage PPO to enhance their performance through iterative self-improvement. Additionally, DPO-based models, such as EURUS \cite{yuan2024advancing} and V-STaR \cite{hosseini2024v}, optimize their outputs based on user preferences and ranking mechanisms. Recent advances, including Step-DPO \cite{lai2024step} and Flow-DPO \cite{deng2024flow}, incorporate structured planning into reinforcement learning to refine problem-solving pathways.

\subsection{Multi-Modal Capabilities}
While most mathematical LLMs focus on text-based problem-solving, a growing number of models support multi-modal reasoning, integrating text and image inputs to solve visual mathematics problems. Notable multi-modal models include GPT-4V \cite{openai2023gpt4}, Gemini \cite{team2023gemini}, and GLM-4V \cite{glm2024chatglm}, which incorporate vision-language capabilities for tasks involving diagrams, plots, and mathematical notation recognition. Similarly, Math-LLaVA \cite{shihu2024mathllava} and AtomThink \cite{xiang2024atomthink} extend multi-modal reasoning by fine-tuning on specialized mathematical datasets. However, many models still lack this capability, indicating an area for further improvement.

\subsection{Open vs. Closed Access Models}
The accessibility of mathematical LLMs varies significantly. Open-source models such as EURUS \cite{yuan2024advancing}, Qwen-Math \cite{yang2024qwen2}, and LINC \cite{olausson2023linc} provide transparency and adaptability, allowing researchers to fine-tune them for specific applications. In contrast, proprietary models such as GPT-4o \cite{hurst2024gpt}, Gemini \cite{team2023gemini}, and o1 \cite{openai_o1_system_card} are closed-source, limiting external modifications. The trade-off between openness and performance remains a key consideration in model selection.

Mathematical LLMs continue to advance in reasoning capabilities, tool integration, and training methodologies. While Chain-of-Thought reasoning has become a standard technique for improving accuracy, models differ in their use of external tools, reinforcement learning, and multi-modal support. Additionally, open-source models provide flexibility for customization, whereas closed-source models typically offer superior performance at the cost of accessibility. Future advancements are likely to focus on enhancing multi-modal mathematical reasoning, better tool integrations, and more robust self-improving learning mechanisms to achieve greater accuracy and efficiency in solving complex mathematical problems.


\section{Performance Analysis of Mathematical Language Models}\label{llm_performance_analysis}

Mathematical LMs have demonstrated varying degrees of success across different mathematical benchmarks. To assess the arithmetic and mathematical capabilities of language models, various datasets and evaluation methods have been proposed. Commonly used benchmarks include MATH \cite{hendrycks2021measuring}, GSM8K \cite{cobbe2021training}, and AGIEval \cite{zhong2023agieval}, which test models across different difficulty levels and problem types.
Traditional NLP evaluation metrics, such as accuracy and perplexity, are insufficient for mathematical reasoning. Instead, numeric predictions are evaluated using regression metrics such as Mean Absolute Error (MAE) and Root Mean Squared Error (RMSE) \cite{berg-kirkpatrick-spokoyny-2020-empirical}. Adjusted Perplexity \cite{spithourakis2018numeracy} has also been proposed to account for numeric token distributions.
		
Some benchmarks adopt specialized evaluation strategies. The MATH dataset \cite{hendrycks2021measuring} enforces strict answer formatting for consistency, while AGIEval \cite{zhong2023agieval} employs Exact Match and F1 scores for fill-in-the-blank questions. Recent research explores automated assessment methods, including large language model-based scoring \cite{he2023socreval}, to provide more nuanced evaluations of reasoning quality.
		
Given the complexity of mathematical problem-solving, developing comprehensive evaluation metrics remains an open challenge. Future directions should focus on refining benchmarks to assess multi-step reasoning, logical consistency, and theorem-proving capabilities more effectively.

Table~\ref{tab_performance} presents the performance of various text-based and multi-modal LMs on multiple datasets, including MATH, GSM8K, MMLU, ASDiv, AIME, and others. 

Among the text-based methods, Codex-math achieves the highest accuracy on the MATH dataset (81.8\%), while QwQ leads with a remarkable 90.6\%. For GSM8K, MCTSr outperforms other models with an accuracy of 96.6\%, followed closely by Diversity-of-Thought (96.3\%) and Step-DPO (94.0\%). In the AIME dataset, Self-check achieves the highest performance with 81.2\%. Similarly, in ASDiv, EURUS attains the best result (93.0\%), demonstrating its effectiveness in arithmetic reasoning tasks.

For dataset-specific performance, PoT leads in MathQA (89.1\%), and CR achieves the highest accuracy on the FOLIO dataset (98.0\%). These results highlight the strengths of specialized models in certain tasks. Meanwhile, InternLM-Math and Qwen-Math show strong overall performance across multiple datasets, particularly with Qwen-Math achieving 91.6\% in GSM8K and 82.8\% in MMLU.

Multi-modal models, which leverage both textual and visual input, generally perform better than their text-based counterparts. o1 outperforms all models on the MATH dataset with an accuracy of 94.8\%, and it also leads in MMLU (92.3\%) and AIME (83.3\%). GPT-4o also exhibits strong performance in MMLU (88.0\%), indicating the advantage of integrating vision and language for mathematical reasoning. Qwen2-VL and AtomThink show competitive results on multimodal datasets, with the latter achieving the highest Math-V score (40.5\%).

Overall, the results suggest that newer models incorporating diverse reasoning strategies, self-consistency, and program-of-thought approaches achieve higher accuracy. Future advancements in multi-modal LMs are likely to further improve mathematical reasoning capabilities by integrating structured symbolic processing with deep learning architectures.

\section{Datasets}
\label{Datasets}

To train and evaluate the arithmetic and mathematical reasoning abilities of language models, various math word problem (MWP) datasets \cite{saxton2018analysing,hendrycks2021measuring,cobbe2021training,kushman-etal-2014-learning} have been developed. We categorize these datasets into training, benchmark, and augmented datasets (Table \ref{table:statistis_datasets}).

\subsection{Training Datasets}
\label{Training Datasets}

Mathematical Calculation datasets extract numerical information from text. Clinical Data \cite{spithourakis2016numerically} contains patient reports with structured attributes, while Scientific Data \cite{spithourakis2018numeracy} includes research papers. DoQ \cite{elazar2019large} captures scalar attribute values for over 350K nouns and verbs. VERBPHYSICS \cite{forbes2017verb} encodes action-object relations, and DROP \cite{dua2019drop} tests discrete reasoning over adversarial questions.

Math Word Problems datasets include AddSub \cite{hosseini-etal-2014-learning}, focusing on addition and subtraction, and SingleOp \cite{roy2015reasoning}, covering multiplication and division. GSM8K \cite{cobbe2021training} comprises high-quality grade school problems requiring multi-step reasoning. Math23k \cite{wang2017deep} and MATHPILE \cite{wang2023generative} provide structured equations. To improve semantic parsing, Dolphin1878 \cite{shi2015automatically} and SingleEq \cite{koncel2015parsing} enable equation mapping. Recent datasets like HMWP \cite{qin2020semantically} and MathQA \cite{amini2019mathqa} expand problem diversity with multilingual and multiple-choice questions.

Theorem Proving datasets such as Feit-Thompson \cite{huang2018gamepad} and IsarStep \cite{li2020isarstep} provide structured theorem proofs. CoqGym \cite{yang2019learning} and LeanStep \cite{han2021proof} integrate proof tactics, while NaturalProofs \cite{welleck2021naturalproofs} covers formal and informal mathematics.

\subsection{Benchmark Datasets}
\label{Benchmark Datasets}

Benchmark datasets evaluate mathematical reasoning. MAWPS \cite{koncel2016mawps} and Dolphin18K \cite{huang2016well} contain diverse MWPs. MATH \cite{hendrycks2021measuring} provides competition-level problems with step-by-step solutions. MGSM \cite{shi2022language} extends GSM8K into 10 languages. miniF2F \cite{zheng2021minif2f} benchmarks neural theorem proving across Lean, Isabelle, and HOL Light. FOLIO \cite{DBLP:conf/emnlp/HanS0QRZCPQBSWS24} offers first-order logic reasoning annotations.

Recent multimodal benchmarks assess vision-language models. MathVista \cite{lu2023mathvista} evaluates diagram-based problems, while MMMU-Math \cite{yue2024mmmumassivemultidisciplinemultimodal} and U-Math \cite{chernyshev2024umathuniversitylevelbenchmarkevaluating} focus on university-level problem solving.

\subsection{Augmented Datasets}
\label{Augmented Datasets}

Augmented datasets enrich existing collections. Aggregate \cite{roy2018mapping} extends AllArith with additional perturbed problems. MetaMathQA \cite{yu2023metamath} reformulates mathematical questions using GPT-4. Math5K \cite{li2023camel} generates 50,000 new problems.

To enhance reasoning transparency, MathQA-Python \cite{austin2021program} converts text to code, and Lila \cite{mishra2022lila} provides Python-based solutions. PEN \cite{kim2022ept} annotates algebraic explanations, while PRM800K \cite{DBLP:conf/iclr/LightmanKBEBLLS24} labels step-wise mathematical reasoning correctness. MiniF2F+informal \cite{jiang2022draft} and NaturalProofs-Gen \cite{welleck2022naturalprover} introduce informal theorem statements to theorem-proving datasets.


\section{LLMs for Time Series Analysis: Challenges, Methods, and Future Directions}\label{challenges_methods}

LLMs, such as GPT-4o and LLaMA~\cite{touvron2023llama2}, have achieved remarkable success in natural language processing. Inspired by this, researchers have started exploring their application in time series analysis for tasks such as forecasting, classification, and anomaly detection~\cite{wu2022timesnet,garza2023timegpt}. By leveraging their strong pattern recognition capabilities, LLMs have been integrated into various domains where time series data play a crucial role. These applications span across finance~\cite{yu2023temporal,lopez2023can}, healthcare~\cite{li2023frozen,jiang2023health}, traffic forecasting~\cite{liu2024spatialtemporal}, and mobility prediction~\cite{wang2023i}. In finance, LLMs assist in stock market forecasting by analyzing textual and numerical data. In healthcare, LLMs are leveraged for Electrocardiogram classification and electronic health record analysis \cite{li2023frozen}.

However, applying LLMs to time series data presents unique challenges: (1) diverse formats and domains make model training complex, and (2) real-world time series data often exhibit non-stationary properties, leading to concept drift that impedes continuous adaptation~\cite{kim2021reversible}.

To address these challenges, recent works exploit pre-trained LLMs to analyze time series data without training new models from scratch. This section systematically categorizes these efforts into five key strategies: (1) direct querying of LLMs, (2) tokenization techniques for numerical data, (3) prompt engineering, (4) fine-tuning methods, and (5) integration of LLMs with existing time series models. Additionally, we discuss domain-specific applications and propose future research directions.

\subsection{Taxonomy of LLMs in Time Series Analysis}
\textbf{General Pipeline.} LLM-based time series analysis can be performed through three main approaches: direct querying, fine-tuning, or integrating LLMs into models. Fine-tuning involves tokenizing time series data, enhancing representation with prompts, and adapting LLMs for downstream tasks.

\textbf{Direct Query.} Works like PromptCast~\cite{xue2023promptcast} and LLMTime~\cite{gruver2023llmtime} use zero-shot or few-shot learning by querying LLMs with structured prompts.

\textbf{Tokenization.} Patching techniques~\cite{zhou2023onefitsall,chang2023llm4ts} and STL decomposition~\cite{cao2023tempo} help convert numerical time series data into a format interpretable by LLMs.

\textbf{Prompt Design.} Context-enhanced and soft prompts~\cite{sun2023test,jin2023time} improve the ability of LLMs to process time series data effectively.

\textbf{Fine-tuning.} Techniques such as LoRA~\cite{hu2021lora} help optimize specific layers of LLMs to adapt them to time series tasks~\cite{chang2023llm4ts,zhou2023onefitsall}.

\textbf{Model Integration.} Some methods use LLMs as feature enhancement tools in multimodal frameworks~\cite{shi2023language,xu2023language,li2023frozen}.

Open challenges include:
\begin{itemize}
    \item Improved tokenization and prompts to better represent time series patterns~\cite{rasul2023lagllama}.
    \item Interpretability via prototype-based or gradient-based methods~\cite{mai2023enhanced}.
    \item Multi-modality fusion techniques for integrating time series with text, images, and tabular data.
    \item Domain generalization techniques for robust performance across diverse datasets.
    \item Scaling laws to understand how model size affects time series performance.
    \item Time series LLM agents for decision-making applications.
    \item Bias and safety mechanisms to ensure reliability in critical domains.
\end{itemize}


\begin{table*}[t]
  \tiny
  \setlength{\tabcolsep}{4.3pt}
  \renewcommand{\arraystretch}{0.85}
  \scriptsize
  \caption{\footnotesize{Taxonomy of LLM-based time series methods. The data type \textbf{TS} denotes standard time series, \textbf{STS} represents spatiotemporal time series, and the prefix \textbf{MM-} indicates multimodal inputs. The task entry \textbf{Multiple} includes forecasting, classification, imputation, and anomaly detection. \textbf{Query} refers to direct LLM-based querying, \textbf{Token} indicates time series tokenization design, \textbf{Prompt} denotes the use of textual or parameterized prompts, \textbf{Tune} represents fine-tuning of LLM parameters, and \textbf{Fusion} specifies whether LLMs are integrated into external models for downstream tasks.}}
  \centering
  \tiny
  \begin{threeparttable}
    \begin{tabular}{l c c c c c c c c c c}
    \toprule
    Method & Data Type & Domain & Task & Query & Token & Prompt & Tune & Fusion & LLM & Code \\
    \midrule
    Time-LLM~\cite{jin2023time} & MM-TS & General & Forecasting & \xmark & \cmark & \cmark & \cmark & \xmark & LLaMA, GPT-2 & Yes$^{\text{[1]}}$ \\ 
    OFA~\cite{zhou2023onefitsall} & TS & General & Multiple & \xmark & \cmark & \xmark & \cmark & \xmark & GPT-2 & Yes$^{\text{[2]}}$ \\ 
    TEMPO~\cite{cao2023tempo} & TS & General & Forecasting & \xmark & \cmark & \cmark & \cmark & \xmark & GPT-2 & No \\ 
    TEST~\cite{sun2023test} & MM-TS & General & Forecasting, Classification & \xmark & \cmark & \cmark & \xmark & \cmark & BERT, GPT-2, ChatGLM, LLaMA2 & Yes$^{\text{[3]}}$ \\ 
    LLM4TS~\cite{chang2023llm4ts} & TS & General & Forecasting & \xmark & \cmark & \xmark & \cmark & \xmark & GPT-2 & No \\ 
    PromptCast~\cite{xue2023promptcast} & TS & General & Forecasting & \cmark & \xmark & \cmark & \xmark & \xmark & Bart, BERT & Yes$^{\text{[4]}}$ \\ 
    LAMP~\cite{shi2023language} & TS & General & Event Prediction & \cmark & \xmark & \cmark & \xmark & \cmark & GPT-3\&3.5, LLaMA-2 & Yes$^{\text{[6]}}$ \\ 
    Yu et al.~\cite{yu2023temporal} & MM-TS & Finance & Forecasting & \cmark & \xmark & \cmark & \cmark & \xmark & GPT-4, Open-LLaMA & No \\ 
    METS~\cite{li2023frozen} & MM-TS & Healthcare & Classification & \cmark & \xmark & \cmark & \xmark & \cmark & ClinicalBERT & No \\ 
    Jiang et al.~\cite{jiang2023health} & MM-TS & Healthcare & Classification & \xmark & \xmark & \xmark & \cmark & \xmark & NYUTron (BERT) & Yes$^{\text{[8]}}$ \\ 
    AuxMobLCast~\cite{xue2022leveraging} & STS & Mobility & Forecasting & \xmark & \xmark & \cmark & \cmark & \cmark & BERT, RoBERTa, GPT-2, XLNet & Yes$^{\text{[9]}}$ \\ 
    ST-LLM~\cite{liu2024spatialtemporal} & STS & Traffic & Forecasting & \xmark & \cmark & \xmark & \cmark & \xmark & LLaMA, GPT-2 & No \\ 
    GATGPT~\cite{chen2023gatgpt} & STS & Traffic & Imputation & \xmark & \cmark & \xmark & \cmark & \xmark & GPT-2 & No \\ 
    LA-GCN~\cite{xu2023language} & MM-STS & Vision & Classification & \xmark & \cmark & \xmark & \xmark & \cmark & BERT & Yes$^{\text{[11]}}$ \\ 
    \bottomrule
    \end{tabular}
    
    \begin{tablenotes}
      \tiny
      \item{} [1] \url{https://github.com/kimmeen/time-llm} \quad [2] \url{https://github.com/DAMO-DI-ML/NeurIPS2023-One-Fits-All} \quad [3] \url{https://openreview.net/forum?id=Tuh4nZVb0g}
      \item{} [4] \url{https://github.com/HaoUNSW/PISA} \quad [5] \url{https://github.com/ngruver/llmtime} \quad [6] \url{https://github.com/iLampard/lamp} \quad [7] \url{https://github.com/ZihanChen1995/ChatGPT-GNN-StockPredict}
      \item{} [8] \url{https://github.com/nyuolab/NYUTron} \quad [9] \url{https://github.com/cruiseresearchgroup/AuxMobLCast} \quad [10] \url{https://github.com/xlwang233/LLM-Mob} \quad [11] \url{https://github.com/damNull/LAGCN}
    \end{tablenotes}
    
  \end{threeparttable}
  \vspace{-0.4cm}
  \label{tab:taxonomy}
\end{table*}




\section{Comparison of commonly used LLMs based on their technical specifications}\label{comparisons_famous_llms}

LLMs have evolved significantly, with various models offering different strengths and weaknesses. In this section we compare some famous LLMs including GPT-4 Turbo, LLaMA3, DeepSeek-R1, Mixtral, Claude3, and Gemini1.5. These models differ in terms of architecture, training data, computational efficiency, response accuracy, and suitability for specific applications. Some models prioritize speed and cost-effectiveness, while others focus on reasoning capabilities, multilingual support, or specialized domain knowledge.
		
		\subsection{GPT-4 Turbo (OpenAI)}
		\textbf{Pros:}
		\begin{itemize}
			\item Best overall reasoning and language understanding.
			\item Supports multi-modal input (text and images).
			\item Handles up to 128K tokens\footnote{100 tokens are about 60-80 English words} \cite{achiam2023gpt}.
			\item Strong API ecosystem and enterprise adoption.
		\end{itemize}
		
		\textbf{Cons:}
		\begin{itemize}
			\item Closed-source, making fine-tuning difficult.
			\item Expensive compared to open-source alternatives.
			\item Some guardrails limit responses to certain queries.
		\end{itemize}
		
		\subsection{LLaMA 2 \& 3 (Meta)}
		\textbf{Pros:}
		\begin{itemize}
			\item Open-source, allowing fine-tuning and self-hosting.
			\item Efficient performance and competitive results.
			\item Strong research community support.
		\end{itemize}
		
		\textbf{Cons:}
		\begin{itemize}
			\item Weaker reasoning than GPT-4 Turbo.
			\item Shorter context windows (4K–32K tokens) \cite{dubey2024llama}.
			\item Requires high-end GPUs for full deployment.
		\end{itemize}

		\subsection{DeepSeek-R1 (DeepSeek)}
		\textbf{Pros:}
		\begin{itemize}
			\item Cost-Effective Development: Achieved performance comparable to GPT-4 at a fraction of the cost.
			\item Efficient Resource Utilization: Trained using about 2,048 NVIDIA H800 GPUs, significantly fewer than competitors \cite{guo2025deepseek}.
			\item Open-Source Commitment: Models are freely available under the MIT license, promoting transparency and customization.
			\item Rapid Adoption: Quickly became the most downloaded application on the social media, surpassing competitors like ChatGPT.
		\end{itemize}
		
		\textbf{Cons:}
		\begin{itemize}
			\item Limited Language Support: Primarily optimized for English and Chinese, with less proficiency in other languages.
			\item Less Established Ecosystem: Being relatively new, it lacks the extensive integration and community support seen with models like GPT-4.
		\end{itemize}
		
		\textbf{Best for:} Users and developers seeking a cost-effective, open-source alternative with strong performance in English and Chinese.

		\subsection{Mixtral (Mistral AI)}
		\textbf{Pros:}
		\begin{itemize}
			\item Mixture of Experts (MoE) architecture for efficiency.
			\item Open-source and highly optimized.
			\item Compact and fast inference speed.
		\end{itemize}
		
		\textbf{Cons:}
		\begin{itemize}
			\item Shorter context length than GPT-4 (4K-32K tokens) \cite{jiang2024mixtral}.
			\item Less adoption in enterprise applications.
		\end{itemize}
		
		\subsection{Claude 3 (Anthropic)}
		\textbf{Pros:}
		\begin{itemize}
			\item Best alignment and ethical AI practices.
			\item Handles very long documents (up to 200K tokens \cite{anthropic2024introducing}).
			\item More human-like responses and high creativity.
		\end{itemize}
		
		\textbf{Cons:}
		\begin{itemize}
			\item Slower response times compared to GPT-4 Turbo.
			\item Closed-source, requiring API access.
		\end{itemize}
		
		\subsection{Gemini 1.5 Pro (Google)}
		\textbf{Pros:}
		\begin{itemize}
			\item Strong multi-modal capabilities (text, images, and audio).
			\item Deep integration with Google services.
			\item Very long context windows (up to 2M tokens \cite{team2024gemini}).
		\end{itemize}
		
		\textbf{Cons:}
		\begin{itemize}
			\item Still catching up to OpenAI in benchmarks.
			\item Limited open-source versions.
		\end{itemize}

Table \ref{tab_llm_comparison} summarizes the technical specifications of commonly used LLMs based on context length, strength, and weaknesses.

        \begin{table*}[ht]
            \centering
            \caption{Comparison of Commonly used LLMs based on their technical specifications}
            \label{tab_llm_comparison}
            \begin{tabular}{lcccc}
                \toprule
                \textbf{Model} & \textbf{Open-Source} & \textbf{Context Length} & \textbf{Strengths} & \textbf{Weaknesses} \\
                \midrule
                GPT-4 Turbo & \xmark & 128K  & Best reasoning, multi-modal & Expensive, closed-source \\
                LLaMA 2/3   & \cmark & 4K--32K & Open-source, efficient & Weaker reasoning \\
                DeepSeek-R1 & \cmark & 12K  & Good multilingual support & Less adoption \\
                Mixtral     & \cmark & 4K--32K  & Efficient, compact & Shorter context \\
                Claude 3    & \xmark & 200K  & Ethical AI, document processing & Slower responses \\
                Gemini 1.5 Pro  & \xmark & 2M    & Multi-modal, long context & Not fully open-source \\
            \bottomrule
    \end{tabular}
\end{table*}



\section{Application of LLms on optimization and control}\label{applications}

LLMs in solving control and optimization problems is essential for understanding their strengths, limitations, and potential applications. As these models continue to advance, their capabilities in multi-step reasoning, broad knowledge retrieval, and intuitive code generation make them promising tools for tackling complex computational tasks. However, their limitations—such as logical inconsistencies, sensitivity to prompt variations, and susceptibility to errors—necessitate a rigorous assessment of their reliability in control and optimization scenarios.

A systematic evaluation of LLMs in these domains allows us to determine the extent to which they can autonomously handle problem-solving, adapt to dynamically changing parameters, and refine solutions iteratively. Moreover, this investigation provides critical insights into how LLMs can complement traditional solvers, optimizing human-AI workflows and enhancing the safety, robustness, and efficiency of solutions in real-world applications. Additionally, by testing these models in structured optimization problems, we may uncover novel solution strategies that lead to improved computational efficiency and innovation in control systems.

To address this objective, we systematically test LLMs across a range of control and optimization tasks, including:

(i) Mixed-Integer Linear Programming (MILP) for Battery Energy Storage Systems (BESS): Optimizing charge/discharge scheduling while minimizing operational costs and adhering to system constraints.

(ii) Linear Quadratic Control (LQC) for water level regulation in interconnected tanks: Evaluating the ability of LLMs to design and analyze optimal control strategies for fluid management.

(iii) Climate dataset analysis: Utilizing LLMs for pattern recognition, forecasting, and decision-making based on historical climate data.

(iv) Multi-Agent Linear Programming Optimization: Exploring how LLMs can assist in coordinating multiple decision-making agents to achieve optimal solutions.

By rigorously testing LLMs in these diverse applications, we aim to identify their strengths and weaknesses in solving complex mathematical optimization and control problems. This research will inform best practices for integrating LLMs into existing computational workflows, ultimately advancing the field of AI-driven optimization and control. 


\paragraph{Code availability statement}
The Python scripts that we use for each example leverage multiple libraries to perform data processing, optimization, and AI-driven analysis. They utilize NumPy \cite{harris2020array} and Pandas \cite{mckinney2010data} for numerical and tabular data manipulation, while Matplotlib \cite{hunter2007matplotlib} is used for visualization. Scipy \cite{mckinney2010data} provides numerical optimization functions, particularly for linear programming, and Scikit-learn \cite{pedregosa2011scikit} is employed for preprocessing tasks like scaling. For geospatial analysis, Cartopy \cite{met2010cartopy} is used to map wind data. The asyncio \cite{alemany2018diseno} library enables asynchronous execution, ensuring smooth parallel processing, especially in federated optimization scenarios. The script also incorporates OpenAI's API to generate insights from wind and battery storage data, assisting in decision-making. Furthermore, PuLP \cite{mitchell2011pulp} is used for mixed-integer linear programming (MILP) to optimize battery charge/discharge schedules dynamically. Finally, AutoGen \cite{wu2023autogen} is used to create multi-agent frameworks for executing linear programming problems asynchronously. In Appendix \ref{repository} we provide the link to the repository containing the codes that are used for the simulation.


\subsection{An example of the LLm application on the BESS}	
This section details an optimization problem for the operation of a BESS with LLMs. Our objective is to minimize the net operational cost (or maximize net revenue) by scheduling the charging and discharging of a battery over a discrete 10-hour planning horizon. We incorporate fuzzy logic rules to capture realistic decision-making heuristics concerning electricity prices, battery state of charge (SOC), and operational switching considerations.

\subsubsection{System Overview}
	
We consider a BESS with the following characteristics:
\begin{itemize}
	\item Battery capacity: $SOC_{\max} = 100\ \text{kWh}$
	\item Initial SOC: $SOC_0 = 50\ \text{kWh}$
	\item Charging efficiency: $\eta = 0.90$
	\item Discharging efficiency: $\frac{1}{\eta} = 0.90$
	\item Max charge/discharge power: $P_{\text{charge}}^{\max} = P_{\text{discharge}}^{\max} = 20\ \text{kW}$
\end{itemize}
We operate over time steps $t = 1,\dots,10$, each representing a one-hour interval.
\subsubsection{Decision Variables}
\begin{itemize}
	\item $P_{\text{charge}}(t)\ [\text{kW}]$: Charging power during hour $t$.
	\item $P_{\text{discharge}}(t)\ [\text{kW}]$: Discharging power during hour $t$.
	\item $SOC(t)\ [\text{kWh}]$: State of charge of the battery at time $t$.
\end{itemize}
	
\subsubsection{Parameters}
	
	\subsubsection*{Electricity Prices and Revenues}
	
	Let
	\[
	C = \bigl[\,20,\,25,\,15,\,10,\,30,\,35,\,40,\,20,\,15,\,10\bigr]\quad \text{(\$/kWh)},
	\]
	\[
	R = \bigl[\,30,\,35,\,25,\,20,\,40,\,45,\,50,\,30,\,25,\,20\bigr]\quad \text{(\$/kWh)}.
	\]
	Here, $C_t$ is the cost of purchasing electricity in hour $t$, and $R_t$ is the revenue rate from selling electricity in hour $t$.
	
	\subsection{Constraints}
	
	\paragraph{Battery Capacity Bound}
	\[
	0 \;\le\; SOC(t) \;\le\; SOC_{\max}, \quad \forall t.
	\]
	
	\paragraph{Power Limits}
	\[
	0 \;\le\; P_{\text{charge}}(t) \;\le\; P_{\text{charge}}^{\max} = 20, \quad \forall t,
	\]
	\[
	0 \;\le\; P_{\text{discharge}}(t) \;\le\; P_{\text{discharge}}^{\max} = 20, \quad \forall t.
	\]
	
	\paragraph{Energy Balance}

    \begin{multline}
    SOC(t+1) = SOC(t)\;+\;\eta \cdot P_{\text{charge}}(t)\\ \;-\;\frac{1}{\eta}\cdot P_{\text{discharge}}(t), 
	\quad t = 1,\ldots,9.
    \end{multline}

	We take $SOC(1)=SOC_0=50\,\text{kWh}$ as the initial condition.
	
	\paragraph{No Simultaneous Charge/Discharge}
	\[
	P_{\text{charge}}(t)\,\cdot\,P_{\text{discharge}}(t) \;=\; 0, \quad \forall t.
	\]
	This ensures the battery cannot charge and discharge in the same hour.
	
	\subsection{Objective Function}
	
	We wish to minimize the total net cost $Z$ over the 10-hour horizon:
	\[
	\min Z \;=\; \sum_{t=1}^{10}\Bigl(C_t \cdot P_{\text{charge}}(t) \;-\; R_t \cdot P_{\text{discharge}}(t)\Bigr).
	\]
	The term $C_t \, P_{\text{charge}}(t)$ represents the cost of charging, while $-\, R_t \, P_{\text{discharge}}(t)$ represents revenue gained from discharging.
	
	\subsubsection{Fuzzy Logic Integration}
	
	We enrich the classical optimization with fuzzy logic rules to guide or penalize decisions that deviate from human-like strategies. The fuzzy rules help shape how aggressively the system charges or discharges based on electricity prices, battery SOC, and the desire to avoid frequent switching.
	
	
	\begin{enumerate}
		\item If the electricity price $C_t$ is high, prioritize discharging.
		\item If the battery SOC is below $30\%$ of $SOC_{\max}$, limit discharging and prioritize charging.
		\item If $C_t$ is low, delay or reduce charging unless SOC is critically low.
		\item Avoid frequent switching between charging and discharging.
	\end{enumerate}
	


	In Table \ref{BESS_llm} (Appendix) we have shown an example of how an hour-by-hour schedule might look based on the output of the LLM. Actual solutions may vary depending on specific fuzzy membership functions, penalty weights, and solver nuances. In addition, the LLM model is also plays an important role for the response to the \texttt{prompt}. In this example we used GPT4O LLM model.

	\paragraph{Cost Computation Example}
	The total cost $Z$ is computed by
	\[
	Z = \sum_{t=1}^{10} 
	\Bigl(
	C_t \cdot P_{\text{charge}}(t) \;-\; R_t \cdot P_{\text{discharge}}(t)
	\Bigr).
	\]
	If, for instance, at hour $2$ we have $P_{\text{charge}}(2) = 0$ and $P_{\text{discharge}}(2) = 10$, then the hour-$2$ cost contribution is
	\[
	\text{Cost}_{2} = 25 \times 0 \;-\; 35 \times 10 \;=\; -350 \;\$,
	\]
	indicating a net profit of \$350 at hour 2 (negative cost).


\subsection{LLM-Driven Fuzzy Logic Control for Water Level Regulation in Interconnected Tanks}
Water level regulation in interconnected tank systems is a fundamental problem in control engineering, frequently addressed using classical and modern control techniques such as Proportional-Integral-Derivative (PID) controllers, Model Predictive Control (MPC), or fuzzy logic controllers. 

In this section, we integrate a LLM as a decision-making component in a fuzzy logic-based control system to optimize the regulation of water levels while minimizing a quadratic cost function. This novel approach leverages natural language processing (NLP)-based reasoning within an LLM to generate real-time control decisions based on predefined fuzzy logic rules.

\subsubsection{System Model}
The dynamics of the interconnected tanks are represented using a state-space formulation, where the water levels in the two tanks form the system state:
\begin{equation}
    \mathbf{x}(t) = \begin{bmatrix} x_1(t) \\ x_2(t) \end{bmatrix},
\end{equation}
The system evolves according to the linear state-space model:
\begin{equation}
    \dot{\mathbf{x}}(t) = \mathbf{A} \mathbf{x}(t) + \mathbf{B} u(t),
\end{equation}
where:
\begin{equation}
    \mathbf{A} = \begin{bmatrix} -0.1 & 0.05 \\ 0.05 & -0.1 \end{bmatrix}, \quad
    \mathbf{B} = \begin{bmatrix} 0.1 \\ 0 \end{bmatrix},
\end{equation}
\noindent The control input \( u(t) \) represents the pump flow rate, which is constrained within the range \( [-u_{\max}, u_{\max}] \), with \( u_{\max} = 0.5 \). The cost function to be minimized is a quadratic penalty function:
\begin{equation}
    J = Q_1 (x_1 - x_{1,\text{desired}})^2 + Q_2 (x_2 - x_{2,\text{desired}})^2 + R u^2,
\end{equation}
where:
\begin{itemize}
    \item \( Q_1 = 10 \), \( Q_2 = 8 \) are state penalty weights,
    \item \( R = 2 \) penalizes excessive control effort,
    \item \( x_{1,\text{desired}} = 2.0 \), \( x_{2,\text{desired}} = 1.5 \) are the target levels.
\end{itemize}

\subsubsection{Fuzzy Logic-Based Control Using LLM}
In this scenario the LLM is integrated as an inference engine that follows predefined fuzzy logic rules to determine the control action \( u(t) \) at each time step. The control strategy involves:
\begin{enumerate}
    \item Assessing Deviation: The LLM compares the actual water levels \( x_1, x_2 \) with their desired levels.
    \item Applying Fuzzy Rules:
    \begin{itemize}
        \item If \( x_1 > x_{1,\text{desired}} \), decrease \( u \).
        \item If \( x_2 < x_{2,\text{desired}} \), increase \( u \).
        \item Balance the adjustments to avoid oscillations.
    \end{itemize}
    \item Minimizing the Cost Function: The LLM aims to select \( u(t) \) that optimally reduces deviations while considering energy efficiency.
\end{enumerate}

At each time step \( t \), a \texttt{prompt} (see Algorithm \ref{tank_level_control_alg}) is dynamically generated and sent to the LLM (see appendix \ref{tank_level_prompt}). In Algorithm \ref{tank_level_control_alg} we present a general procedure that is used for tank level control.

The LLM generates a response, which is parsed to extract the control input. If parsing fails, a fallback \( u = 0 \) is used.

\begin{algorithm}\label{tank_level_control_alg}
    \caption{LLM-Based Fuzzy Control for Interconnected Tanks}
    \begin{algorithmic}[1]
        \State \textbf{Input:} Initial state $\mathbf{x}_0 = [x_1, x_2]^\top$, desired state $\mathbf{x}_d = [x_{1,\text{desired}}, x_{2,\text{desired}}]^\top$
        \State \textbf{Parameters:} 
        \State \quad System matrices $\mathbf{A}$, $\mathbf{B}$
        \State \quad Quadratic cost weights $Q_1, Q_2, R$
        \State \quad Maximum pump flow rate $u_{\max}$
        \State \quad Time step $\Delta t$, total time $T$
        
        \State \textbf{Initialize:} Set $t = 0$, state $\mathbf{x} = \mathbf{x}_0$, control input $u = 0$

        \While{$t < T$}
            \State Construct \textbf{\texttt{\emph{prompt}}} with current state and system parameters
            \State Send prompt to LLM to obtain fuzzy logic-based control input $u(t)$
            \State Parse the LLM response to extract $u(t)$
            \State Ensure control input is within bounds: 
            \[
                u(t) \leftarrow \text{clip}(u(t), -u_{\max}, u_{\max})
            \]
            \State Compute state derivative:
            \[
                \dot{\mathbf{x}} = \mathbf{A} \mathbf{x} + \mathbf{B} u
            \]
            \State Update system state:
            \[
                \mathbf{x} \leftarrow \mathbf{x} + \dot{\mathbf{x}} \cdot \Delta t
            \]
            \State  $(t, x_1, x_2, u)$ for analysis
            \State Increment time: $t \leftarrow t + \Delta t$
        \EndWhile

        \State \textbf{Output:} Time trajectory of states and control inputs
    \end{algorithmic}
\end{algorithm}

The model is simulated over a 10-second horizon, with the LLM making decisions at each step. The summary of the simulation results produced by LLM is shown in Table \ref{tank_level_table}.

\subsection{Application of Large Language Models in Time Series dataset analysis}
In this section we consider the application of LLMs for Wind and Pressure Data Analysis for renewable energies. We describe the implementation of a climate data analysis framework integrating machine learning techniques for wind speed and pressure assessment in Germany \cite{forootani2024climate}. The system processes CMIP6 climate datasets, extracts and interpolates meteorological variables, computes statistical features, and formulates queries for a LLM to analyze patterns and trends. Here we detail the algorithmic structure and highlights how LLMs assist in interpreting climate data for renewable energy applications. Wind energy forecasting and optimal wind turbine placement require an extensive analysis of historical meteorological data. Traditional statistical methods provide limited insights due to their inability to capture complex interactions among different environmental factors. The integration of advanced machine learning techniques, specifically LLMs, offers a promising approach to identifying patterns, anomalies, and correlations in meteorological datasets. The objective here is to: (i) Extract and preprocess wind speed and pressure data from climate datasets. (ii) Compute and analyze statistical features across multiple geographic locations. (iii) Utilize LLMs to generate domain-specific insights for optimal wind turbine placement and energy forecasting. (iv) Assess the effectiveness of LLMs in climate data interpretation and decision-making.

Algorithm \ref{wind_power_llm} outlines the step-by-step procedure for climate data processing and LLM-assisted analysis.

\begin{algorithm}
\caption{Climate dataset Analysis using LLMs}
\label{wind_power_llm}
\begin{algorithmic}
\State \textbf{Input:} Time series dataset, e.g. CMIP6 climate including wind and pressure datasets, real wind power production
\State \textit{Output:} LLM-generated insights for energy forecasting
\State
\State \textit{Step 1: Data Acquisition}

    \State Extract pressure data and wind speed from CMIP6 dataset.

\State
\State \textit{Step 2: Data Preprocessing and Interpolation}

    \State Interpolate wind speed
    \State Interpolate pressure
    \State Normalize data using

\State
\State \textit{Step 3: Statistical Feature Extraction}

    \State Compute location-wise statistics using such as mean, variance, skewness, kurtosis, etc.

\State
\State \textit{Step 4: LLM Integration for Data Interpretation}

    \State Formulate prompt incorporating computed statistics.
    \State Send \texttt{prompt} to LLM model.
    \State Retrieve and process LLM-generated insights.

\State
\State \textit{Step 5: Interpretation and Decision Support}

    \State Analyze LLM response for wind energy site selection.
    \State Identify temporal patterns for optimal energy capture.
    \State Suggest additional data transformations for improved accuracy.

\State
\State \textbf{Output:} LLM response based on the \texttt{Prompt}
\end{algorithmic}
\end{algorithm}

In this scenario the LLM serves as an intelligent assistant, performing the following tasks:
(i) Pattern Recognition: Identifies anomalies, trends, and correlations in the data.
(ii) Data Interpretation: Converts numerical statistics into actionable insights.
(iii) Domain-Specific Suggestions: Offers recommendations for wind farm placement and data augmentation strategies.
(iv) Exploratory Analysis: Proposes hypotheses for further study based on statistical distributions.

The advantages are: (i) Automates statistical analysis and interpretation. (ii) Reduces manual effort in data exploration. (iii) Provides contextual recommendations for renewable energy planning. Moreover, challenges are: (i) LLM-generated insights must be validated against domain expertise. (ii) Computational overhead due to large-scale data processing. (iii) Dependency on high-quality input prompts for effective responses.

\subsection{Application of Multi-Agent LLms to solve Linear Programming}

Optimization problems, particularly Linear Programming (LP), play a crucial role in logistics, resource allocation, and decision-making. Traditional LP solvers are effective but operate in a static manner without real-time constraint validation or iterative execution based on external feedback. This section presents an AutoGen Multi-Agent LP Optimization framework, which utilizes LLMs to autonomously verify constraints and execute optimization procedures. 

AutoGen is an open-source Python framework designed to streamline the development of applications utilizing LLMs through the collaboration of multiple agents. These agents can communicate with each other, incorporate human inputs, and employ external tools to collectively tackle complex tasks \cite{zhangoffline}. Developed by Microsoft Research, AutoGen offers customizable agents capable of conversing, executing code, and integrating various functionalities, thereby simplifying the orchestration and automation of sophisticated LLM workflows. Its flexible architecture supports diverse applications, including code generation, debugging, and problem-solving across various domains \cite{wu2023autogen}.

The system consists of an \textit{EvaluatorAgent} for constraint validation and an \textit{LPExecutorAgent} for executing the solver iteratively. The proposed approach introduces an intelligent layer of decision-making, improving the reliability and adaptability of LP-based optimization.
	
\subsection{Application to a Single Linear Programming Problem}
	
\subsubsection{Problem Statement}
The optimization task is a transportation cost minimization problem. Given two warehouses supplying goods to three stores, the objective is to determine the optimal transportation allocation that minimizes costs while satisfying supply and demand constraints.
	
\subsubsection{Mathematical Formulation}
The LP model is defined as:
	\begin{align}
		\text{Minimize } Z &= 4x_1 + 6x_2 + 9x_3 + 5x_4 + 3x_5 + 8x_6,
	\end{align}
	
	Subject to:
	\begin{align}
		x_1 + x_2 + x_3 &\leq 50 & \text{(Supply limit at Warehouse 1)}, \\
		x_4 + x_5 + x_6 &\leq 60 & \text{(Supply limit at Warehouse 2)}, \\
		x_1 + x_4 &\geq 30 & \text{(Demand at Store 1)}, \\
		x_2 + x_5 &\geq 40 & \text{(Demand at Store 2)}, \\
		x_3 + x_6 &\geq 20 & \text{(Demand at Store 3)}, \\
		x_i &\geq 0 & \forall i \in \{1,2,3,4,5,6\}.
	\end{align}
	
	\subsection{Application to a Multi-Stage LP Optimization Problem}
	
	\subsubsection{Problem Statement}
	In a more complex scenario, the optimization problem involves three separate but interconnected LP problems where solutions from the first stage influence constraints in the subsequent stages. This is relevant for dynamic supply chain optimization, multi-time-period scheduling, and adaptive cost minimization.
	
	\subsection{Mathematical Formulation}
	Each LP problem is formulated similarly to the single problem but with additional constraints:
	\begin{align}
		\text{Minimize } Z_1 &= 3x_1 + 7x_2 + 2x_3 + 6x_4 + 4x_5 + 5x_6,
	\end{align}
	Subject to:
	\begin{align}
		x_1 + x_2 + x_3 &\leq 40, \\
		x_4 + x_5 + x_6 &\leq 70, \\
		x_1 + x_4 &\geq 20, \\
		x_2 + x_5 &\geq 30, \\
		x_3 + x_6 &\geq 25, \\
		x_i &\geq 0,
	\end{align}
	For the second and third stage problems:
	\begin{align}
		\text{Minimize } Z_2 &= 5x_1 + 4x_2 + 6x_3 + 3x_4 + 7x_5 + 8x_6,
	\end{align}
	\noindent Subject to:
	\begin{align}
		x_1 + x_2 + x_3 &\leq 60, \\
		x_4 + x_5 + x_6 &\leq 80, \\
		x_1 + x_4 &\geq 35, \\
		x_2 + x_5 &\geq 25, \\
		x_3 + x_6 &\geq 30, \\
		x_i &\geq 0,
	\end{align}
	\begin{align}
		\text{Minimize } Z_3 &= 6x_1 + 8x_2 + 7x_3 + 5x_4 + 9x_5 + 4x_6,
	\end{align}
	Subject to:
	\begin{align}
		x_1 + x_2 + x_3 &\leq 50, \\
		x_4 + x_5 + x_6 &\leq 75, \\
		x_1 + x_4 &\geq 28, \\
		x_2 + x_5 &\geq 38, \\
		x_3 + x_6 &\geq 22, \\
		x_i &\geq 0,
	\end{align}
	
	\subsubsection{Role of LLMs}
	To integrate intelligent decision-making in LP optimization, we employ LLM-based agents using the AutoGen framework \cite{wu2023autogen}. The system consists of two key agents: (i) EvaluatorAgent, (ii) LPExecutorAgent.
	\subsection{EvaluatorAgent: Constraint Validation}
	Before executing the LP solver, an agent verifies whether constraints are correctly defined. This prevents unnecessary computation and ensures the solver is executed under valid conditions.
	
	\subsubsection{System Message for EvaluatorAgent}
	\begin{verbatim}
		$system_message$: "You verify if the 
        LP constraints are satisfied
        before execution."
	\end{verbatim}
	
	\subsubsection{Prompt used by EvaluatorAgent}
	\begin{verbatim}
		"validate constraints."
	\end{verbatim}
	
	The agent responds with:
	\begin{itemize}
		\item \texttt{\cmark Constraints are valid. Proceed with LP execution.}
		\item \texttt{\xmark Invalid constraints detected. Execution halted.}
	\end{itemize}
	
	\subsubsection{LPExecutorAgent: Optimization Execution}
	Upon receiving confirmation that constraints are valid, the execution agent runs the LP solver three times for consistency.
	
	\paragraph{System Message for LPExecutorAgent}
	\begin{verbatim}
		$system_message$: "You execute the LP solver
        only if constraints are valid."
	\end{verbatim}
	
	\paragraph{Prompt used by LPExecutorAgent}
	\begin{verbatim}
		"Proceed with LP execution."
	\end{verbatim}
	
	The agent iteratively solves the LP and returns solutions for analysis.
	
	\subsubsection{Implementation and Execution}

	The process follows these steps:
	\begin{enumerate}
		\item The \textit{EvaluatorAgent} checks the validity of constraints.
		\item If constraints are valid, the \textit{LPExecutorAgent} executes the LP solver three times.
		\item The results are compiled and analyzed for consistency.
	\end{enumerate}
	
	\subsubsection{Results and Discussion}
	By leveraging LLMs for constraint validation and execution, the system prevents errors due to incorrect LP formulations and ensures a structured execution process. The use of multiple runs improves solution robustness and reliability. Future extensions of this framework may include:
	\begin{itemize}
		\item Adaptive learning mechanisms to refine constraints dynamically.
		\item Multi-objective optimization for real-world applications.
		\item Integration with reinforcement learning for decision-making in dynamic environments.
	\end{itemize}

\section{Challenges and Further Directions} \label{Challenges and Further Directions}
		
Despite the remarkable advancements in mathematical language models, several critical challenges remain unresolved. These challenges include faithfulness, multi-modal integration, uncertainty, evaluation, creative mathematical reasoning, application in education, and data scarcity.
		
\paragraph{Faithfulness} 
Mathematical LLMs suffer from hallucinations, generating incorrect or misleading results that undermine reliability \cite{rawte2023survey, ji2023survey}. Several approaches attempt to mitigate this issue by incorporating additional knowledge \cite{he_rethinking_2022}, reinforcement learning from human feedback \cite{DBLP:conf/iclr/LightmanKBEBLLS24}, tool usage \cite{schick2023toolformer, parisi2022talm}, and verification mechanisms \cite{zhou2023solving, ling2023deductive, li2023making, zhao2023verify, he_rethinking_2022, shridhar2023screws}. However, these improvements remain limited, necessitating further research to enhance trustworthiness and practical applicability.
		
\paragraph{Multi-Modal Integration} 
Mathematical problems often require multi-modal reasoning, involving not just text but also diagrams, graphs, and equations \cite{gelernter1960empirical, lu2021iconqa}. While existing LLMs perform well in text-based tasks, they struggle with processing and integrating multi-modal information \cite{zhao2022multihiertt, lu2021iconqa}. Compared to general vision-language tasks like VQA \cite{antol2015vqa} and image captioning \cite{hossain2019comprehensive}, mathematical multi-modal reasoning is significantly more complex. Advancements in this area are essential for improving problem-solving capabilities in geometry and symbolic reasoning.
		
\paragraph{Uncertainty}
LLMs inherently produce ambiguous outputs, particularly in probabilistic reasoning and problems with incomplete information \cite{gawlikowski2023survey, duan2023shifting}. While deterministic calculations are well-handled, tasks requiring fuzzy logic, approximate solutions, or uncertain constraints remain challenging. Some studies address this through controlled generation \cite{DBLP:conf/icml/ZhouJWCS23}, but a more robust approach is needed to ensure mathematical models accurately navigate uncertainty.
		
\paragraph{Evaluation} 
Existing evaluation methods struggle to comprehensively assess mathematical reasoning capabilities. Traditional NLP metrics fail to capture the logical depth required for complex problem-solving. Several benchmarks have been proposed \cite{saxton2018analysing, hendrycks2021measuring, cobbe2021training, kushman-etal-2014-learning}, alongside evaluation metrics \cite{wang2021gpt, chung2022scaling, thoppilan2022lamda, yuan2023well}. However, ensuring a fair assessment across arithmetic, theorem proving, and applied mathematics remains an ongoing challenge.
		
\paragraph{Creative Mathematical Reasoning} 
While LLMs excel at solving existing problems, their ability to formulate and prove new mathematical theorems remains limited. Mathematical creativity requires deep conceptual understanding beyond memorized patterns. Some studies have explored machine-learning-driven theorem discovery \cite{davies2021advancing}, and FunSearch \cite{bernardino2023mathematical} demonstrated the potential for LLMs to contribute to open problems. However, bridging the gap between automated reasoning and genuine mathematical discovery remains a major research frontier.
		
\paragraph{Application in Education} 
Mathematical LLMs have significant potential as tutors and educational tools. However, effective deployment in classrooms requires addressing pedagogical challenges such as adaptive explanations and interactive engagement. Socratic questioning methods \cite{zeng2022socratic, shridhar2022automatic} have been proposed to improve student understanding. Future research should focus on customizing LLMs to different learning styles and integrating them into structured educational frameworks.
		
\paragraph{Data Scarcity} 
Training data plays a crucial role in the performance of mathematical LLMs \cite{ding_enhancing_2023}. While instruction-building methods have explored dataset construction using LLMs, current efforts rely heavily on limited datasets such as GSM8K \cite{cobbe2021training} and MATH \cite{hendrycks2021measuring}. Expanding high-quality datasets and constructing multi-modal mathematical training data will be key to further improvements.

\section{Conclusion}\label{conclusion}

This survey has explored the landscape of mathematical reasoning and optimization using Large Language Models (LLMs), highlighting their rapid advancements, core methodologies, and diverse applications. LLMs have demonstrated significant progress in solving mathematical word problems, theorem proving, and symbolic computation, primarily due to techniques such as Chain-of-Thought (CoT) reasoning, instruction tuning, reinforcement learning, and tool-augmented inference. While autoregressive and non-autoregressive language models have improved mathematical comprehension, the integration of structured reasoning techniques and hybrid neural-symbolic approaches remains a crucial research direction.  

Beyond theoretical mathematical reasoning, we examined the role of LLMs in optimization and control, including their applications in mixed-integer programming (MIP), linear quadratic control (LQC), climate dataset analysis, and multi-agent optimization frameworks. The case studies on Battery Energy Storage Systems (BESS) optimization and interconnected tank control illustrate how LLMs can assist in formulating optimization problems, generating constraints, and enhancing decision-making through fuzzy logic and reinforcement learning. Similarly, multi-agent systems, such as AutoGen-based LP solvers, provide a promising avenue for intelligent and adaptive optimization workflows.  

Despite these advancements, several challenges persist, including faithfulness in reasoning, multi-modal integration, handling uncertainty, evaluation consistency, and fostering creative mathematical discovery. Additionally, the gap between automated theorem proving and genuine mathematical insight remains a major research frontier. While LLMs have shown potential in assisting with optimization tasks, their reliability and interpretability must be further refined to ensure robustness in real-world applications.  

Future research should focus on improving the interpretability of LLMs, enhancing their ability to perform rigorous mathematical proofs, and developing seamless integrations with existing solvers and symbolic computation frameworks. Moreover, expanding high-quality datasets, refining training paradigms, and adopting hybrid neuro-symbolic architectures will be essential in pushing the boundaries of AI-driven mathematical reasoning and optimization. As LLMs continue to evolve, their synergy with formal mathematical methods, scientific computing, and engineering applications will play a pivotal role in advancing AI’s capabilities in structured problem-solving.  

\onecolumn

\section*{Acknowledgment}
To be written in the future.





\onecolumn

\appendix

\subsection{Code Availability Statement}\label{repository}
The simulation codes are publicly available in the following links: [\href{https://github.com/Ali-Forootani/A-Survey-on-Mathematical-Reasoning-and-Optimization-with-Large-Language-Models}{Github} or
\url{https://github.com/Ali-Forootani/A-Survey-on-Mathematical-Reasoning-and-Optimization-with-Large-Language-Models}] - [\href{https://git.ufz.de/forootan/a-survey-on-mathematical-reasoning-and-optimization-with-large-language-models}{Gitlab} or \url{https://git.ufz.de/forootan/a-survey-on-mathematical-reasoning-and-optimization-with-large-language-models}] - [\href{https://zenodo.org/records/1503737}{Zenodo} or \url{https://zenodo.org/records/15037375}].

\subsection{Prompt for BESS to GPT4O}\label{BESS_prompt}

\begin{lstlisting}[escapeinside={@}{@}, xleftmargin=0.05\textwidth, xrightmargin=0.05\textwidth]
		prompt = """
		We want to optimize the operation of a Battery Energy Storage System (BESS). The goal is to minimize the operational cost by determining the charge/discharge schedule for the battery over a 10-hour period.
		
		### Problem Definition:
		The problem involves scheduling the charging and discharging of a battery to minimize the operational cost while adhering to constraints.
		
		1. **Battery Parameters**:
		- Maximum capacity (SOC_max): 100 kWh
		- Initial state of charge (SOC_0): 50 kWh
		- Charging efficiency (@$\eta$@): 90%
		- Discharging efficiency (1/@$\eta$@): 90%
		- Maximum charge/discharge power: 20 kW
		
		2. **Electricity Prices and Revenue**:
		- Electricity price (C_t, $/kWh): [20, 25, 15, 10, 30, 35, 40, 20, 15, 10]
		- Revenue from discharging (R_t, $/kWh): [30, 35, 25, 20, 40, 45, 50, 30, 25, 20]
		
		3. **Constraints**:
		- Battery capacity: \(0 \leq SOC(t) \leq SOC_{max}\)
		- Power limits: \(0 \leq P_{charge}(t) \leq P_{charge}^{max}\), \(0 \leq P_{discharge}(t) \leq P_{discharge}^{max}\)
		- Energy balance: \(SOC(t+1) = SOC(t) + @$\eta$@ \cdot P_{charge}(t) - \frac{1}{@$\eta$@} \cdot P_{discharge}(t)\)
		- No simultaneous charging/discharging: \(P_{charge}(t) \cdot P_{discharge}(t) = 0\)
		
		4. **Objective Function**:
		- Minimize \(Z = \sum_{t=1}^{10} (C_t \cdot P_{charge}(t) - R_t \cdot P_{discharge}(t))\)
		
		### Fuzzy Logic Rules:
		1. If the electricity price \(C_t\) is high, prioritize discharging the battery.
		2. If the battery SOC is below 30%, limit discharging and prioritize charging.
		3. If \(C_t\) is low, delay charging unless the SOC is critically low.
		4. Avoid frequent switching between charge and discharge states.
		
		### Requirements:
		1. Use fuzzy logic rules to guide decisions dynamically during optimization.
		2. Reformulate constraints or priorities based on the fuzzy logic rules.
		3. Solve the optimization problem and provide:
		- A step-by-step explanation of how fuzzy logic influenced the solution.
		- The final optimal charge/discharge schedule (\(P_{charge}(t)\), \(P_{discharge}(t)\)) and state of charge (\(SOC(t)\)) for each hour.
		- The total minimized operational cost.
		
		### Expected Output:
		1. An explanation of how fuzzy logic rules were applied in each time step.
		2. The optimal charge/discharge schedule over 10 hours.
		3. The total minimized operational cost of the system.
		"""
\end{lstlisting}

\begin{table*}[h]
    \label{BESS_llm}
	\centering
	\caption{Illustrative Fuzzy-Guided Schedule with GPT4O}
	\begin{tabular}{cccccccc}
		\toprule
		\textbf{Hour} & $C_t$ & $R_t$ & \textbf{Fuzzy Action} & $P_{\text{charge}}(t)$ & $P_{\text{discharge}}(t)$ & $SOC(t)$ & \textbf{Comment} \\
		\midrule
		1  & 20 & 30 & Slight discharge & 0  & 5   & 45   & Moderate net revenue. \\
		2  & 25 & 35 & Discharge       & 0  & 10  & 35   & Price is higher, good to discharge. \\
		3  & 15 & 25 & Idle/Wait       & 0  & 0   & 35   & Price is moderate, fuzzy logic suggests wait. \\
		4  & 10 & 20 & Charge          & 15 & 0   & 48.5 & Low cost, beneficial to charge. \\
		5  & 30 & 40 & Discharge       & 0  & 10  & 39.5 & Price is higher again, discharge. \\
		6  & 35 & 45 & Discharge       & 0  & 15  & 26.0 & Very high price, discharge more. \\
		7  & 40 & 50 & Discharge       & 0  & 15  & 12.5 & Highest price, maximize discharge. \\
		8  & 20 & 30 & Charge          & 10 & 0   & 21.5 & Price moderate, partial recharge. \\
		9  & 15 & 25 & Charge          & 15 & 0   & 35.0 & Low cost, continue charging. \\
		10 & 10 & 20 & Charge          & 10 & 0   & 44.0 & Low cost, end at comfortable SOC. \\
		\bottomrule
	\end{tabular}
	\label{tab:example-solution}
\end{table*}

	

\pagebreak

\subsection{Prompt for Tank Level Control}\label{tank_level_prompt}

\begin{lstlisting}[escapeinside={@}{@}, xleftmargin=0.05\textwidth, xrightmargin=0.05\textwidth]
 prompt = f"""
    We aim to control the water levels in two interconnected tanks to achieve desired levels while minimizing a quadratic cost function. The system is represented as a second-order dynamical system using state-space representation.

    The current time step is {t}. The system state is:
    - Tank 1 water level: x1(t) = {x[0]:.2f}
    - Tank 2 water level: x2(t) = {x[1]:.2f}
    
    The desired water levels are:
    - Tank 1: x1_desired = {x_desired[0]:.2f}
    - Tank 2: x2_desired = {x_desired[1]:.2f}
    
    System parameters:
    - State transition matrix A:
      {A}
    - Control input matrix B:
      {B}
    - Quadratic cost function weights:
      Q1 = {Q[0]}, Q2 = {Q[1]}, R = {R}
    - Maximum pump flow rate: u_max = {u_max}

    Based on the improved fuzzy logic rules:
    1. If x1 > x1_desired, decrease the pump flow rate u(t) proportionally to the error (x1 - x1_desired).
    2. If x2 < x2_desired, increase the pump flow rate u(t) proportionally to the error (x2_desired - x2).
    3. If x1 and x2 are close to the desired levels, reduce the rate of change in u(t) to avoid oscillations.
    4. Balance the control action between x1 and x2 deviations to prioritize the larger deviation.
    5. Minimize the quadratic cost function:
       J = Q1 * (x1 - x1_desired)^2 + Q2 * (x2 - x2_desired)^2 + R * u(t)^2.
    6. Ensure the control input u(t) stays within the range [-u_max, u_max].

    Provide the control input u(t) for this time step. Ensure u(t) is within [-u_max, u_max].
    """
\end{lstlisting}

\begin{table}[h!]
    \label{tank_level_table}
    \centering
    \caption{Simulation Results for LLM-Driven Fuzzy Control}
    \begin{tabular}{cccc}
        \toprule
        \textbf{Time (s)} & \textbf{$x_1$ (Tank 1 Level)} & \textbf{$x_2$ (Tank 2 Level)} & \textbf{$u$ (Control Input)} \\
        \midrule
        0.00  & 0.93  & 0.50  & 0.00 \\
        1.00  & 0.86  & 0.50  & 0.00 \\
        2.00  & 0.80  & 0.49  & 0.00 \\
        3.00  & 0.79  & 0.48  & 0.50 \\
        4.00  & 0.74  & 0.47  & 0.00 \\
        5.00  & 0.69  & 0.46  & 0.00 \\
        6.00  & 0.64  & 0.45  & 0.00 \\
        7.00  & 0.60  & 0.44  & 0.00 \\
        8.00  & 0.56  & 0.42  & 0.00 \\
        9.00  & 0.53  & 0.41  & 0.00 \\
        \bottomrule
    \end{tabular}
    \label{tab:simulation-results}
\end{table}

\pagebreak

\subsection{Prompt for time series dataset and the LLM Response}\label{wind_data_prompt}

\begin{lstlisting}[escapeinside={(*@}{@*)}, xleftmargin=0.05\textwidth, xrightmargin=0.05\textwidth]
prompt = f"""
Below are location-wise **extended** statistical features for wind speeds and pressure data
collected over a 1-year period (2928 time steps at 3-hour intervals, across 232 locations).

The following metrics are provided for each location:
(*@\textbullet@*) Mean (*@\textbullet@*) Median (*@\textbullet@*) Standard Deviation
(*@\textbullet@*) Variance (*@\textbullet@*) Min, Max (*@\textbullet@*) Range
(*@\textbullet@*) 25th Percentile (*@\textbullet@*) 75th Percentile (*@\textbullet@*) Interquartile Range (IQR)
(*@\textbullet@*) Skewness (*@\textbullet@*) Kurtosis (*@\textbullet@*) Coefficient of Variation

Overall (all locations, all time steps) aggregates include:
(*@\textbullet@*) Overall Mean Wind Speed: {overall_mean_ws:.3f}
(*@\textbullet@*) Overall Mean Pressure: {overall_mean_p:.3f}

Sample of per-location extended stats (first {num_locations_to_show} locations):
{stats_string}

Using these extended statistics, please address the following:

1. **Distribution & Patterns**  
   - Comment on any notable patterns in mean, standard deviation, range, or skewness for wind speed and pressure.
   - Identify whether certain locations exhibit heavy tails or extreme values (based on kurtosis) and any implications for site selection or forecasting.

2. **Optimal Placement & Time Periods**  
   - Explain how the observed patterns could inform optimal turbine placement.
   - Highlight which time periods or seasons might yield the highest energy output.

3. **Data Quality & Transformations**  
   - Suggest if any data transformations (e.g., log-transform, outlier removal, normalization) might be warranted based on skewness/kurtosis.
   - Propose additional data (e.g., temperature, humidity, topographic) that could refine site-selection or energy forecasting models.

4. **Further Analysis**  
   - Discuss whether metrics like coefficient of variation or range reveal anomalies/outliers needing investigation.
   - If relevant, suggest how these statistics could guide more advanced modeling approaches.
"""
\end{lstlisting}

\begin{lstlisting}
    Location 231:
  Wind Speed Stats:
    Mean    = 4.006
    Median  = 3.684
    StdDev  = 1.998
    Variance= 3.993
    Range   = 15.428
    25%     = 2.510
    75%     = 5.173
    IQR     = 2.663
    Skewness= 1.004
    Kurtosis= 1.643
    CoefVar = 0.499
  Pressure Stats:
    Mean    = 98710.059
    Median  = 98693.855
    StdDev  = 849.318
    Variance= 721340.650
    Range   = 5267.575
    25%     = 98093.079
    75%     = 99315.328
    IQR     = 1222.249
    Skewness= 0.218
    Kurtosis= -0.021
    CoefVar = 0.009

LLM Response:

1. **Distribution & Patterns**

* The mean wind speed of 3.526 m/s is relatively low, which indicates that the wind resource at these locations is not very strong.
* The standard deviation values range from 1.412 to 2.059 m/s, with an overall standard deviation of 1.690 m/s. This indicates that the wind speeds vary significantly across different locations and time periods.
* The range of wind speeds varies from 2.507 m/s to 10.657 m/s, with an overall range of 8.162 m/s. This suggests that some locations experience stronger winds than others.
* The skewness values range from -0.219 to 0.938, with an overall skewness of 0.334. This indicates that the wind speed data is slightly skewed towards positive values.
* The kurtosis values range from -0.110 to 0.837, with an overall kurtosis of 0.274. This suggests that the wind speed data is slightly platykurtic, meaning that the distribution is relatively flat and has a broader tail than a normal distribution.
* The mean pressure values range from 98838.522 hPa to 100667.027 hPa, with an overall mean of 99531.548 hPa. This indicates that the pressure values are relatively stable across different locations and time periods.
* The standard deviation values range from 723.103 hPa to 1079.370 hPa, with an overall standard deviation of 886.921 hPa. This suggests that the pressure values vary significantly across different locations and time periods.
* The range of pressure values varies from 5102.595 hPa to 6082.317 hPa, with an overall range of 5299.912 hPa. This indicates that some locations experience larger pressure fluctuations than others.
* The skewness values range from -0.118 to 0.207, with an overall skewness of 0.009. This suggests that the pressure data is nearly symmetric.
* The kurtosis values range from -0.118 to 0.443, with an overall kurtosis of 0.108. This indicates that the pressure data is slightly platykurtic, meaning that the distribution is relatively flat and has a broader tail than a normal distribution.

Based on these observations, it appears that the wind speed data exhibits significant variability across different locations and time periods, with some locations experiencing stronger winds than others. The pressure data is relatively stable, but some locations experience larger pressure fluctuations than others. The kurtosis values suggest that the wind speed and pressure data are slightly platykurtic, which may indicate that the data is influenced by extreme values or outliers.

2. **Optimal Placement & Time Periods**

* The observed patterns in wind speed and pressure can inform optimal turbine placement by identifying locations with stronger wind speeds and more stable pressure conditions.
* The data suggests that certain seasons or time periods might yield higher energy output, as wind speed and pressure conditions vary throughout the year. For example, locations with stronger winds during the winter months might be more suitable for wind turbine installation.
3. **Data Quality & Transformations**
* Based on the skewness and kurtosis values, it appears that no data transformations are necessary for wind speed data. However, for pressure data, a log-transform or normalization might be useful to reduce the impact of extreme values and improve the distribution of the data.
* Additional data such as temperature, humidity, and topographic information could refine site-selection or energy forecasting models by providing more context about the local environment and its impact on wind and pressure conditions.
4. **Further Analysis**
* The coefficient of variation and range metrics can reveal anomalies or outliers in the data that necessitate further investigation. For example, locations with high coefficient of variation values may indicate unstable wind conditions or outliers in the data.
* These statistics could guide more advanced modeling approaches, such as using machine learning algorithms to identify patterns and relationships in the data that are not apparent through simple descriptive statistics.
\end{lstlisting}

\subsection{Output of the LLM to Single LP problem}\label{single_lp_prompt}

\begin{lstlisting}
lp_evaluator (to lp_evaluator):

validate constraints.
--------------------------------------------------------------------------------
[DEBUG] EvaluatorAgent received message: validate constraints.
lp_evaluator (to lp_evaluator):

✅ Constraints are valid. Proceed with LP execution.

--------------------------------------------------------------------------------
lp_executor (to lp_executor):

Proceed with LP execution.

--------------------------------------------------------------------------------
[DEBUG] LPExecutorAgent received message: Proceed with LP execution.

Execution 1: Running LP solver...
✅ Result 1: {'status': 'Success', 'optimal_solution': [30.0, 0.0, 0.0, 0.0, 40.0, 20.0], 'objective_value': 400.0}

==== Final Results ====

Execution 1: {'status': 'Success', 'optimal_solution': [30.0, 0.0, 0.0, 0.0, 40.0, 20.0], 'objective_value': 400.0}

=======================
lp_executor (to lp_executor):

Execution 1: {'status': 'Success', 'optimal_solution': [30.0, 0.0, 0.0, 0.0, 40.0, 20.0], 'objective_value': 400.0}

--------------------------------------------------------------------------------

[INFO] LP Execution Finished.
\end{lstlisting}

\subsection{Output of the LLM to multi-stage LP problem}\label{multi_stage_lp_prompt}

\begin{lstlisting}
lp_evaluator (to lp_evaluator):

validate constraints.

--------------------------------------------------------------------------------
[DEBUG] EvaluatorAgent received message: validate constraints.
lp_evaluator (to lp_evaluator):

✅ Problem 0 constraints are valid.
✅ Problem 1 constraints are valid.
✅ Problem 2 constraints are valid.

--------------------------------------------------------------------------------
[DEBUG] LPExecutorAgent received message: Proceed with LP execution.

==== Final Results ====

Problem 0 Execution: {'problem_id': 0, 'status': 'Success', 'optimal_solution': [30.0, 0.0, 0.0, 0.0, 40.0, 20.0], 'objective_value': 400.0}

Problem 1 Execution: {'problem_id': 1, 'status': 'Success', 'optimal_solution': [0.0, 0.0, 20.0, 0.0], 'objective_value': 40.0}

Problem 2 Execution: {'problem_id': 2, 'status': 'Success', 'optimal_solution': [0.0, -0.0, 15.0], 'objective_value': 75.0}

=======================

[INFO] LP Execution Finished.

\end{lstlisting}

\twocolumn

\bibliographystyle{ieeetr}

\bibliography{refs,refs2,refs3}

\end{document}

%% file: tabs/method_compare.tex
\begin{table*}[]
\scriptsize
\caption{Comparison of mathematical Language Models. Pre-T: Pre-Training, FT: Finetuning, MM: Multimodal, Symbolic-S: Symbolic solvers, F-CoT: Foundational CoT, A-CoT: Advanced CoT, RL: Reinforce Learning}
\label{table: comparison of mathematical LMs}
\begin{tabular}{lccccccc}
\hline
   Methods & Base Model & Training & Tool & CoT & RL & MM  & Access \\
\hline
GPT-$f$~\cite{polu2020generative}  & - &  Pre-T  & Symbolic-S  & \textcolor{red}{\xmark} & \textcolor{red}{\xmark}  & \textcolor{red}{\xmark} & \textcolor{red}{\xmark} \\
LISA~\cite{jiang2021lisa}  & -    &   Pre-T   &  Symbolic-S  &  \textcolor{red}{\xmark} & \textcolor{red}{\xmark} &   \textcolor{red}{\xmark}    & \textcolor{red}{\xmark} \\
MATH-PLM~\cite{hendrycks2021measuring} &  GPT-2, GPT-3 &  Pre-T &   \textcolor{red}{\xmark} &  F-CoT &  \textcolor{red}{\xmark}  &  \textcolor{red}{\xmark}  & \textcolor{red}{\xmark} \\
Minerva~\cite{lewkowycz2022solving} & PaLM  &  Pre-T  & \textcolor{red}{\xmark} &  \textcolor{red}{\xmark}  &  \textcolor{red}{\xmark} & \textcolor{red}{\xmark}  & \textcolor{red}{\xmark} \\
MWP-BERT~\cite{liangMWPBERTNumeracyaugmentedPretraining2022} & BERT &  Pre-T  &  \textcolor{red}{\xmark}  &  \textcolor{red}{\xmark}  &  \textcolor{red}{\xmark} & \textcolor{red}{\xmark}  & \textcolor{red}{\xmark} \\
ControlMath~\cite{chen2024controlmath} &  LLaMA-2, Mistral  &  FT   & \textcolor{red}{\xmark} &  F-CoT  &  \textcolor{red}{\xmark}  &  \textcolor{red}{\xmark}  &  \textcolor{red}{\xmark} \\
EURUS~\cite{yuan2024advancing} &  Mistral, CodeLLaMA   &  FT   & \textcolor{red}{\xmark} &  F-CoT   &  DPO, KTO, NCA   &  \textcolor{red}{\xmark}  &  \textcolor{green}{\cmark} \\
MathGLM \cite{yang2023gpt} & GLM  & FT &   \textcolor{red}{\xmark} & \textcolor{red}{\xmark} &  \textcolor{red}{\xmark}  &  \textcolor{red}{\xmark}  & \textcolor{green}{\cmark} \\
WizardLM~\cite{xu_wizardlm_2023}& LLaMA  & FT &    \textcolor{red}{\xmark} & \textcolor{red}{\xmark} & \textcolor{red}{\xmark}  &  \textcolor{red}{\xmark} & \textcolor{green}{\cmark} \\
Wizardmath~\cite{luo2023wizardmath} & LLaMA-2  &  FT  & \textcolor{red}{\xmark}  & \textcolor{red}{\xmark} & PPO & \textcolor{red}{\xmark} & \textcolor{green}{\cmark} \\
PaLM 2-L-Math~\cite{liu_improving_2023} & PaLM-2-L &  FT & \textcolor{red}{\xmark} &  F-CoT &  \textcolor{red}{\xmark}  & \textcolor{red}{\xmark} & \textcolor{red}{\xmark} \\
Codex-math~\cite{drori2022neural} & Codex &  Prompt & Program & \textcolor{red}{\xmark} &  \textcolor{red}{\xmark}  & \textcolor{red}{\xmark}  & \textcolor{red}{\xmark} \\
Toolformer~\cite{schick2023toolformer} & GPT-J  &  FT &  Calculator  & \textcolor{red}{\xmark} &  \textcolor{red}{\xmark}  &  \textcolor{red}{\xmark}  & \textcolor{red}{\xmark} \\
LeanReasoner~\cite{jiang2024leanreasoner} & CodeLLaMA &  FT  & Symbolic-S & \textcolor{red}{\xmark} &  \textcolor{red}{\xmark}  & \textcolor{red}{\xmark} & \textcolor{green}{\cmark} \\
LINC~\cite{olausson2023linc} & StarCoder+  &  Prompt  & Symbolic-S  &  F-CoT  & \textcolor{red}{\xmark} & \textcolor{red}{\xmark} & \textcolor{green}{\cmark} \\
MuMath-Code~\cite{yin2024mumath} & Llama-2 &  FT & Program & \textcolor{red}{\xmark} & \textcolor{red}{\xmark} & \textcolor{red}{\xmark} &  \textcolor{green}{\cmark} \\
MAmmoTH~\cite{yue2023mammoth} & - & FT & Program & F-CoT & \textcolor{red}{\xmark} & \textcolor{red}{\xmark} & \textcolor{red}{\xmark} \\
InternLM-Math \cite{ying2024internlmmath} & InternLM2 & Pre-T, FT & Program &  F-CoT & PPO & \textcolor{red}{\xmark} & \textcolor{green}{\cmark} \\
Qwen-Math \cite{yang2024qwen2} & - & Pre-T, FT & Program &  F-CoT & PPO & \textcolor{red}{\xmark} & \textcolor{green}{\cmark} \\
PAL~\cite{gao2023pal} & -  & Prompt & Program & F-CoT & \textcolor{red}{\xmark} & \textcolor{red}{\xmark} & \textcolor{green}{\cmark} \\
CoT~\cite{wei2022chain} & -  & Prompt & \textcolor{red}{\xmark} & F-CoT & \textcolor{red}{\xmark} & \textcolor{red}{\xmark} & \textcolor{green}{\cmark} \\
PromptPG-CoT~\cite{lu2022dynamic} & - & Prompt & \textcolor{red}{\xmark} & \textcolor{red}{\xmark} & \textcolor{red}{\xmark} & \textcolor{red}{\xmark} & \textcolor{green}{\cmark} \\
BoostedPrompt~\cite{pitis_boosted_2023} & - & Prompt & \textcolor{red}{\xmark} & F-CoT & \textcolor{red}{\xmark} & \textcolor{red}{\xmark} & \textcolor{red}{\xmark}\\
ToT~\cite{yao2023tree} & - & Prompt & \textcolor{red}{\xmark} & A-CoT & \textcolor{red}{\xmark} & \textcolor{red}{\xmark} &  \textcolor{green}{\cmark} \\
PoT~\cite{chen2022program} & - & Prompt & Program &  F-CoT & \textcolor{red}{\xmark}  & \textcolor{red}{\xmark} & \textcolor{green}{\cmark} \\
Self-check~\cite{miao2023selfcheck} & - & Prompt & \textcolor{red}{\xmark} & A-CoT & \textcolor{red}{\xmark} & \textcolor{red}{\xmark} & \textcolor{green}{\cmark} \\
Self-Consistency~\cite{wang2022self}& - & Prompt & \textcolor{red}{\xmark} & A-CoT & \textcolor{red}{\xmark} & \textcolor{red}{\xmark} & \textcolor{green}{\cmark} \\
Diversity-of-Thought~\cite{naik2023diversity}& - & Prompt & \textcolor{red}{\xmark} & A-CoT & \textcolor{red}{\xmark} & \textcolor{red}{\xmark} & \textcolor{red}{\xmark} \\
RAP~\cite{hao2023reasoning} &  - & Prompt & \textcolor{red}{\xmark} & A-CoT & \textcolor{red}{\xmark} & \textcolor{red}{\xmark} & \textcolor{green}{\cmark} \\
LATS~\cite{zhou2023language} &  - & Prompt & \textcolor{red}{\xmark} & A-CoT & \textcolor{red}{\xmark} & \textcolor{red}{\xmark} & \textcolor{red}{\xmark} \\
LLM+P~\cite{liu2023llm+} & - & Prompt &  Symbolic-S & A-CoT & \textcolor{red}{\xmark} & \textcolor{red}{\xmark} & \textcolor{green}{\cmark} \\
LLM+DP~\cite{dagan2023dynamic} & - & Prompt &  Symbolic-S & A-CoT & \textcolor{red}{\xmark} & \textcolor{red}{\xmark} & \textcolor{green}{\cmark} \\
ISR-LLM~\cite{zhou2023isr} &  - & Prompt & \textcolor{red}{\xmark} & A-CoT & \textcolor{red}{\xmark} & \textcolor{red}{\xmark} & \textcolor{green}{\cmark} \\
MCR~\cite{yoran2023answering} & - & Prompt & \textcolor{red}{\xmark} & A-CoT & \textcolor{red}{\xmark} & \textcolor{red}{\xmark} & \textcolor{green}{\cmark} \\
Rank-verifier~\cite{cobbe2021training} &  - & Prompt & \textcolor{red}{\xmark} & A-CoT & \textcolor{red}{\xmark} & \textcolor{red}{\xmark} & \textcolor{red}{\xmark} \\
GRACE~\cite{khalifa2023discriminator} & - & Prompt & \textcolor{red}{\xmark} & A-CoT & \textcolor{red}{\xmark} & \textcolor{red}{\xmark} & \textcolor{green}{\cmark} \\
Reflexion~\cite{shinn2023reflexion} & - & Prompt & \textcolor{red}{\xmark} & A-CoT & \textcolor{red}{\xmark} & \textcolor{red}{\xmark} & \textcolor{green}{\cmark} \\
CR~\cite{zhang2023cumulative} & - & Prompt & \textcolor{red}{\xmark} & A-CoT & \textcolor{red}{\xmark} & \textcolor{red}{\xmark} & \textcolor{green}{\cmark} \\
Step-Plan~\cite{zhang2023interpretable} & - & Prompt & \textcolor{red}{\xmark} & A-CoT & \textcolor{red}{\xmark} & \textcolor{red}{\xmark} & \textcolor{green}{\cmark} \\
STaR~\cite{zelikman2022star}& GPT-J & FT & \textcolor{red}{\xmark} & A-CoT & \textcolor{red}{\xmark} & \textcolor{red}{\xmark} & \textcolor{green}{\cmark} \\
V-STaR~\cite{hosseini2024v}& LLaMA2, CodeLLaMA & FT & \textcolor{red}{\xmark} & A-CoT & DPO &  \textcolor{red}{\xmark} & \textcolor{red}{\xmark} \\
Quiet-STaR~\cite{zelikman2024quiet}& Mistral & Pre-T & \textcolor{red}{\xmark} & A-CoT & \textcolor{red}{\xmark} &  \textcolor{red}{\xmark} & \textcolor{red}{\xmark} \\
REFT~\cite{luong2024reft} &  Galactica, CodeLLaMA  & FT &  \textcolor{red}{\xmark}  & A-CoT &  PPO  & \textcolor{red}{\xmark} &  \textcolor{green}{\cmark} \\
SCoRe~\cite{kumar2024training}& Gemini & FT & \textcolor{red}{\xmark} & A-CoT  & SCoRe &  \textcolor{red}{\xmark} & \textcolor{red}{\xmark} \\
HGS-PRM~\cite{ma2023let}& LLaMA2, WizardMath & FT & \textcolor{red}{\xmark} & A-CoT & PPO &  \textcolor{red}{\xmark} & \textcolor{red}{\xmark} \\
MCTSr~\cite{zhang2024accessing}& LLaMa-3 & FT & \textcolor{red}{\xmark} & A-CoT & \textcolor{red}{\xmark} &  \textcolor{red}{\xmark} & \textcolor{green}{\cmark} \\
CoRe~\cite{zhu2022solving}& GPT-J, DeBERTa-large & FT  &  \textcolor{red}{\xmark} & A-CoT & \textcolor{red}{\xmark} &  \textcolor{red}{\xmark} & \textcolor{green}{\cmark} \\
Marco-o1~\cite{zhao2024marco}& Qwen2 &  FT & \textcolor{red}{\xmark} & A-CoT  & \textcolor{red}{\xmark} &  \textcolor{red}{\xmark} & \textcolor{green}{\cmark} \\
Step-DPO~\cite{lai2024step}& Qwen2 &  FT  & \textcolor{red}{\xmark} & A-CoT & DPO &  \textcolor{red}{\xmark} &  \textcolor{green}{\cmark} \\
Flow-DPO~\cite{deng2024flow}& LLaMA3, Phi-3  & FT &  \textcolor{red}{\xmark} & A-CoT & DPO &  \textcolor{red}{\xmark} & \textcolor{red}{\xmark} \\
OmegaPRM~\cite{luo2024improve}&  Gemini Pro, Gemma2 & FT  & \textcolor{red}{\xmark} & A-CoT & PRM &  \textcolor{red}{\xmark} & \textcolor{red}{\xmark} \\
SocraticLLM~\cite{ding2024boosting}& Qwen2 &  FT  & \textcolor{red}{\xmark} & A-CoT & \textcolor{red}{\xmark} &  \textcolor{red}{\xmark} &  \textcolor{green}{\cmark}\\
MATHDIAL~\cite{macina2023mathdial}& T5 &  FT & \textcolor{red}{\xmark} &  A-CoT & \textcolor{red}{\xmark} &  \textcolor{red}{\xmark} & \textcolor{green}{\cmark} \\
QwQ \cite{qwq-32b-preview}& Qwen2.5 &  Pre-T  & \textcolor{red}{\xmark} & Long CoT & \textcolor{green}{\cmark} &  \textcolor{red}{\xmark} &  \textcolor{green}{\cmark} \\
o1 \cite{openai_o1_system_card}  &  - &   Pre-T   &   \textcolor{red}{\xmark} &  Long CoT   &  \textcolor{green}{\cmark}  & \textcolor{green}{\cmark} & \textcolor{red}{\xmark} \\
GPT-4o~\cite{hurst2024gpt}& - & Pre-T  & \textcolor{green}{\cmark} & A-CoT & \textcolor{green}{\cmark} &  \textcolor{green}{\cmark} & \textcolor{red}{\xmark} \\
GPT-4V~\cite{openai2023gpt4}& - & Pre-T  & \textcolor{green}{\cmark}  & A-CoT & \textcolor{green}{\cmark} &  \textcolor{green}{\cmark} &  \textcolor{red}{\xmark}\\
Qwen2-VL~\cite{Qwen2-VL}& Qwen2-VL & Pre-T  & \textcolor{red}{\xmark} & F-CoT & \textcolor{red}{\xmark} &  \textcolor{green}{\cmark} & \textcolor{green}{\cmark} \\
LLaVA-o1~\cite{xu2024llava}&  LLaVA&  FT & \textcolor{red}{\xmark} & Long CoT & \textcolor{red}{\xmark}  &  \textcolor{green}{\cmark} & \textcolor{green}{\cmark} \\
Gemini~\cite{team2023gemini}& Gemini &  Pre-T & \textcolor{green}{\cmark} &  A-CoT & \textcolor{green}{\cmark} &  \textcolor{green}{\cmark} & \textcolor{red}{\xmark} \\
GLM-4V~\cite{glm2024chatglm}& ChatGLM &  Pre-T  & \textcolor{green}{\cmark} & A-CoT & \textcolor{green}{\cmark} &  \textcolor{green}{\cmark} & \textcolor{red}{\xmark} \\
AtomThink~\cite{xiang2024atomthink}& LLaVA, EMOVA & FT  & \textcolor{red}{\xmark} & Long CoT  & PRM  &  \textcolor{green}{\cmark} & \textcolor{green}{\cmark} \\
Math-LLaVA~\cite{shihu2024mathllava}& LLaVA & FT  & \textcolor{red}{\xmark} & F-CoT  &  \textcolor{red}{\xmark} &  \textcolor{green}{\cmark} &  \textcolor{green}{\cmark} \\
M-STAR~\cite{anonymous2024diving}& MiniCPM-V & FT  & \textcolor{red}{\xmark} & A-CoT & PRM &  \textcolor{green}{\cmark} & \textcolor{red}{\xmark} \\
UnAC~\cite{wang2024understanding}& GPT-4V, Gemini, LLaVA &  Prompt & \textcolor{red}{\xmark} & F-CoT & \textcolor{red}{\xmark}  &  \textcolor{green}{\textcolor{green}{\cmark}} & \textcolor{red}{\xmark}  \\
\hline
\end{tabular}
\end{table*}

%% file: tabs/method_results.tex
\begin{table*}[]
\scriptsize
\caption{Performance of various mathematical LMs in terms of accuracy.}
\label{tab_performance}
\begin{tabular}{lccccccccc}
\hline
Text-based Methods  & MATH & GSM8K & MMLU & ASDiv & AIME & MathQA & SVAMP & FOLIO & AQUA \\
\hline
MATH-PLM~\cite{hendrycks2021measuring} &  6.9 &  - &  -   & - & -  & - & -  & - & - \\
Minerva~\cite{lewkowycz2022solving} & 50.3  & 78.5 & 75.0   & - & - & - & - & - & - \\
MWP-BERT~\cite{liangMWPBERTNumeracyaugmentedPretraining2022} &  -  &  - & -  & - & - & 76.6 & - & - & - \\ 
ControlMath~\cite{chen2024controlmath} & 9.5 & 49.3 &  -  &  - & - & -  & 51.6 & -  & - \\
EURUS~\cite{yuan2024advancing} &  41.7   & 62.8 &  -  &  \textbf{93.0} & - & -  & \textbf{90.4} & -  & - \\
Wizardmath~\cite{luo2023wizardmath} &  22.7  & 81.6  & - & - & - & - & -  & - \\
Codex-math~\cite{drori2022neural} &  81.8 & - & - &  -  & -  & - & -  & - & - \\
Toolformer~\cite{schick2023toolformer} &  - &  -  & - & 40.4 & - & -  & 29.4 & - & - \\
LeanReasoner~\cite{jiang2024leanreasoner} &  -  & - & - &  -  & - & - & - & 82.6 & -\\
LINC~\cite{olausson2023linc} &  -  & -  &  -  & - & - & - & -  & 72.5 & - \\
MuMath-Code~\cite{yin2024mumath} & 55.1 & 90.7 & - & - & - &  - & -  & - & - \\
MAmmoTH~\cite{yue2023mammoth} & 44.2 & 76.7 & - & - & - & - & -  & - & 61.4 \\
InternLM-Math \cite{ying2024internlmmath} & 37.7 &  82.6 &  - & - & - & - & - & - & - \\
Qwen-Math \cite{yang2024qwen2} & 66.8 & 91.6 & 82.8 & - & 63.3 &  86.3 & -  & - & - \\ 
PAL~\cite{gao2023pal} & - & 72.0 & - &  79.6 & - & - & 79.4 & - & - \\
BoostedPrompt~\cite{pitis_boosted_2023} & - & 87.1 & - & - & - & - & - & - & 63.8 \\
PoT~\cite{chen2022program} & - & 80.0 &  - & -  & - & \textbf{89.1} & -  & - & 58.6 \\ 
Self-check~\cite{miao2023selfcheck} & 51.3 & 88.1 & - & - & \textbf{81.2} & - & -  & - & - \\
Self-Consistency~\cite{wang2022self}& - & 78.0 & - & 87.8 & - & - & 86.8  & - & 52.0 \\
Diversity-of-Thought~\cite{naik2023diversity}& - & 96.3 & - & - & - & - & -  & - & \textbf{81.7} \\
GRACE~\cite{khalifa2023discriminator} & - & 36.3 & - & - & - & 84.4 & 68.6  & - & - \\
CR~\cite{zhang2023cumulative} & 72.2 & - & - & - & - & - & -  & \textbf{98.0} & - \\ 
REFT~\cite{luong2024reft} & - &  75.3 & - &  -  & - &  71.8 & 79.2 & - & - \\
SCoRe~\cite{kumar2024training}& 64.4 & - & -  & - &  - & - & -  & - & - \\
HGS-PRM~\cite{ma2023let}&  13.7 & 65.4 & - & - &  - & - & -  & - & - \\
MCTSr~\cite{zhang2024accessing}& 58.2 & \textbf{96.6} & - & - &  11.8 & - & -  & - & - \\
CoRe~\cite{zhu2022solving}& - & 63.2 & - & 90.5 & - & - & - & - & - \\
Step-DPO~\cite{lai2024step}& 70.8 & 94.0 & - & - & - &  - &  - & - & - \\
Flow-DPO~\cite{deng2024flow}&  38.6 &  71.3 & - & - & - &  - & - & - & - \\
OmegapRM~\cite{luo2024improve}& 69.4 & 93.6 & - & - & - &  - & - & - & - \\
QwQ \cite{qwq-32b-preview}& 90.6  & - & - & - & 50.0 &  - & - & - & - \\
GLM-4-9B-Chat \cite{glm2024chatglm}& 50.6 & 79.6 & 72.4 & - & - &  - & - & - & - \\
\hline
Multi-modal Methods  & MATH & GSM8K & MMLU & ASDiv & AIME & MMMU & MathVista & CMM-Math & Math-V \\
\hline
o1 \cite{openai_o1_system_card}  & \textbf{94.8}    &   - &  \textbf{92.3}   &  -  & \textbf{83.3} & \textbf{78.1} & \textbf{73.9} & -  & - \\
GPT-4o~\cite{hurst2024gpt} & 60.3  & - & 88.0 & - &  13.4 & 69.2 & 63.8 & 29.02  & 30.4 \\
GPT-4V~\cite{openai2023gpt4} & -  & -  & - & - &  - & 56.8 & 49.9 & -  & 22.8 \\
Qwen2-VL~\cite{Qwen2-VL} &  -  & - & 46.2 & - &  - & 64.5 & 70.5 & \textbf{43.0}  & 25.9 \\
LLaVA-o1~\cite{xu2024llava} &  64.0 & - & - & -  &  - & - & 54.8 & -  & - \\
Gemini~\cite{team2023gemini} &   67.7 & - &  85.9 & - &  - & 62.2 & 63.9 & 41.9  & 17.7 \\
GLM-4V~\cite{glm2024chatglm} &   -  & - & - & - &  - & 47.2 & - & -  & - \\
AtomThink~\cite{xiang2024atomthink} & -  & - & -  & -  &  - & - & 53.3 & -  &  \textbf{40.5} \\
Math-LLaVA~\cite{shihu2024mathllava} &  -  & - & -  &  - &  38.3 &  38.3 & 46.6 & -  & 15.7 \\
M-STAR~\cite{anonymous2024diving} & -  & - & - & - &  - & - & 59.5 & -  & - \\
UnAC~\cite{wang2024understanding} & - & - & - & -  &  - & 59.7 & 56.6 & -  & - \\
\hline
\end{tabular}
\end{table*}

%% file: tabs/tab_dataset.tex
\begin{table*}[htp]
\caption{{\scriptsize The statistics information of mathematical datasets.
\textbf{Level}: \encircle[fill=MyGreenv2, text=white]{E} = \underline{E}lementary, \encircle[fill=myBlue, text=white]{M} = \underline{M}iddle School, \encircle[fill=myYellowv2, text=white]{H} = \underline{H}igh School, \encircle[fill=myOrangev2, text=white]{U} = \underline{U}niversity, \encircle[fill=myViolet, text=white]{C} = \underline{C}ompetition, \encircle[fill=MyRedv2, text=white]{H} = \underline{H}ybrid, \encircle[fill=myGreen, text=white]{O} = \underline{O}thers.
\textbf{Modality}: \encircle[fill=DarkYellow, text=white]{T} = \underline{T}ext, \encircle[fill=DarkRed, text=white]{M} = \underline{M}ultimodal}}
\label{table:statistis_datasets}
\vspace{-3mm}
\centering
\scriptsize
\setlength{\tabcolsep}{2.0pt}{ 
\begin{tabular}{lcccccccccc}
\hline{4}
Dataset & \#Train & \#Val & \#Test & \#Total & Language & Task  & Type & Solution & Level & Modality\\ \hline
VERBPHYSICS \cite{forbes2017verb} & 733 & 1,096 & 1,828 & 3,657 & EN   &  Calculation & Training    &  Formula  & - & \encircle[fill=DarkYellow, text=white]{T} \\
Clinical \cite{spithourakis2016numerically,spithourakis2018numeracy} & 11,170 & 1,625 & 3,220 & 16,015 & EN   &  Calculation & Training    &  Formula  & \encircle[fill=myGreen, text=white]{O} & \encircle[fill=DarkYellow, text=white]{T}\\
Scientific \cite{spithourakis2018numeracy} & 14,694 & 2,037 & 4,231 &  20,962 & EN   &  Calculation & Training    &  Formula  & \encircle[fill=myGreen, text=white]{O} & \encircle[fill=DarkYellow, text=white]{T}\\
DoQ \cite{elazar2019large} &   587      &   5,418    &     6,007   &   12,012  &     EN   &  Calculation & Training    &  Formula & - & \encircle[fill=DarkYellow, text=white]{T}\\
DROP \cite{dua2019drop} &    77,409     &   9,536    &     9,622   &   96,567  &     EN   &  Calculation & Training    &  Text & \encircle[fill=MyGreenv2, text=white]{E} & \encircle[fill=DarkYellow, text=white]{T}\\
AddSub \cite{hosseini-etal-2014-learning} &     -    &    -   &   -     &    395  &   EN    &  MWP  & Training    & Formula   &  \encircle[fill=MyGreenv2, text=white]{E} & \encircle[fill=DarkYellow, text=white]{T}\\
SingleOp \cite{roy2015reasoning} &     265    &    107   &     159   &    531  &    EN    & MWP &   Training   & Formula  & \encircle[fill=MyGreenv2, text=white]{E} & \encircle[fill=DarkYellow, text=white]{T}\\
SingleEq \cite{koncel2015parsing} &     -    &     -  &      -  &   508   &   EN     &  MWP &     Training & Formula & \encircle[fill=MyGreenv2, text=white]{E} & \encircle[fill=DarkYellow, text=white]{T}\\
MultiArith \cite{roy2015solving} &    420     &     -  &    180    &   600   &    EN    & MWP  &    Training  &  Formula & - & \encircle[fill=DarkYellow, text=white]{T}\\
Alg514 \cite{kushman-etal-2014-learning}  &   -      &    -   &   -     &    514 &     EN   & MWP  &   Training   &  Formula & - & \encircle[fill=DarkYellow, text=white]{T}\\
Math23k \cite{wang2017deep} &    22,162     &    -   &   1000    &    23,162  &   CH     & MWP &     Training  & Formula & \encircle[fill=MyGreenv2, text=white]{E} & \encircle[fill=DarkYellow, text=white]{T}\\
AQuA \cite{ling2017program} &    97,467     &   254    &    254    &    97,975  &    EN    &  MWP &      Training  & Text & \encircle[fill=myOrangev2, text=white]{U} & \encircle[fill=DarkYellow, text=white]{T}\\
GSM8K \cite{cobbe2021training} &    7,473     &       &  1,319      &   8,792 &     EN   &  MWP & Training  & Text & \encircle[fill=MyGreenv2, text=white]{E}  & \encircle[fill=DarkYellow, text=white]{T}\\
SVAMP \cite{patel2021nlp} &    700     &       &   300     &   1,000  &    EN    &  MWP &     Training &  Formula & \encircle[fill=MyGreenv2, text=white]{E} & \encircle[fill=DarkYellow, text=white]{T}\\
DRAW \cite{upadhyay2015draw}  &   -      &    -   &   -     &   1,000  &    EN    & MWP &   Training   &   Formula & \encircle[fill=myGreen, text=white]{O} & \encircle[fill=DarkYellow, text=white]{T}\\
Dolphin1878 \cite{shi2015automatically}  &      -   &    374   &    1,504    &   1,878  &     EN   & MWP &  Training  & Formula & - & \encircle[fill=DarkYellow, text=white]{T}\\
HMWP \cite{qin2020semantically}  &   -      &  -     &     -   &  5,470   &     CH   & MWP &  Training  &  Formula & - & \encircle[fill=DarkYellow, text=white]{T}\\
ArMATH \cite{alghamdi2022armath}  &   -      &   -    &     -   & 6,000    &    Arabic    & MWP &   Training  &  Formula & \encircle[fill=MyGreenv2, text=white]{E}  & \encircle[fill=DarkYellow, text=white]{T}\\
TabMWP \cite{lu2022dynamic}  &     -    &    -   &    -    &    38,431 &     EN   & MWP &  Training   &  Text &  \encircle[fill=MyGreenv2, text=white]{E} \hspace{1pt} \encircle[fill=myBlue, text=white]{M} & \encircle[fill=DarkYellow, text=white]{T}\\
TAL-SCQ5K \footnote{https://github.com/math-eval/TAL-SCQ5K}  &    3,000     &  -   &     2,000     &   5,000  &   CH/EN     & MWP &   Training      &   Text & \encircle[fill=myViolet, text=white]{C} & \encircle[fill=DarkYellow, text=white]{T}\\
REALFP \cite{kalyan2021much}  &    185     &    185   &     558   &  928  &   EN     & MWP  &   Training      &    Formula  & \encircle[fill=myViolet, text=white]{C} & \encircle[fill=DarkYellow, text=white]{T}\\
SYNTHFP \cite{kalyan2021much}  &    10,000     &   1,000    &   1,000     &  12,000  &   EN     &  MWP &      Training   &    Formula & \encircle[fill=myGreen, text=white]{O}  & \encircle[fill=DarkYellow, text=white]{T}\\
MultiHiertt \cite{zhao2022multihiertt}  &     7,830     &    1,044   &     1,566    &  10,440  &    EN    &  MWP &    Training    &  Formula & \encircle[fill=myGreen, text=white]{O} & \encircle[fill=DarkRed, text=white]{M}\\ 
MATHPILE \cite{wang2023generative} &     -     &    -   &     -    &  903,180  &    EN    &  MWP &    Training    &  Text & \encircle[fill=MyRedv2, text=white]{H} & \encircle[fill=DarkYellow, text=white]{T}\\ 
OpenWebMath \cite{paster2023openwebmath}  &     -    &     -  &   -     &  -   &   EN     & MWP  &    Training   &   Formula  &  \encircle[fill=MyRedv2, text=white]{H} & \encircle[fill=DarkYellow, text=white]{T}\\
MathQA \cite{amini2019mathqa} &    29,837     &    4,475   &   28,985     &  37,297   &     EN   & \highlight{MQA} &    Training  & Formula & - & \encircle[fill=DarkYellow, text=white]{T}\\
FinQA \cite{chen2021finqa}  &    6,251     &   883    &    1,147    &  8,281  &    EN    &  \highlight{MQA}   &    Training    &   Formula  & \encircle[fill=myGreen, text=white]{O} & \encircle[fill=DarkYellow, text=white]{T}\\
TAT-QA \cite{zhu2021tat}  &     -    &   -    &    -    &  16,552  &      EN  &  \highlight{MQA} &     Training     &   Text & \encircle[fill=myGreen, text=white]{O}& \encircle[fill=DarkRed, text=white]{M}\\
MML \cite{grabowski2015four}  &     -    &     -  &   -     &  57,882   &   EN     & TP  &    Training   &   Formula &  -  & \encircle[fill=DarkYellow, text=white]{T}\\
HolStep \cite{kaliszyk2016holstep}  &    2,013,046     &    -   &   196,030     &  2,209,076   &    EN    & TP  &     Training & Formula   &   -  & \encircle[fill=DarkYellow, text=white]{T}\\
CoqGym \cite{yang2019learning}  &      -   & -    &  -      &  71,000   &     EN   & TP  &   Training    &    Formula  & - & \encircle[fill=DarkYellow, text=white]{T}\\
HOList \cite{bansal2019holist}  &    -     &    -   &    -    &  29,462   &     EN   & TP  &    Training    &   Formula  & - & \encircle[fill=DarkYellow, text=white]{T}\\
IsarStep \cite{li2020isarstep}  &    820,000     &   5,000    &  5,000      &  830,000   &   EN     & TP  &   Training   &   Formula    & \encircle[fill=myOrangev2, text=white]{U} & \encircle[fill=DarkYellow, text=white]{T}\\
LISA \cite{jiang2021lisa}  &     -    &   -    &   -     &  183,000   &    EN    & TP  &    Training    &   Formula & - & \encircle[fill=DarkYellow, text=white]{T}\\
NaturalProofs \cite{welleck2021naturalproofs}  &         &       &        &  32,000   &     EN   & TP  &    Training   &   Text   & \encircle[fill=myGreen, text=white]{O} & \encircle[fill=DarkYellow, text=white]{T}\\
LeanStep \cite{han2021proof}  &    -     &     -  &     -   &  21,606,000   &    EN    & TP  &      Training    &   Formula & - & \encircle[fill=DarkYellow, text=white]{T}\\
\hline
NumGLUE \cite{mishra2022numglue}  &   -    &    -   &    -    &  101,835   &     EN   &  Calculation &   Benchmark   &  Text  & \encircle[fill=MyRedv2, text=white]{H} & \encircle[fill=DarkYellow, text=white]{T}\\
Dophin18k \cite{huang2016well} &    -     &   -    &     -   &   18,460   &     EN   &   MWP &   Benchmark  & Text & \encircle[fill=MyGreenv2, text=white]{E} & \encircle[fill=DarkYellow, text=white]{T}\\
MAWPS \cite{koncel2016mawps} &    -     &    -   &      -  &    3,320  &    EN    &  MWP &     Benchmark  &  Formula & \encircle[fill=MyRedv2, text=white]{H} & \encircle[fill=DarkYellow, text=white]{T}\\
ASDiv \cite{miao2020diverse} &    -     &   -    &    -    &   2,305  &   EN     &  MWP &      Benchmark & Formula & \encircle[fill=MyGreenv2, text=white]{E} & \encircle[fill=DarkYellow, text=white]{T}\\
MATH \cite{hendrycks2021measuring} &     7,500    &       &  5,000      &   12,500  & EN       & MWP &     Benchmark  &  Text & \encircle[fill=myYellowv2, text=white]{H} \hspace{1pt} \encircle[fill=myGreen, text=white]{O} & \encircle[fill=DarkYellow, text=white]{T}\\
MGSM \cite{shi2022language}  &    -     &  -     &   -     &  -   &     Multilingual   & MWP &   Benchmark   &  Text & \encircle[fill=MyGreenv2, text=white]{E} & \encircle[fill=DarkYellow, text=white]{T}\\
Mathematics \cite{saxton2018analysing}  &   2,000,000      &     &     100,000     &   2,100,000   &    EN    & MWP &    Benchmark  &    Formula & - & \encircle[fill=DarkYellow, text=white]{T}\\
MMLU-Math \cite{hendryckstest2021}  &      -   &    -   &    -    &   906  &    EN    &  WMP &     Benchmark  & Formula & \encircle[fill=MyGreenv2, text=white]{E} \hspace{1pt} \encircle[fill=myYellowv2, text=white]{H} \hspace{1pt} \encircle[fill=myViolet, text=white]{C} & \encircle[fill=DarkYellow, text=white]{T}\\
AGIEval \cite{zhong2023agieval}  &    -     &    -   &    -    &  469/220   &    CH/EN    & MWP  &  Benchmark     &    Formula & \encircle[fill=myYellowv2, text=white]{H} \hspace{1pt} \encircle[fill=myViolet, text=white]{C} \hspace{1pt} \encircle[fill=myGreen, text=white]{O}  & \encircle[fill=DarkYellow, text=white]{T}\\
\highlight{AIME\footnote{https://www.kaggle.com/datasets/hemishveeraboina/aime-problem-set-1983-2024}} &    -     &    -   &     -   &  \highlight{933}   &    \highlight{EN}    & \highlight{MWP}  &     \highlight{Benchmark}    &   \highlight{Formula} & \encircle[fill=myViolet, text=white]{C} & \encircle[fill=DarkYellow, text=white]{T}\\
\highlight{MATHTRAP \cite{zhao2024exploring}} & - & - & - & \highlight{105/155} & \highlight{EN} & \highlight{MWP} & \highlight{Benchmark} & \highlight{Formula} & \encircle[fill=MyRedv2, text=white]{H} & \encircle[fill=DarkYellow, text=white]{T}\\
\highlight{MathVista \cite{lu2023mathvista}} & - & - & - & \highlight{6,141} & \highlight{EN} & \highlight{MWP} & \highlight{Benchmark} & \highlight{Formula} & \encircle[fill=MyGreenv2, text=white]{E} \hspace{1pt} \encircle[fill=myBlue, text=white]{M} \hspace{1pt} \encircle[fill=myYellowv2, text=white]{H} \hspace{1pt} \encircle[fill=myOrangev2, text=white]{U} & \encircle[fill=DarkRed, text=white]{M} \\
\highlight{Math-V \cite{wang2024measuring}} & - & - & - & \highlight{3,040} & \highlight{EN} & \highlight{MWP} & \highlight{Benchmark} & \highlight{Formula} & \encircle[fill=MyGreenv2, text=white]{E} \hspace{1pt} \encircle[fill=myYellowv2, text=white]{H}  \hspace{1pt} \encircle[fill=myGreen, text=white]{O} & \encircle[fill=DarkRed, text=white]{M} \\
\highlight{CMM-Math \cite{liu2024cmm} } &     \highlight{22,248}    &   -    &     \highlight{5,821}   &  \highlight{28,069}  &   \highlight{CH}    &  \highlight{MWP}   &  \highlight{Benchmark}    &    \highlight{Text} & \encircle[fill=MyGreenv2, text=white]{E} \hspace{1pt} \encircle[fill=myBlue, text=white]{M} \hspace{1pt} \encircle[fill=myYellowv2, text=white]{H} & \encircle[fill=DarkRed, text=white]{M} \\
\highlight{MathVerse \cite{zhang2025mathverse}} & - & - & - & \highlight{2,612} & \highlight{EN} & \highlight{MWP} & \highlight{Benchmark} & \highlight{Formula} & \encircle[fill=MyRedv2, text=white]{H} & \encircle[fill=DarkRed, text=white]{M} \\
\highlight{GeoQA \cite{chen2021geoqa}} & \highlight{3,499}  & \highlight{745}  & \highlight{754} & \highlight{4,998} & \highlight{EN} & \highlight{MWP} & \highlight{Benchmark} & \highlight{Formula} & \encircle[fill=myBlue, text=white]{M} & \encircle[fill=DarkRed, text=white]{M} \\
\highlight{M3CoT \cite{chen-etal-2024-m3cot}} & - & - & - & \highlight{11,459} & \highlight{EN} & \highlight{MWP} & \highlight{Benchmark} & \highlight{Formula} & \encircle[fill=myGreen, text=white]{O} & \encircle[fill=DarkRed, text=white]{M} \\
\highlight{MR-MATH \cite{xia2024evaluating}} & - & - & - & - & \highlight{EN} & \highlight{MWP} & \highlight{Benchmark} & \highlight{Formula} & \encircle[fill=MyGreenv2, text=white]{E} & \encircle[fill=DarkYellow, text=white]{T}\\
\highlight{MMMU-Math \cite{yue2024mmmumassivemultidisciplinemultimodal}} & - & - & - & \highlight{505} & \highlight{EN} & \highlight{MWP} & \highlight{Benchmark} & \highlight{Text} & \encircle[fill=myOrangev2, text=white]{U} & \encircle[fill=DarkRed, text=white]{M} \\
\highlight{We-Math \cite{qiao2024wemathdoeslargemultimodal}} & - & - & - & \highlight{1700} & \highlight{EN} & \highlight{MWP} & \highlight{Benchmark} & \highlight{Formula} & \encircle[fill=MyGreenv2, text=white]{E} \hspace{1pt} \encircle[fill=myYellowv2, text=white]{H}  \hspace{1pt} 
 \encircle[fill=myOrangev2, text=white]{U} & \encircle[fill=DarkRed, text=white]{M} \\
\highlight{U-MATH \cite{chernyshev2024umathuniversitylevelbenchmarkevaluating}}  &     -    &   -    &   -     &  \highlight{1,080}   &     \highlight{EN}   & \highlight{MWP}  &    \highlight{Benchmark}      &   \highlight{Formula} & \encircle[fill=myOrangev2, text=white]{U} & \encircle[fill=DarkRed, text=white]{M} \\
INT \cite{wu2020int}  &     -    &  -     &     -   &  -   &   EN     & TP  &     Benchmark     &  Formula & - & \encircle[fill=DarkYellow, text=white]{T}\\
miniF2F \cite{zheng2021minif2f}  &     -    &   244    &   244     &  488   &     EN   & TP  &    Benchmark      &   Formula & \encircle[fill=MyGreenv2, text=white]{E} \hspace{1pt} \encircle[fill=myYellowv2, text=white]{H}  \hspace{1pt} \encircle[fill=myGreen, text=white]{O} & \encircle[fill=DarkYellow, text=white]{T}\\
\highlight{FOLIO \cite{DBLP:conf/emnlp/HanS0QRZCPQBSWS24}}  &    -     &     -  &     -   &  \highlight{1,430}   &    \highlight{EN}    & \highlight{TP}  &   \highlight{Benchmark}    & \highlight{Formula} & \encircle[fill=MyRedv2, text=white]{H} & \encircle[fill=DarkYellow, text=white]{T}\\
\hline
Aggregate \cite{roy2018mapping}  &     -    &    -   &   -     &   1,492  &    EN    & MWP &    Augmented   &  Formula & \encircle[fill=MyRedv2, text=white]{H} & \encircle[fill=DarkYellow, text=white]{T}\\
MathQA-Python \cite{austin2021program}  &   19,209      &   2,822    &    2,822    &  23,914   &    EN    &  MWP &   Augmented   &  Code & - & \encircle[fill=DarkYellow, text=white]{T}\\
Math50k \cite{li2023camel}  &     -    &    -   &   -     &   50,000  &    EN    &  WMP &  Augmented    &   Text &  -& \encircle[fill=DarkYellow, text=white]{T}\\
PRM800K \cite{DBLP:conf/iclr/LightmanKBEBLLS24}  &      -   &    -   &   -     &   2,868  &    EN    &   WMP &    Augmented  &   Text & \encircle[fill=myYellowv2, text=white]{H} \hspace{1pt} \encircle[fill=myGreen, text=white]{O} & \encircle[fill=DarkYellow, text=white]{T}\\
MetaMathQA \cite{yu2023metamath}  &   -      &    -   &    -    &    395,000  &     EN   &  MWP &  Augmented   &   Text & \encircle[fill=MyRedv2, text=white]{H}  & \encircle[fill=DarkYellow, text=white]{T}\\
Lila \cite{mishra2022lila}  &      -   &    -   &    -    &   134,000  &     EN   &  MWP &  Augmented   & Code  & \encircle[fill=MyRedv2, text=white]{H} & \encircle[fill=DarkYellow, text=white]{T}\\
PEN \cite{kim2022ept}  &     -    &     -  & -       &   3,581   &     EN   & MWP &   Augmented   &  Formula & \encircle[fill=MyRedv2, text=white]{H} & \encircle[fill=DarkYellow, text=white]{T}\\
miniF2F+informal \cite{jiang2022draft}  &    -     &    244   &     244   &  488   &    EN    & TP  &     Augmented    &   Formula & \encircle[fill=MyGreenv2, text=white]{E} \hspace{1pt} \encircle[fill=myYellowv2, text=white]{H}  \hspace{1pt} \encircle[fill=myGreen, text=white]{O} & \encircle[fill=DarkYellow, text=white]{T}\\
NaturalProofs-Gen \cite{welleck2022naturalprover}  &     12,500    &  1,000     &   1,000     &  14,500   &    EN    & TP  &      Augmented &    Text  & \encircle[fill=myGreen, text=white]{O} & \encircle[fill=DarkYellow, text=white]{T}\\
\hline{4}
\end{tabular}}
\end{table*}

%% file: llm_opt_con_survey.bbl
\begin{thebibliography}{100}

\bibitem{feigenbaum1963computers}
E.~A. Feigenbaum, J.~Feldman, {\em et~al.}, {\em Computers and thought},
  vol.~7.
\newblock 1963.

\bibitem{bobrow1964natural}
D.~Bobrow {\em et~al.}, ``Natural language input for a computer problem solving
  system,'' 1964.

\bibitem{briars1984integrated}
D.~J. Briars and J.~H. Larkin, ``An integrated model of skill in solving
  elementary word problems,'' {\em Cognition and instruction}, vol.~1, no.~3,
  pp.~245--296, 1984.

\bibitem{fletcher1985understanding}
C.~R. Fletcher, ``Understanding and solving arithmetic word problems: A
  computer simulation,'' {\em Behavior Research Methods, Instruments, \&
  Computers}, vol.~17, no.~5, pp.~565--571, 1985.

\bibitem{wang2017deep}
Y.~Wang, X.~Liu, and S.~Shi, ``Deep neural solver for math word problems,'' in
  {\em EMNLP}, pp.~845--854, 2017.

\bibitem{hosseini-etal-2014-learning}
M.~J. Hosseini, H.~Hajishirzi, O.~Etzioni, and N.~Kushman, ``Learning to solve
  arithmetic word problems with verb categorization,'' in {\em EMNLP},
  pp.~523--533, 2014.

\bibitem{kushman-etal-2014-learning}
N.~Kushman, Y.~Artzi, L.~Zettlemoyer, and R.~Barzilay, ``Learning to
  automatically solve algebra word problems,'' in {\em ACL}, pp.~271--281, June
  2014.

\bibitem{zhou2015learn}
L.~Zhou, S.~Dai, and L.~Chen, ``Learn to solve algebra word problems using
  quadratic programming,'' in {\em EMNLP}, pp.~817--822, 2015.

\bibitem{mitra2016learning}
A.~Mitra and C.~Baral, ``Learning to use formulas to solve simple arithmetic
  problems,'' in {\em ACL}, pp.~2144--2153, 2016.

\bibitem{roy2015reasoning}
S.~Roy, T.~Vieira, and D.~Roth, ``Reasoning about quantities in natural
  language,'' {\em TACL}, vol.~3, pp.~1--13, 2015.

\bibitem{roy2015solving}
S.~Roy and D.~Roth, ``Solving general arithmetic word problems,'' in {\em
  EMNLP}, pp.~1743--1752, 2015.

\bibitem{shi2015automatically}
S.~Shi, Y.~Wang, C.-Y. Lin, X.~Liu, and Y.~Rui, ``Automatically solving number
  word problems by semantic parsing and reasoning,'' in {\em EMNLP},
  pp.~1132--1142, 2015.

\bibitem{koncel2015parsing}
R.~Koncel-Kedziorski, H.~Hajishirzi, A.~Sabharwal, O.~Etzioni, and S.~D. Ang,
  ``Parsing algebraic word problems into equations,'' {\em TACL}, vol.~3,
  pp.~585--597, 2015.

\bibitem{couperus2023large}
J.~Couperus, ``Large language models and mathematical understanding,'' Master's
  thesis, 2023.

\bibitem{wei2022chain}
J.~Wei, X.~Wang, D.~Schuurmans, M.~Bosma, F.~Xia, E.~Chi, Q.~V. Le, D.~Zhou,
  {\em et~al.}, ``Chain-of-thought prompting elicits reasoning in large
  language models,'' {\em NeurIPS}, vol.~35, pp.~24824--24837, 2022.

\bibitem{matzakos2023learning}
N.~Matzakos, S.~Doukakis, and M.~Moundridou, ``Learning mathematics with large
  language models: A comparative study with computer algebra systems and other
  tools,'' {\em International Journal of Emerging Technologies in Learning
  (Online)}, vol.~18, no.~20, p.~51, 2023.

\bibitem{devlinBERTPretrainingDeep2019}
J.~Devlin, M.-W. Chang, K.~Lee, and K.~Toutanova, ``{{BERT}}: {{Pre-training}}
  of {{Deep Bidirectional Transformers}} for {{Language Understanding}},''
  2019.

\bibitem{liuRoBERTaRobustlyOptimized2019}
Y.~Liu, M.~Ott, N.~Goyal, J.~Du, M.~Joshi, D.~Chen, O.~Levy, M.~Lewis,
  L.~Zettlemoyer, and V.~Stoyanov, ``{{RoBERTa}}: {{A}} robustly optimized
  {{BERT}} pretraining approach,'' 2019.

\bibitem{lewis2020bart}
M.~Lewis, Y.~Liu, N.~Goyal, M.~Ghazvininejad, A.~Mohamed, O.~Levy, V.~Stoyanov,
  and L.~Zettlemoyer, ``Bart: Denoising sequence-to-sequence pre-training for
  natural language generation, translation, and comprehension,'' in {\em ACL},
  pp.~7871--7880, 2020.

\bibitem{radford_gpt-1_2018}
A.~Radford, K.~Narasimhan, T.~Salimans, and I.~Sutskever, ``{Improving}
  {Language} {Understanding} by {Generative} {Pre}-{Training},'' p.~12, June
  2018.

\bibitem{radford_gpt-2_2019}
A.~Radford, J.~Wu, R.~Child, D.~Luan, D.~Amodei, and I.~Sutskever, ``{Language}
  {Models} are {Unsupervised} {Multitask} {Learners},'' p.~24, Feb. 2019.

\bibitem{clarkRegentsScienceExams2021}
P.~Clark, O.~Etzioni, D.~Khashabi, T.~Khot, B.~D. Mishra, K.~Richardson,
  A.~Sabharwal, C.~Schoenick, O.~Tafjord, N.~Tandon, S.~Bhakthavatsalam,
  D.~Groeneveld, M.~Guerquin, and M.~Schmitz, ``From '{{F}}' to 'a' on the
  {{N}}.{{Y}}. regents science exams: {{An}} overview of the aristo project,''
  2021.

\bibitem{gevaInjectingNumericalReasoning2020}
M.~Geva, A.~Gupta, and J.~Berant, ``Injecting numerical reasoning skills into
  language models,'' in {\em ACL} (D.~Jurafsky, J.~Chai, N.~Schluter, and
  J.~Tetreault, eds.), pp.~946--958, July 2020.

\bibitem{fengInjectingNumericalReasoning2022}
Y.~Feng, J.~Zhang, X.~Zhang, L.~Liu, C.~Li, and H.~Chen, ``Injecting numerical
  reasoning skills into knowledge base question answering models,'' 2022.

\bibitem{pengMathBERTPretrainedModel2021}
S.~Peng, K.~Yuan, L.~Gao, and Z.~Tang, ``{{MathBERT}}: {{A}} pre-trained model
  for mathematical formula understanding,'' 2021.

\bibitem{jiang2021lisa}
A.~Q. Jiang, W.~Li, J.~M. Han, and Y.~Wu, ``Lisa: Language models of isabelle
  proofs,'' in {\em AITP}, pp.~378--392, 2021.

\bibitem{zhu2021tat}
F.~Zhu, W.~Lei, Y.~Huang, C.~Wang, S.~Zhang, J.~Lv, F.~Feng, and T.-S. Chua,
  ``Tat-qa: A question answering benchmark on a hybrid of tabular and textual
  content in finance,'' in {\em ACL}, pp.~3277--3287, 2021.

\bibitem{zhao2022multihiertt}
Y.~Zhao, Y.~Li, C.~Li, and R.~Zhang, ``Multihiertt: Numerical reasoning over
  multi hierarchical tabular and textual data,'' in {\em ACL}, pp.~6588--6600,
  2022.

\bibitem{jieLearningReasonDeductively2022}
Z.~Jie, J.~Li, and W.~Lu, ``Learning to reason deductively: {{Math}} word
  problem solving as complex relation extraction,'' 2022.

\bibitem{liSeekingPatternsNot2022}
Z.~Li, W.~Zhang, C.~Yan, Q.~Zhou, C.~Li, H.~Liu, and Y.~Cao, ``Seeking
  patterns, not just memorizing procedures: {{Contrastive}} learning for
  solving math word problems,'' 2022.

\bibitem{openai2023gpt4}
OpenAI, ``Gpt-4 technical report,'' 2023.

\bibitem{min2022metaicl}
S.~Min, M.~Lewis, L.~Zettlemoyer, and H.~Hajishirzi, ``Metaicl: Learning to
  learn in context,'' in {\em NAACL}, pp.~2791--2809, 2022.

\bibitem{brown2020language}
T.~Brown, B.~Mann, N.~Ryder, M.~Subbiah, J.~D. Kaplan, P.~Dhariwal,
  A.~Neelakantan, P.~Shyam, G.~Sastry, A.~Askell, {\em et~al.}, ``Language
  models are few-shot learners,'' {\em NeurIPS}, vol.~33, pp.~1877--1901, 2020.

\bibitem{chen2022meta}
Y.~Chen, R.~Zhong, S.~Zha, G.~Karypis, and H.~He, ``Meta-learning via language
  model in-context tuning,'' in {\em ACL}, pp.~719--730, 2022.

\bibitem{yang2023gpt}
Z.~Yang, M.~Ding, Q.~Lv, Z.~Jiang, Z.~He, Y.~Guo, J.~Bai, and J.~Tang, ``Gpt
  can solve mathematical problems without a calculator,'' {\em arXiv}, 2023.

\bibitem{liu_goat_2023}
T.~Liu and B.~K.~H. Low, ``Goat: Fine-tuned llama outperforms gpt-4 on
  arithmetic tasks,'' {\em arXiv}, 2023.

\bibitem{wang2022self}
X.~Wang, J.~Wei, D.~Schuurmans, Q.~V. Le, E.~H. Chi, S.~Narang, A.~Chowdhery,
  and D.~Zhou, ``Self-consistency improves chain of thought reasoning in
  language models,'' in {\em ICLR}, 2023.

\bibitem{chowdhery2022palm}
A.~Chowdhery, S.~Narang, J.~Devlin, M.~Bosma, G.~Mishra, A.~Roberts, P.~Barham,
  H.~W. Chung, C.~Sutton, S.~Gehrmann, {\em et~al.}, ``Palm: Scaling language
  modeling with pathways,'' {\em arXiv}, 2022.

\bibitem{chen2022program}
W.~Chen, X.~Ma, X.~Wang, and W.~W. Cohen, ``Program of thoughts prompting:
  Disentangling computation from reasoning for numerical reasoning tasks,''
  {\em arXiv}, 2022.

\bibitem{lu2022dynamic}
P.~Lu, L.~Qiu, K.-W. Chang, Y.~N. Wu, S.-C. Zhu, T.~Rajpurohit, P.~Clark, and
  A.~Kalyan, ``Dynamic prompt learning via policy gradient for semi-structured
  mathematical reasoning,'' in {\em ICLR}, 2023.

\bibitem{zhang2023automatic}
Z.~Zhang, A.~Zhang, M.~Li, and A.~Smola, ``Automatic chain of thought prompting
  in large language models,'' in {\em ICLR}, 2023.

\bibitem{fu2022complexity}
Y.~Fu, H.~Peng, A.~Sabharwal, P.~Clark, and T.~Khot, ``Complexity-based
  prompting for multi-step reasoning,'' in {\em ICLR}, 2023.

\bibitem{gao2023pal}
L.~Gao, A.~Madaan, S.~Zhou, U.~Alon, P.~Liu, Y.~Yang, J.~Callan, and G.~Neubig,
  ``Pal: Program-aided language models,'' in {\em ICML}, pp.~10764--10799,
  2023.

\bibitem{drori2022neural}
I.~Drori, S.~Zhang, R.~Shuttleworth, L.~Tang, A.~Lu, E.~Ke, K.~Liu, L.~Chen,
  S.~Tran, N.~Cheng, {\em et~al.}, ``A neural network solves, explains, and
  generates university math problems by program synthesis and few-shot learning
  at human level,'' {\em PNAS}, vol.~119, no.~32, p.~e2123433119, 2022.

\bibitem{he2023solving}
J.~He-Yueya, G.~Poesia, R.~E. Wang, and N.~D. Goodman, ``Solving math word
  problems by combining language models with symbolic solvers,'' {\em arXiv},
  2023.

\bibitem{lu-etal-2023-survey}
P.~Lu, L.~Qiu, W.~Yu, S.~Welleck, and K.-W. Chang, ``A survey of deep learning
  for mathematical reasoning,'' in {\em ACL} (A.~Rogers, J.~Boyd-Graber, and
  N.~Okazaki, eds.), pp.~14605--14631, 2023.

\bibitem{qiao-etal-2023-reasoning}
S.~Qiao, Y.~Ou, N.~Zhang, X.~Chen, Y.~Yao, S.~Deng, C.~Tan, F.~Huang, and
  H.~Chen, ``Reasoning with language model prompting: A survey,'' in {\em ACL},
  pp.~5368--5393, 2023.

\bibitem{chuCoTReasoningSurvey2023}
Z.~Chu, J.~Chen, Q.~Chen, W.~Yu, T.~He, H.~Wang, W.~Peng, M.~Liu, B.~Qin, and
  T.~Liu, ``A survey of chain of thought reasoning: Advances, frontiers and
  future,'' 2023.

\bibitem{qiu2020pre}
X.~Qiu, T.~Sun, Y.~Xu, Y.~Shao, N.~Dai, and X.~Huang, ``Pre-trained models for
  natural language processing: A survey,'' {\em Science China Technological
  Sciences}, vol.~63, no.~10, pp.~1872--1897, 2020.

\bibitem{zhao2023survey}
W.~X. Zhao, K.~Zhou, J.~Li, T.~Tang, X.~Wang, Y.~Hou, Y.~Min, B.~Zhang,
  J.~Zhang, Z.~Dong, {\em et~al.}, ``A survey of large language models,'' {\em
  arXiv}, 2023.

\bibitem{yan2024survey}
Y.~Yan, J.~Su, J.~He, F.~Fu, X.~Zheng, Y.~Lyu, K.~Wang, S.~Wang, Q.~Wen, and
  X.~Hu, ``A survey of mathematical reasoning in the era of multimodal large
  language model: Benchmark, method \& challenges,'' {\em arXiv}, 2024.

\bibitem{liu2023mathematical}
W.~Liu, H.~Hu, J.~Zhou, Y.~Ding, J.~Li, J.~Zeng, M.~He, Q.~Chen, B.~Jiang,
  A.~Zhou, {\em et~al.}, ``Mathematical language models: A survey,'' {\em arXiv
  preprint arXiv:2312.07622}, 2023.

\bibitem{imani2023mathprompter}
S.~Imani, L.~Du, and H.~Shrivastava, ``Mathprompter: Mathematical reasoning
  using large language models,'' {\em arXiv}, 2023.

\bibitem{wang2024leveraging}
T.~Wang, W.-Y. Yu, R.~She, W.~Yang, T.~Chen, and J.~Zhang, ``Leveraging large
  language models for solving rare {MIP} challenges,'' {\em arXiv preprint
  arXiv:2409.04464}, 2024.

\bibitem{ahmed2403lm4opt}
T.~Ahmed and S.~Choudhury, ``Lm4opt: Unveiling the potential of large language
  models in formulating mathematical optimization problems, 2024,'' {\em URL:
  https://arxiv. org/abs/2403.01342}.

\bibitem{Xuhan2025}
X.~Huang, Q.~Shen, Y.~Hu, A.~Gao, and B.~Wang, ``Llms for mathematical
  modeling: Towards bridging the gap between natural and mathematical
  languages,'' {\em arXiv:2405.13144v3}, 2025.

\bibitem{lai2024executing}
J.~Lai, J.~Xu, Y.~Yang, Y.~Huang, C.~Cao, and J.~Xu, ``Executing arithmetic:
  Fine-tuning large language models as turing machines,'' {\em arXiv preprint
  arXiv:2410.07896}, 2024.

\bibitem{dua2019drop}
D.~Dua, Y.~Wang, P.~Dasigi, G.~Stanovsky, S.~Singh, and M.~Gardner, ``Drop: A
  reading comprehension benchmark requiring discrete reasoning over
  paragraphs,'' in {\em NAACL}, pp.~2368--2378, 2019.

\bibitem{zhang2020language}
X.~Zhang, D.~Ramachandran, I.~Tenney, Y.~Elazar, and D.~Roth, ``Do language
  embeddings capture scales?,'' in {\em BlackboxNLP Workshop on Analyzing and
  Interpreting Neural Networks for NLP}, pp.~292--299, 2020.

\bibitem{spithourakis2018numeracy}
G.~Spithourakis and S.~Riedel, ``Numeracy for language models: Evaluating and
  improving their ability to predict numbers,'' in {\em ACL}, vol.~56,
  pp.~2104--2115, 2018.

\bibitem{wallace2019nlp}
E.~Wallace, Y.~Wang, S.~Li, S.~Singh, and M.~Gardner, ``Do nlp models know
  numbers? probing numeracy in embeddings,'' in {\em EMNLP}, pp.~5307--5315,
  2019.

\bibitem{berg-kirkpatrickEmpiricalInvestigationContextualized2020}
T.~Berg-Kirkpatrick and D.~Spokoyny, ``An empirical investigation of
  contextualized number prediction,'' in {\em EMNLP}, p.~4754–4764, Nov.
  2020.

\bibitem{nogueira2021investigating}
R.~Nogueira, Z.~Jiang, and J.~Lin, ``Investigating the limitations of
  transformers with simple arithmetic tasks,'' {\em arXiv}, 2021.

\bibitem{wang2021exploring}
C.~Wang, B.~Zheng, Y.~Niu, and Y.~Zhang, ``Exploring generalization ability of
  pretrained language models on arithmetic and logical reasoning,'' in {\em
  NLPCC}, pp.~758--769, 2021.

\bibitem{muffo_evaluating_nodate}
M.~Muffo, A.~Cocco, and E.~Bertino, ``Evaluating {Transformer} {Language}
  {Models} on {Arithmetic} {Operations} {Using} {Number} {Decomposition},''

\bibitem{yuan2023well}
Z.~Yuan, H.~Yuan, C.~Tan, W.~Wang, and S.~Huang, ``How well do large language
  models perform in arithmetic tasks?,'' {\em arXiv}, 2023.

\bibitem{taylor_galactica_2022}
R.~Taylor, M.~Kardas, G.~Cucurull, T.~Scialom, A.~Hartshorn, E.~Saravia,
  A.~Poulton, V.~Kerkez, and R.~Stojnic, ``Galactica: {A} {Large} {Language}
  {Model} for {Science},'' 2022.

\bibitem{touvron2023llama}
H.~Touvron, T.~Lavril, G.~Izacard, X.~Martinet, M.-A. Lachaux, T.~Lacroix,
  B.~Rozi{\`e}re, N.~Goyal, E.~Hambro, F.~Azhar, {\em et~al.}, ``Llama: Open
  and efficient foundation language models,'' {\em arXiv}, 2023.

\bibitem{zhouTeachingAlgorithmicReasoning2022}
H.~Zhou, A.~Nova, H.~Larochelle, A.~Courville, B.~Neyshabur, and H.~Sedghi,
  ``Teaching algorithmic reasoning via in-context learning,'' 2022.

\bibitem{jelassiLengthGeneralizationArithmetic2023}
S.~Jelassi, S.~d'Ascoli, C.~Domingo-Enrich, Y.~Wu, Y.~Li, and F.~Charton,
  ``Length generalization in arithmetic transformers,'' {\em arXiv}, 2023.

\bibitem{nye_show_2021}
M.~Nye, A.~J. Andreassen, G.~Gur-Ari, H.~Michalewski, J.~Austin, D.~Bieber,
  D.~Dohan, A.~Lewkowycz, M.~Bosma, D.~Luan, C.~Sutton, and A.~Odena, ``Show
  {Your} {Work}: {Scratchpads} for {Intermediate} {Computation} with {Language}
  {Models},'' 2021.

\bibitem{yuan2023scaling}
Z.~Yuan, H.~Yuan, C.~Li, G.~Dong, C.~Tan, and C.~Zhou, ``Scaling relationship
  on learning mathematical reasoning with large language models,'' {\em arXiv},
  2023.

\bibitem{yu2023metamath}
L.~Yu, W.~Jiang, H.~Shi, J.~Yu, Z.~Liu, Y.~Zhang, J.~T. Kwok, Z.~Li, A.~Weller,
  and W.~Liu, ``Metamath: Bootstrap your own mathematical questions for large
  language models,'' {\em arXiv}, 2023.

\bibitem{luo2023wizardmath}
H.~Luo, Q.~Sun, C.~Xu, P.~Zhao, J.~Lou, C.~Tao, X.~Geng, Q.~Lin, S.~Chen, and
  D.~Zhang, ``Wizardmath: Empowering mathematical reasoning for large language
  models via reinforced evol-instruct,'' {\em arXiv}, 2023.

\bibitem{zhou2023mathattack}
Z.~Zhou, Q.~Wang, M.~Jin, J.~Yao, J.~Ye, W.~Liu, W.~Wang, X.~Huang, and
  K.~Huang, ``Mathattack: Attacking large language models towards math solving
  ability,'' 2023.

\bibitem{azerbayev8llemma}
Z.~Azerbayev, H.~Schoelkopf, K.~Paster, M.~Dos~Santos, S.~McAleer, A.~Q. Jiang,
  J.~Deng, S.~Biderman, and S.~Welleck, ``Llemma: An open language model for
  mathematics,'' {\em Minerva}, vol.~8, p.~164B.

\bibitem{polu2020generative}
S.~Polu and I.~Sutskever, ``Generative language modeling for automated theorem
  proving,'' 2020.

\bibitem{jiang2022draft}
A.~Q. Jiang, S.~Welleck, J.~P. Zhou, T.~Lacroix, J.~Liu, W.~Li, M.~Jamnik,
  G.~Lample, and Y.~Wu, ``Draft, sketch, and prove: Guiding formal theorem
  provers with informal proofs,'' in {\em ICLR}, 2023.

\bibitem{welleck2021naturalproofs}
S.~Welleck, J.~Liu, R.~Le~Bras, H.~Hajishirzi, Y.~Choi, and K.~Cho,
  ``Naturalproofs: Mathematical theorem proving in natural language,'' in {\em
  NeurIPS Datasets and Benchmarks Track (Round 1)}, 2021.

\bibitem{irving2016deepmath}
G.~Irving, C.~Szegedy, A.~A. Alemi, N.~E{\'e}n, F.~Chollet, and J.~Urban,
  ``Deepmath-deep sequence models for premise selection,'' {\em NeurIPS},
  vol.~29, 2016.

\bibitem{wu2020int}
Y.~Wu, A.~Jiang, J.~Ba, and R.~B. Grosse, ``Int: An inequality benchmark for
  evaluating generalization in theorem proving,'' in {\em ICLR}, 2021.

\bibitem{yang2019learning}
K.~Yang and J.~Deng, ``Learning to prove theorems via interacting with proof
  assistants,'' in {\em ICML}, pp.~6984--6994, 2019.

\bibitem{maynez-etal-2020-faithfulness}
J.~Maynez, S.~Narayan, B.~Bohnet, and R.~McDonald, ``On faithfulness and
  factuality in abstractive summarization,'' in {\em ACL}, pp.~1906--1919, July
  2020.

\bibitem{ji2023survey}
Z.~Ji, N.~Lee, R.~Frieske, T.~Yu, D.~Su, Y.~Xu, E.~Ishii, Y.~J. Bang,
  A.~Madotto, and P.~Fung, ``Survey of hallucination in natural language
  generation,'' {\em ACM Computing Surveys}, vol.~55, no.~12, pp.~1--38, 2023.

\bibitem{vaswani2017attention}
A.~Vaswani, N.~Shazeer, N.~Parmar, J.~Uszkoreit, L.~Jones, A.~N. Gomez,
  {\L}.~Kaiser, and I.~Polosukhin, ``Attention is all you need,'' {\em
  NeurIPS}, vol.~30, 2017.

\bibitem{noorbakhsh2021pretrained}
K.~Noorbakhsh, M.~Sulaiman, M.~Sharifi, K.~Roy, and P.~Jamshidi, ``Pretrained
  language models are symbolic mathematics solvers too!,'' {\em arXiv}, 2021.

\bibitem{rael_exploring_nodate}
C.~Raffel, N.~Shazeer, A.~Roberts, K.~Lee, S.~Narang, M.~Matena, Y.~Zhou,
  W.~Li, and P.~J. Liu, ``Exploring the {Limits} of {Transfer} {Learning} with
  a {Uniﬁed} {Text}-to-{Text} {Transformer},'' {\em JMLR}, vol.~21, no.~140,
  pp.~1--67, 2020.

\bibitem{kenton2019bert}
J.~D. M.-W.~C. Kenton and L.~K. Toutanova, ``Bert: Pre-training of deep
  bidirectional transformers for language understanding,'' in {\em NAACL},
  pp.~4171--4186, 2019.

\bibitem{kim_point_2020}
B.~Kim, K.~S. Ki, D.~Lee, and G.~Gweon, ``Point to the {Expression}: {Solving}
  {Algebraic} {Word} {Problems} using the {Expression}-{Pointer} {Transformer}
  {Model},'' in {\em EMNLP}, (Online), pp.~3768--3779, 2020.

\bibitem{shen_generate_2021}
J.~Shen, Y.~Yin, L.~Li, L.~Shang, X.~Jiang, M.~Zhang, and Q.~Liu, ``Generate \&
  {Rank}: {A} {Multi}-task {Framework} for {Math} {Word} {Problems},'' 2021.

\bibitem{jiang2022thor}
A.~Q. Jiang, W.~Li, S.~Tworkowski, K.~Czechowski, T.~Odrzygóźdź, P.~Miłoś,
  Y.~Wu, and M.~Jamnik, ``Thor: Wielding hammers to integrate language models
  and automated theorem provers,'' 2022.

\bibitem{lample_hypertree_nodate}
G.~Lample, M.-A. Lachaux, T.~Lavril, X.~Martinet, A.~Hayat, G.~Ebner,
  A.~Rodriguez, and T.~Lacroix, ``{HyperTree} {Proof} {Search} for {Neural}
  {Theorem} {Proving},''

\bibitem{hendryckstest2021}
D.~Hendrycks, C.~Burns, S.~Basart, A.~Zou, M.~Mazeika, D.~Song, and
  J.~Steinhardt, ``Measuring massive multitask language understanding,'' {\em
  ICLR}, 2021.

\bibitem{hendrycks2021measuring}
D.~Hendrycks, C.~Burns, S.~Kadavath, A.~Arora, S.~Basart, E.~Tang, D.~Song, and
  J.~Steinhardt, ``Measuring mathematical problem solving with the math
  dataset,'' in {\em NeurIPS Datasets and Benchmarks Track (Round 2)}, 2021.

\bibitem{han2021proof}
J.~M. Han, J.~Rute, Y.~Wu, E.~Ayers, and S.~Polu, ``Proof artifact co-training
  for theorem proving with language models,'' in {\em ICLR}, 2022.

\bibitem{lewkowycz2022solving}
A.~Lewkowycz, A.~Andreassen, D.~Dohan, E.~Dyer, H.~Michalewski, V.~Ramasesh,
  A.~Slone, C.~Anil, I.~Schlag, T.~Gutman-Solo, {\em et~al.}, ``Solving
  quantitative reasoning problems with language models,'' {\em NeurIPS},
  vol.~35, pp.~3843--3857, 2022.

\bibitem{wu_lime_nodate}
Y.~Wu, M.~Rabe, and W.~Li, ``{LIME}: {Learning} {Inductive} {Bias} for
  {Primitives} of {Mathematical} {Reasoning},''

\bibitem{liangMWPBERTNumeracyaugmentedPretraining2022}
Z.~Liang, J.~Zhang, L.~Wang, W.~Qin, Y.~Lan, J.~Shao, and X.~Zhang,
  ``{{MWP-BERT}}: {{Numeracy-augmented}} pre-training for math word problem
  solving,'' 2022.

\bibitem{nakamoto2023enhancing}
R.~NAKAMOTO, B.~Flanagan, T.~Yamauchi, D.~Yilling, K.~Takami, and H.~Ogata,
  ``Enhancing automated scoring of math self-explanation quality using
  llm-generated datasets: A semi-supervised approach,'' 2023.

\bibitem{zhang2024learn}
Z.~Zhang, T.~Ge, Z.~Liang, W.~Yu, D.~Yu, M.~Jia, D.~Yu, and M.~Jiang, ``Learn
  beyond the answer: Training language models with reflection for mathematical
  reasoning,'' {\em arXiv}, 2024.

\bibitem{xu_wizardlm_2023}
C.~Xu, Q.~Sun, K.~Zheng, X.~Geng, P.~Zhao, J.~Feng, C.~Tao, and D.~Jiang,
  ``Wizardlm: Empowering large language models to follow complex
  instructions,'' {\em arXiv}, 2023.

\bibitem{yuan2024advancing}
L.~Yuan, G.~Cui, H.~Wang, N.~Ding, X.~Wang, J.~Deng, B.~Shan, H.~Chen, R.~Xie,
  Y.~Lin, {\em et~al.}, ``Advancing llm reasoning generalists with preference
  trees,'' {\em arXiv}, 2024.

\bibitem{rafailov2024direct}
R.~Rafailov, A.~Sharma, E.~Mitchell, C.~D. Manning, S.~Ermon, and C.~Finn,
  ``Direct preference optimization: Your language model is secretly a reward
  model,'' {\em NeurIPS}, vol.~36, 2024.

\bibitem{chen2024noise}
H.~Chen, G.~He, L.~Yuan, G.~Cui, H.~Su, and J.~Zhu, ``Noise contrastive
  alignment of language models with explicit rewards,'' {\em arXiv}, 2024.

\bibitem{zhang_instruction_2023}
S.~Zhang, L.~Dong, X.~Li, S.~Zhang, X.~Sun, S.~Wang, J.~Li, R.~Hu, T.~Zhang,
  F.~Wu, {\em et~al.}, ``Instruction tuning for large language models: A
  survey,'' {\em arXiv}, 2023.

\bibitem{liu_improving_2023}
Y.~Liu, A.~Singh, C.~D. Freeman, J.~D. Co-Reyes, and P.~J. Liu, ``Improving
  large language model fine-tuning for solving math problems,'' {\em arXiv},
  2023.

\bibitem{yang2024qwen2}
A.~Yang, B.~Zhang, B.~Hui, B.~Gao, B.~Yu, C.~Li, D.~Liu, J.~Tu, J.~Zhou,
  J.~Lin, {\em et~al.}, ``Qwen2. 5-math technical report: Toward mathematical
  expert model via self-improvement,'' {\em arXiv}, 2024.

\bibitem{ying2024internlmmath}
H.~Ying, S.~Zhang, L.~Li, Z.~Zhou, Y.~Shao, Z.~Fei, Y.~Ma, J.~Hong, K.~Liu,
  Z.~Wang, Y.~Wang, Z.~Wu, S.~Li, F.~Zhou, H.~Liu, S.~Zhang, W.~Zhang, H.~Yan,
  X.~Qiu, J.~Wang, K.~Chen, and D.~Lin, ``Internlm-math: Open math large
  language models toward verifiable reasoning,'' 2024.

\bibitem{kazemi2022lambada}
M.~Kazemi, N.~Kim, D.~Bhatia, X.~Xu, and D.~Ramachandran, ``Lambada: Backward
  chaining for automated reasoning in natural language,'' {\em arXiv}, 2022.

\bibitem{yang2022logicsolver}
Z.~Yang, J.~Qin, J.~Chen, L.~Lin, and X.~Liang, ``Logicsolver: Towards
  interpretable math word problem solving with logical prompt-enhanced
  learning,'' {\em arXiv}, 2022.

\bibitem{schick2023toolformer}
T.~Schick, J.~Dwivedi-Yu, R.~Dessì, R.~Raileanu, M.~Lomeli, L.~Zettlemoyer,
  N.~Cancedda, and T.~Scialom, ``Toolformer: Language models can teach
  themselves to use tools,'' 2023.

\bibitem{parisi2022talm}
A.~Parisi, Y.~Zhao, and N.~Fiedel, ``Talm: Tool augmented language models,''
  2022.

\bibitem{jiang2024leanreasoner}
D.~Jiang, M.~Fonseca, and S.~B. Cohen, ``Leanreasoner: Boosting complex logical
  reasoning with lean,'' {\em arXiv}, 2024.

\bibitem{paranjape2023art}
B.~Paranjape, S.~Lundberg, S.~Singh, H.~Hajishirzi, L.~Zettlemoyer, and M.~T.
  Ribeiro, ``Art: Automatic multi-step reasoning and tool-use for large
  language models,'' 2023.

\bibitem{ling2023deductive}
Z.~Ling, Y.~Fang, X.~Li, Z.~Huang, M.~Lee, R.~Memisevic, and H.~Su, ``Deductive
  verification of chain-of-thought reasoning,'' {\em arXiv}, 2023.

\bibitem{zhao2024stepwise}
Z.~Zhao, Y.~Rong, D.~Guo, E.~G{\"o}zl{\"u}kl{\"u}, E.~G{\"u}lboy, and
  E.~Kasneci, ``Stepwise self-consistent mathematical reasoning with large
  language models,'' {\em arXiv}, 2024.

\bibitem{yoran2023answering}
O.~Yoran, T.~Wolfson, B.~Bogin, U.~Katz, D.~Deutch, and J.~Berant, ``Answering
  questions by meta-reasoning over multiple chains of thought,'' {\em arXiv},
  2023.

\bibitem{yao2023tree}
S.~Yao, D.~Yu, J.~Zhao, I.~Shafran, T.~L. Griffiths, Y.~Cao, and K.~Narasimhan,
  ``Tree of thoughts: Deliberate problem solving with large language models,''
  {\em arXiv}, 2023.

\bibitem{yao2023beyond}
Y.~Yao, Z.~Li, and H.~Zhao, ``Beyond chain-of-thought, effective
  graph-of-thought reasoning in large language models,'' {\em arXiv}, 2023.

\bibitem{zelikman2022star}
E.~Zelikman, Y.~Wu, J.~Mu, and N.~Goodman, ``Star: Bootstrapping reasoning with
  reasoning,'' {\em NeurIPS}, vol.~35, pp.~15476--15488, 2022.

\bibitem{zhu2022solving}
X.~Zhu, J.~Wang, L.~Zhang, Y.~Zhang, R.~Gan, J.~Zhang, and Y.~Yang, ``Solving
  math word problems via cooperative reasoning induced language models,'' {\em
  arXiv}, 2022.

\bibitem{luo2024improve}
L.~Luo, Y.~Liu, R.~Liu, S.~Phatale, H.~Lara, Y.~Li, L.~Shu, Y.~Zhu, L.~Meng,
  J.~Sun, {\em et~al.}, ``Improve mathematical reasoning in language models by
  automated process supervision,'' {\em arXiv}, 2024.

\bibitem{nelson1980socratic}
L.~Nelson, ``The socratic method,'' vol.~2, pp.~34--38, 1980.

\bibitem{zeng2022socratic}
A.~Zeng, M.~Attarian, B.~Ichter, K.~Choromanski, A.~Wong, S.~Welker,
  F.~Tombari, A.~Purohit, M.~Ryoo, V.~Sindhwani, {\em et~al.}, ``Socratic
  models: Composing zero-shot multimodal reasoning with language,'' {\em
  arXiv}, 2022.

\bibitem{zhou2024don}
J.~P. Zhou, C.~Staats, W.~Li, C.~Szegedy, K.~Q. Weinberger, and Y.~Wu, ``Don't
  trust: Verify--grounding llm quantitative reasoning with autoformalization,''
  {\em arXiv preprint arXiv:2403.18120}, 2024.

\bibitem{dong2024formal}
K.~Dong, A.~Mahankali, and T.~Ma, ``Formal theorem proving by rewarding llms to
  decompose proofs hierarchically,'' {\em arXiv preprint arXiv:2411.01829},
  2024.

\bibitem{li2025neuro}
Z.~Li, Z.~Zhou, Y.~Yao, X.~Zhang, Y.-F. Li, C.~Cao, F.~Yang, and X.~Ma,
  ``Neuro-symbolic data generation for math reasoning,'' {\em Advances in
  Neural Information Processing Systems}, vol.~37, pp.~23488--23515, 2025.

\bibitem{wu2024enhancing}
Z.~Wu, Q.~Zeng, Z.~Zhang, Z.~Tan, C.~Shen, and M.~Jiang, ``Enhancing
  mathematical reasoning in llms by stepwise correction,'' {\em arXiv preprint
  arXiv:2410.12934}, 2024.

\bibitem{collins2024evaluating}
K.~M. Collins, A.~Q. Jiang, S.~Frieder, L.~Wong, M.~Zilka, U.~Bhatt,
  T.~Lukasiewicz, Y.~Wu, J.~B. Tenenbaum, W.~Hart, {\em et~al.}, ``Evaluating
  language models for mathematics through interactions,'' {\em Proceedings of
  the National Academy of Sciences}, vol.~121, no.~24, p.~e2318124121, 2024.

\bibitem{chen2024controlmath}
N.~Chen, N.~Wu, J.~Chang, and J.~Li, ``Controlmath: Controllable data
  generation promotes math generalist models,'' {\em arXiv}, 2024.

\bibitem{olausson2023linc}
T.~X. Olausson, A.~Gu, B.~Lipkin, C.~E. Zhang, A.~Solar-Lezama, J.~B.
  Tenenbaum, and R.~Levy, ``Linc: A neurosymbolic approach for logical
  reasoning by combining language models with first-order logic provers,'' {\em
  arXiv}, 2023.

\bibitem{yin2024mumath}
S.~Yin, W.~You, Z.~Ji, G.~Zhong, and J.~Bai, ``Mumath-code: Combining tool-use
  large language models with multi-perspective data augmentation for
  mathematical reasoning,'' {\em arXiv}, 2024.

\bibitem{yue2023mammoth}
X.~Yue, X.~Qu, G.~Zhang, Y.~Fu, W.~Huang, H.~Sun, Y.~Su, and W.~Chen,
  ``Mammoth: Building math generalist models through hybrid instruction
  tuning,'' {\em arXiv}, 2023.

\bibitem{pitis_boosted_2023}
S.~Pitis, M.~R. Zhang, A.~Wang, and J.~Ba, ``Boosted prompt ensembles for large
  language models,'' {\em arXiv}, 2023.

\bibitem{miao2023selfcheck}
N.~Miao, Y.~W. Teh, and T.~Rainforth, ``Selfcheck: Using llms to zero-shot
  check their own step-by-step reasoning,'' {\em arXiv}, 2023.

\bibitem{naik2023diversity}
R.~Naik, V.~Chandrasekaran, M.~Yuksekgonul, H.~Palangi, and B.~Nushi,
  ``Diversity of thought improves reasoning abilities of large language
  models,'' {\em arXiv}, 2023.

\bibitem{hao2023reasoning}
S.~Hao, Y.~Gu, H.~Ma, J.~J. Hong, Z.~Wang, D.~Z. Wang, and Z.~Hu, ``Reasoning
  with language model is planning with world model,'' {\em arXiv}, 2023.

\bibitem{zhou2023language}
A.~Zhou, K.~Yan, M.~Shlapentokh-Rothman, H.~Wang, and Y.-X. Wang, ``Language
  agent tree search unifies reasoning acting and planning in language models,''
  {\em arXiv}, 2023.

\bibitem{liu2023llm+}
B.~Liu, Y.~Jiang, X.~Zhang, Q.~Liu, S.~Zhang, J.~Biswas, and P.~Stone, ``Llm+
  p: Empowering large language models with optimal planning proficiency,'' {\em
  arXiv}, 2023.

\bibitem{dagan2023dynamic}
G.~Dagan, F.~Keller, and A.~Lascarides, ``Dynamic planning with a llm,'' {\em
  arXiv}, 2023.

\bibitem{zhou2023isr}
Z.~Zhou, J.~Song, K.~Yao, Z.~Shu, and L.~Ma, ``Isr-llm: Iterative self-refined
  large language model for long-horizon sequential task planning,'' {\em
  arXiv}, 2023.

\bibitem{cobbe2021training}
K.~Cobbe, V.~Kosaraju, M.~Bavarian, M.~Chen, H.~Jun, L.~Kaiser, M.~Plappert,
  J.~Tworek, J.~Hilton, R.~Nakano, C.~Hesse, and J.~Schulman, ``Training
  verifiers to solve math word problems,'' 2021.

\bibitem{khalifa2023discriminator}
M.~Khalifa, L.~Logeswaran, M.~Lee, H.~Lee, and L.~Wang, ``Discriminator-guided
  multi-step reasoning with language models,'' {\em arXiv}, 2023.

\bibitem{shinn2023reflexion}
N.~Shinn, F.~Cassano, A.~Gopinath, K.~R. Narasimhan, and S.~Yao, ``Reflexion:
  Language agents with verbal reinforcement learning,'' in {\em NeurIPS}, 2023.

\bibitem{zhang2023cumulative}
Y.~Zhang, J.~Yang, Y.~Yuan, and A.~C.-C. Yao, ``Cumulative reasoning with large
  language models,'' {\em arXiv}, 2023.

\bibitem{zhang2023interpretable}
M.~Zhang, Z.~Wang, Z.~Yang, W.~Feng, and A.~Lan, ``Interpretable math word
  problem solution generation via step-by-step planning,'' {\em arXiv}, 2023.

\bibitem{hosseini2024v}
A.~Hosseini, X.~Yuan, N.~Malkin, A.~Courville, A.~Sordoni, and R.~Agarwal,
  ``V-star: Training verifiers for self-taught reasoners,'' {\em arXiv}, 2024.

\bibitem{zelikman2024quiet}
E.~Zelikman, G.~Harik, Y.~Shao, V.~Jayasiri, N.~Haber, and N.~D. Goodman,
  ``Quiet-star: Language models can teach themselves to think before
  speaking,'' {\em arXiv}, 2024.

\bibitem{luong2024reft}
T.~Q. Luong, X.~Zhang, Z.~Jie, P.~Sun, X.~Jin, and H.~Li, ``Reft: Reasoning
  with reinforced fine-tuning,'' {\em arXiv}, 2024.

\bibitem{kumar2024training}
A.~Kumar, V.~Zhuang, R.~Agarwal, Y.~Su, J.~D. Co-Reyes, A.~Singh, K.~Baumli,
  S.~Iqbal, C.~Bishop, R.~Roelofs, {\em et~al.}, ``Training language models to
  self-correct via reinforcement learning,'' {\em arXiv}, 2024.

\bibitem{ma2023let}
Q.~Ma, H.~Zhou, T.~Liu, J.~Yuan, P.~Liu, Y.~You, and H.~Yang, ``Let's reward
  step by step: Step-level reward model as the navigators for reasoning,'' {\em
  arXiv}, 2023.

\bibitem{zhang2024accessing}
D.~Zhang, X.~Huang, D.~Zhou, Y.~Li, and W.~Ouyang, ``Accessing gpt-4 level
  mathematical olympiad solutions via monte carlo tree self-refine with llama-3
  8b,'' {\em arXiv}, 2024.

\bibitem{zhao2024marco}
Y.~Zhao, H.~Yin, B.~Zeng, H.~Wang, T.~Shi, C.~Lyu, L.~Wang, W.~Luo, and
  K.~Zhang, ``Marco-o1: Towards open reasoning models for open-ended
  solutions,'' {\em arXiv}, 2024.

\bibitem{lai2024step}
X.~Lai, Z.~Tian, Y.~Chen, S.~Yang, X.~Peng, and J.~Jia, ``Step-dpo: Step-wise
  preference optimization for long-chain reasoning of llms,'' {\em arXiv},
  2024.

\bibitem{deng2024flow}
Y.~Deng and P.~Mineiro, ``Flow-dpo: Improving llm mathematical reasoning
  through online multi-agent learning,'' {\em arXiv}, 2024.

\bibitem{ding2024boosting}
Y.~Ding, H.~Hu, J.~Zhou, Q.~Chen, B.~Jiang, and L.~He, ``Boosting large
  language models with socratic method for conversational mathematics
  teaching,'' in {\em CIKM}, pp.~3730--3735, 2024.

\bibitem{macina2023mathdial}
J.~Macina, N.~Daheim, S.~P. Chowdhury, T.~Sinha, M.~Kapur, I.~Gurevych, and
  M.~Sachan, ``Mathdial: A dialogue tutoring dataset with rich pedagogical
  properties grounded in math reasoning problems,'' {\em arXiv}, 2023.

\bibitem{qwq-32b-preview}
Q.~Team, ``Qwq: Reflect deeply on the boundaries of the unknown,'' 2024.

\bibitem{openai_o1_system_card}
{OpenAI}, ``{OpenAI O1 System Card},'' 2024.

\bibitem{hurst2024gpt}
A.~Hurst, A.~Lerer, A.~P. Goucher, A.~Perelman, A.~Ramesh, A.~Clark, A.~Ostrow,
  A.~Welihinda, A.~Hayes, A.~Radford, {\em et~al.}, ``Gpt-4o system card,''
  {\em arXiv}, 2024.

\bibitem{Qwen2-VL}
P.~Wang, S.~Bai, S.~Tan, S.~Wang, Z.~Fan, J.~Bai, K.~Chen, X.~Liu, J.~Wang,
  W.~Ge, Y.~Fan, K.~Dang, M.~Du, X.~Ren, R.~Men, D.~Liu, C.~Zhou, J.~Zhou, and
  J.~Lin, ``Qwen2-vl: Enhancing vision-language model's perception of the world
  at any resolution,'' {\em arXiv}, 2024.

\bibitem{xu2024llava}
G.~Xu, P.~Jin, L.~Hao, Y.~Song, L.~Sun, and L.~Yuan, ``Llava-o1: Let vision
  language models reason step-by-step,'' {\em arXiv}, 2024.

\bibitem{team2023gemini}
G.~Team, R.~Anil, S.~Borgeaud, J.-B. Alayrac, J.~Yu, R.~Soricut, J.~Schalkwyk,
  A.~M. Dai, A.~Hauth, K.~Millican, {\em et~al.}, ``Gemini: a family of highly
  capable multimodal models,'' {\em arXiv}, 2023.

\bibitem{glm2024chatglm}
T.~GLM, A.~Zeng, B.~Xu, B.~Wang, C.~Zhang, D.~Yin, D.~Zhang, D.~Rojas, G.~Feng,
  H.~Zhao, {\em et~al.}, ``Chatglm: A family of large language models from
  glm-130b to glm-4 all tools,'' {\em arXiv}, 2024.

\bibitem{xiang2024atomthink}
K.~Xiang, Z.~Liu, Z.~Jiang, Y.~Nie, R.~Huang, H.~Fan, H.~Li, W.~Huang, Y.~Zeng,
  J.~Han, {\em et~al.}, ``Atomthink: A slow thinking framework for multimodal
  mathematical reasoning,'' {\em arXiv}, 2024.

\bibitem{shihu2024mathllava}
W.~Shi, Z.~Hu, Y.~Bin, J.~Liu, Y.~Yang, S.-K. Ng, L.~Bing, and R.~K.-W. Lee,
  ``Math-llava: Bootstrapping mathematical reasoning for multimodal large
  language models,'' 2024.

\bibitem{anonymous2024diving}
Anonymous, ``Diving into self-evolve training for multimodal reasoning,'' in
  {\em Submitted to The Thirteenth International Conference on Learning
  Representations}, 2024.

\bibitem{wang2024understanding}
Y.~wang and Y.~Fu, ``Understanding, abstracting and checking: Evoking
  complicated multimodal reasoning in {LMM}s,'' 2024.

\bibitem{forbes2017verb}
M.~Forbes and Y.~Choi, ``Verb physics: Relative physical knowledge of actions
  and objects,'' in {\em ACL}, pp.~266--276, 2017.

\bibitem{spithourakis2016numerically}
G.~Spithourakis, I.~Augenstein, and S.~Riedel, ``Numerically grounded language
  models for semantic error correction,'' in {\em EMNLP}, pp.~987--992, 2016.

\bibitem{elazar2019large}
Y.~Elazar, A.~Mahabal, D.~Ramachandran, T.~Bedrax-Weiss, and D.~Roth, ``How
  large are lions? inducing distributions over quantitative attributes,'' in
  {\em ACL}, pp.~3973--3983, 2019.

\bibitem{ling2017program}
W.~Ling, D.~Yogatama, C.~Dyer, and P.~Blunsom, ``Program induction by rationale
  generation: Learning to solve and explain algebraic word problems,'' in {\em
  ACL}, pp.~158--167, 2017.

\bibitem{patel2021nlp}
A.~Patel, S.~Bhattamishra, and N.~Goyal, ``Are nlp models really able to solve
  simple math word problems?,'' in {\em NAACL}, pp.~2080--2094, 2021.

\bibitem{upadhyay2015draw}
S.~Upadhyay and M.-W. Chang, ``Draw: A challenging and diverse algebra word
  problem set,'' tech. rep., Citeseer, 2015.

\bibitem{qin2020semantically}
J.~Qin, L.~Lin, X.~Liang, R.~Zhang, and L.~Lin, ``Semantically-aligned
  universal tree-structured solver for math word problems,'' in {\em EMNLP},
  pp.~3780--3789, 2020.

\bibitem{alghamdi2022armath}
R.~Alghamdi, Z.~Liang, and X.~Zhang, ``Armath: a dataset for solving arabic
  math word problems,'' in {\em LREC}, pp.~351--362, 2022.

\bibitem{kalyan2021much}
A.~Kalyan, A.~Kumar, A.~Chandrasekaran, A.~Sabharwal, and P.~Clark, ``How much
  coffee was consumed during emnlp 2019? fermi problems: A new reasoning
  challenge for ai,'' in {\em EMNLP}, pp.~7318--7328, 2021.

\bibitem{wang2023generative}
Z.~Wang, R.~Xia, and P.~Liu, ``Generative ai for math: Part i--mathpile: A
  billion-token-scale pretraining corpus for math,'' {\em arXiv}, 2023.

\bibitem{paster2023openwebmath}
K.~Paster, M.~D. Santos, Z.~Azerbayev, and J.~Ba, ``Openwebmath: An open
  dataset of high-quality mathematical web text,'' 2023.

\bibitem{amini2019mathqa}
A.~Amini, S.~Gabriel, S.~Lin, R.~Koncel-Kedziorski, Y.~Choi, and H.~Hajishirzi,
  ``Mathqa: Towards interpretable math word problem solving with
  operation-based formalisms,'' in {\em NAACL}, pp.~2357--2367, 2019.

\bibitem{chen2021finqa}
Z.~Chen, W.~Chen, C.~Smiley, S.~Shah, I.~Borova, D.~Langdon, R.~Moussa,
  M.~Beane, T.-H. Huang, B.~R. Routledge, {\em et~al.}, ``Finqa: A dataset of
  numerical reasoning over financial data,'' in {\em EMNLP}, pp.~3697--3711,
  2021.

\bibitem{grabowski2015four}
A.~Grabowski, A.~Korni{\l}owicz, and A.~Naumowicz, ``Four decades of mizar:
  Foreword,'' {\em Journal of Automated Reasoning}, vol.~55, pp.~191--198,
  2015.

\bibitem{kaliszyk2016holstep}
C.~Kaliszyk, F.~Chollet, and C.~Szegedy, ``Holstep: A machine learning dataset
  for higher-order logic theorem proving,'' in {\em ICLR}, 2017.

\bibitem{bansal2019holist}
K.~Bansal, S.~Loos, M.~Rabe, C.~Szegedy, and S.~Wilcox, ``Holist: An
  environment for machine learning of higher order logic theorem proving,'' in
  {\em ICML}, pp.~454--463, 2019.

\bibitem{li2020isarstep}
W.~Li, L.~Yu, Y.~Wu, and L.~C. Paulson, ``Isarstep: a benchmark for high-level
  mathematical reasoning,'' in {\em ICLR}, 2021.

\bibitem{mishra2022numglue}
S.~Mishra, A.~Mitra, N.~Varshney, B.~Sachdeva, P.~Clark, C.~Baral, and
  A.~Kalyan, ``Numglue: A suite of fundamental yet challenging mathematical
  reasoning tasks,'' in {\em ACL}, pp.~3505--3523, 2022.

\bibitem{huang2016well}
D.~Huang, S.~Shi, C.-Y. Lin, J.~Yin, and W.-Y. Ma, ``How well do computers
  solve math word problems? large-scale dataset construction and evaluation,''
  in {\em ACL}, pp.~887--896, 2016.

\bibitem{koncel2016mawps}
R.~Koncel-Kedziorski, S.~Roy, A.~Amini, N.~Kushman, and H.~Hajishirzi, ``Mawps:
  A math word problem repository,'' in {\em NAACL}, pp.~1152--1157, 2016.

\bibitem{miao2020diverse}
S.-Y. Miao, C.-C. Liang, and K.-Y. Su, ``A diverse corpus for evaluating and
  developing english math word problem solvers,'' in {\em ACL}, pp.~975--984,
  2020.

\bibitem{shi2022language}
F.~Shi, M.~Suzgun, M.~Freitag, X.~Wang, S.~Srivats, S.~Vosoughi, H.~W. Chung,
  Y.~Tay, S.~Ruder, D.~Zhou, {\em et~al.}, ``Language models are multilingual
  chain-of-thought reasoners,'' in {\em ICLR}, 2022.

\bibitem{saxton2018analysing}
D.~Saxton, E.~Grefenstette, F.~Hill, and P.~Kohli, ``Analysing mathematical
  reasoning abilities of neural models,'' in {\em ICLR}, 2019.

\bibitem{zhong2023agieval}
W.~Zhong, R.~Cui, Y.~Guo, Y.~Liang, S.~Lu, Y.~Wang, A.~Saied, W.~Chen, and
  N.~Duan, ``Agieval: A human-centric benchmark for evaluating foundation
  models,'' 2023.

\bibitem{zhao2024exploring}
J.~Zhao, J.~Tong, Y.~Mou, M.~Zhang, Q.~Zhang, and X.-J. Huang, ``Exploring the
  compositional deficiency of large language models in mathematical reasoning
  through trap problems,'' in {\em EMNLP}, pp.~16361--16376, 2024.

\bibitem{lu2023mathvista}
P.~Lu, H.~Bansal, T.~Xia, J.~Liu, C.~Li, H.~Hajishirzi, H.~Cheng, K.-W. Chang,
  M.~Galley, and J.~Gao, ``Mathvista: Evaluating mathematical reasoning of
  foundation models in visual contexts,'' {\em arXiv}, 2023.

\bibitem{wang2024measuring}
K.~Wang, J.~Pan, W.~Shi, Z.~Lu, M.~Zhan, and H.~Li, ``Measuring multimodal
  mathematical reasoning with math-vision dataset,'' {\em arXiv}, 2024.

\bibitem{liu2024cmm}
W.~Liu, Q.~Pan, Y.~Zhang, Z.~Liu, J.~Wu, J.~Zhou, A.~Zhou, Q.~Chen, B.~Jiang,
  and L.~He, ``Cmm-math: A chinese multimodal math dataset to evaluate and
  enhance the mathematics reasoning of large multimodal models,'' {\em arXiv},
  2024.

\bibitem{zhang2025mathverse}
R.~Zhang, D.~Jiang, Y.~Zhang, H.~Lin, Z.~Guo, P.~Qiu, A.~Zhou, P.~Lu, K.-W.
  Chang, Y.~Qiao, {\em et~al.}, ``Mathverse: Does your multi-modal llm truly
  see the diagrams in visual math problems?,'' in {\em ECCV}, pp.~169--186,
  2025.

\bibitem{chen2021geoqa}
J.~Chen, J.~Tang, J.~Qin, X.~Liang, L.~Liu, E.~P. Xing, and L.~Lin, ``Geoqa: A
  geometric question answering benchmark towards multimodal numerical
  reasoning,'' {\em arXiv}, 2021.

\bibitem{chen-etal-2024-m3cot}
Q.~Chen, L.~Qin, J.~Zhang, Z.~Chen, X.~Xu, and W.~Che, ``{M}$^3${C}o{T}: A
  novel benchmark for multi-domain multi-step multi-modal chain-of-thought,''
  in {\em ACL}, pp.~8199--8221, 2024.

\bibitem{xia2024evaluating}
S.~Xia, X.~Li, Y.~Liu, T.~Wu, and P.~Liu, ``Evaluating mathematical reasoning
  beyond accuracy,'' {\em arXiv}, 2024.

\bibitem{yue2024mmmumassivemultidisciplinemultimodal}
X.~Yue, Y.~Ni, K.~Zhang, T.~Zheng, R.~Liu, G.~Zhang, S.~Stevens, D.~Jiang,
  W.~Ren, Y.~Sun, C.~Wei, B.~Yu, R.~Yuan, R.~Sun, M.~Yin, B.~Zheng, Z.~Yang,
  Y.~Liu, W.~Huang, H.~Sun, Y.~Su, and W.~Chen, ``Mmmu: A massive
  multi-discipline multimodal understanding and reasoning benchmark for expert
  agi,'' 2024.

\bibitem{qiao2024wemathdoeslargemultimodal}
R.~Qiao, Q.~Tan, G.~Dong, M.~Wu, C.~Sun, X.~Song, Z.~GongQue, S.~Lei, Z.~Wei,
  M.~Zhang, R.~Qiao, Y.~Zhang, X.~Zong, Y.~Xu, M.~Diao, Z.~Bao, C.~Li, and
  H.~Zhang, ``We-math: Does your large multimodal model achieve human-like
  mathematical reasoning?,'' 2024.

\bibitem{chernyshev2024umathuniversitylevelbenchmarkevaluating}
K.~Chernyshev, V.~Polshkov, E.~Artemova, A.~Myasnikov, V.~Stepanov,
  A.~Miasnikov, and S.~Tilga, ``U-math: A university-level benchmark for
  evaluating mathematical skills in llms,'' 2024.

\bibitem{zheng2021minif2f}
K.~Zheng, J.~M. Han, and S.~Polu, ``minif2f: a cross-system benchmark for
  formal olympiad-level mathematics,'' in {\em ICLR}, 2023.

\bibitem{DBLP:conf/emnlp/HanS0QRZCPQBSWS24}
S.~Han, H.~Schoelkopf, Y.~Zhao, Z.~Qi, M.~Riddell, W.~Zhou, J.~Coady, D.~Peng,
  Y.~Qiao, L.~Benson, L.~Sun, A.~Wardle{-}Solano, H.~Szab{\'{o}}, E.~Zubova,
  M.~Burtell, J.~Fan, Y.~Liu, B.~Wong, M.~Sailor, A.~Ni, L.~Nan, J.~Kasai,
  T.~Yu, R.~Zhang, A.~R. Fabbri, W.~Kryscinski, S.~Yavuz, Y.~Liu, X.~V. Lin,
  S.~Joty, Y.~Zhou, C.~Xiong, R.~Ying, A.~Cohan, and D.~Radev, ``{FOLIO:}
  natural language reasoning with first-order logic,'' in {\em Proceedings of
  the 2024 Conference on Empirical Methods in Natural Language Processing,
  {EMNLP} 2024, Miami, FL, USA, November 12-16, 2024} (Y.~Al{-}Onaizan,
  M.~Bansal, and Y.~Chen, eds.), pp.~22017--22031, Association for
  Computational Linguistics, 2024.

\bibitem{roy2018mapping}
S.~Roy and D.~Roth, ``Mapping to declarative knowledge for word problem
  solving,'' {\em TACL}, vol.~6, pp.~159--172, 2018.

\bibitem{austin2021program}
J.~Austin, A.~Odena, M.~Nye, M.~Bosma, H.~Michalewski, D.~Dohan, E.~Jiang,
  C.~Cai, M.~Terry, Q.~Le, {\em et~al.}, ``Program synthesis with large
  language models,'' {\em arXiv}, 2021.

\bibitem{li2023camel}
G.~Li, H.~A. A.~K. Hammoud, H.~Itani, D.~Khizbullin, and B.~Ghanem, ``Camel:
  Communicative agents for" mind" exploration of large language model
  society,'' in {\em NeurIPS}, 2023.

\bibitem{DBLP:conf/iclr/LightmanKBEBLLS24}
H.~Lightman, V.~Kosaraju, Y.~Burda, H.~Edwards, B.~Baker, T.~Lee, J.~Leike,
  J.~Schulman, I.~Sutskever, and K.~Cobbe, ``Let's verify step by step,'' in
  {\em ICLR}, 2024.

\bibitem{mishra2022lila}
S.~Mishra, M.~Finlayson, P.~Lu, L.~Tang, S.~Welleck, C.~Baral, T.~Rajpurohit,
  O.~Tafjord, A.~Sabharwal, P.~Clark, {\em et~al.}, ``Lila: A unified benchmark
  for mathematical reasoning,'' in {\em EMNLP}, pp.~5807--5832, 2022.

\bibitem{kim2022ept}
B.~Kim, K.~S. Ki, S.~Rhim, and G.~Gweon, ``Ept-x: An expression-pointer
  transformer model that generates explanations for numbers,'' in {\em ACL},
  pp.~4442--4458, 2022.

\bibitem{welleck2022naturalprover}
S.~Welleck, J.~Liu, X.~Lu, H.~Hajishirzi, and Y.~Choi, ``Naturalprover:
  Grounded mathematical proof generation with language models,'' {\em NeurIPS},
  vol.~35, pp.~4913--4927, 2022.

\bibitem{MATH}
D.~Hendrycks, C.~Burns, S.~Kadavath, A.~Arora, S.~Basart, E.~Tang, D.~Song, and
  J.~Steinhardt, ``Measuring mathematical problem solving with the math
  dataset,'' {\em arXiv preprint arXiv:2103.03874}, 2021.

\bibitem{trinh2024solving}
T.~H. Trinh, Y.~Wu, Q.~V. Le, H.~He, and T.~Luong, ``Solving olympiad geometry
  without human demonstrations,'' {\em Nature}, vol.~625, no.~7995,
  pp.~476--482, 2024.

\bibitem{patil2025advancing}
A.~Patil, ``Advancing reasoning in large language models: Promising methods and
  approaches,'' {\em arXiv preprint arXiv:2502.03671}, 2025.

\bibitem{gpt4}
OpenAI, ``{GPT4} technical report,'' {\em arXiv preprint arXiv:2303.08774},
  2023.

\bibitem{gemini}
R.~Anil, S.~Borgeaud, Y.~Wu, J.~Alayrac, J.~Yu, R.~Soricut, J.~Schalkwyk, A.~M.
  Dai, A.~Hauth, K.~Millican, D.~Silver, S.~Petrov, M.~Johnson, I.~Antonoglou,
  J.~Schrittwieser, A.~Glaese, J.~Chen, E.~Pitler, T.~P. Lillicrap,
  A.~Lazaridou, O.~Firat, J.~Molloy, M.~Isard, P.~R. Barham, T.~Hennigan,
  B.~Lee, F.~Viola, M.~Reynolds, Y.~Xu, R.~Doherty, E.~Collins, C.~Meyer,
  E.~Rutherford, E.~Moreira, K.~Ayoub, M.~Goel, G.~Tucker, E.~Piqueras,
  M.~Krikun, I.~Barr, N.~Savinov, I.~Danihelka, B.~Roelofs, A.~White,
  A.~Andreassen, T.~von Glehn, L.~Yagati, M.~Kazemi, L.~Gonzalez, M.~Khalman,
  J.~Sygnowski, and et~al., ``Gemini: {A} family of highly capable multimodal
  models,'' {\em CoRR}, vol.~abs/2312.11805, 2023.

\bibitem{deepseek-llm}
DeepSeek-AI, ``Deepseek {LLM:} scaling open-source language models with
  longtermism,'' {\em CoRR}, vol.~abs/2401.02954, 2024.

\bibitem{joulin2016fasttext}
A.~Joulin, E.~Grave, P.~Bojanowski, M.~Douze, H.~J{\'e}gou, and T.~Mikolov,
  ``Fasttext. zip: Compressing text classification models,'' {\em arXiv
  preprint arXiv:1612.03651}, 2016.

\bibitem{deepseek-coder}
D.~Guo, Q.~Zhu, D.~Yang, Z.~Xie, K.~Dong, W.~Zhang, G.~Chen, X.~Bi, Y.~Wu,
  Y.~K. Li, F.~Luo, Y.~Xiong, and W.~Liang, ``Deepseek-coder: When the large
  language model meets programming -- the rise of code intelligence,'' 2024.

\bibitem{minerva}
A.~Lewkowycz, A.~Andreassen, D.~Dohan, E.~Dyer, H.~Michalewski, V.~Ramasesh,
  A.~Slone, C.~Anil, I.~Schlag, T.~Gutman-Solo, Y.~Wu, B.~Neyshabur,
  G.~Gur-Ari, and V.~Misra, ``Solving quantitative reasoning problems with
  language models,'' {\em arXiv preprint arXiv:2206.14858}, 2022.

\bibitem{wei2023cmath}
T.~Wei, J.~Luan, W.~Liu, S.~Dong, and B.~Wang, ``Cmath: Can your language model
  pass chinese elementary school math test?,'' 2023.

\bibitem{agieval}
W.~Zhong, R.~Cui, Y.~Guo, Y.~Liang, S.~Lu, Y.~Wang, A.~Saied, W.~Chen, and
  N.~Duan, ``{AGIEval}: {A} human-centric benchmark for evaluating foundation
  models,'' {\em CoRR}, vol.~abs/2304.06364, 2023.

\bibitem{cot}
J.~Wei, X.~Wang, D.~Schuurmans, M.~Bosma, B.~Ichter, F.~Xia, E.~H. Chi, Q.~V.
  Le, and D.~Zhou, ``Chain-of-thought prompting elicits reasoning in large
  language models,'' in {\em NeurIPS}, 2022.

\bibitem{pot}
W.~Chen, X.~Ma, X.~Wang, and W.~W. Cohen, ``Program of thoughts prompting:
  Disentangling computation from reasoning for numerical reasoning tasks,''
  {\em CoRR}, vol.~abs/2211.12588, 2022.

\bibitem{pal}
L.~Gao, A.~Madaan, S.~Zhou, U.~Alon, P.~Liu, Y.~Yang, J.~Callan, and G.~Neubig,
  ``{PAL:} program-aided language models,'' in {\em International Conference on
  Machine Learning, {ICML} 2023, 23-29 July 2023, Honolulu, Hawaii, {USA}}
  (A.~Krause, E.~Brunskill, K.~Cho, B.~Engelhardt, S.~Sabato, and J.~Scarlett,
  eds.), vol.~202 of {\em Proceedings of Machine Learning Research},
  pp.~10764--10799, {PMLR}, 2023.

\bibitem{tora}
Z.~Gou, Z.~Shao, Y.~Gong, Y.~Shen, Y.~Yang, M.~Huang, N.~Duan, and W.~Chen,
  ``Tora: {A} tool-integrated reasoning agent for mathematical problem
  solving,'' {\em CoRR}, vol.~abs/2309.17452, 2023.

\bibitem{dpo}
R.~Rafailov, A.~Sharma, E.~Mitchell, S.~Ermon, C.~D. Manning, and C.~Finn,
  ``Direct preference optimization: Your language model is secretly a reward
  model,'' 2023.

\bibitem{schulman2017proximal}
J.~Schulman, F.~Wolski, P.~Dhariwal, A.~Radford, and O.~Klimov, ``Proximal
  policy optimization algorithms,'' {\em arXiv}, 2017.

\bibitem{gsm8k}
K.~Cobbe, V.~Kosaraju, M.~Bavarian, M.~Chen, H.~Jun, L.~Kaiser, M.~Plappert,
  J.~Tworek, J.~Hilton, R.~Nakano, {\em et~al.}, ``Training verifiers to solve
  math word problems,'' {\em arXiv preprint arXiv:2110.14168}, 2021.

\bibitem{mistral}
A.~Q. Jiang, A.~Sablayrolles, A.~Mensch, C.~Bamford, D.~S. Chaplot, D.~d.~l.
  Casas, F.~Bressand, G.~Lengyel, G.~Lample, L.~Saulnier, {\em et~al.},
  ``Mistral 7b,'' {\em arXiv preprint arXiv:2310.06825}, 2023.

\bibitem{llemma}
Z.~Azerbayev, H.~Schoelkopf, K.~Paster, M.~D. Santos, S.~McAleer, A.~Q. Jiang,
  J.~Deng, S.~Biderman, and S.~Welleck, ``Llemma: An open language model for
  mathematics,'' {\em arXiv preprint arXiv:2310.10631}, 2023.

\bibitem{minif2f}
K.~Zheng, J.~M. Han, and S.~Polu, ``Minif2f: a cross-system benchmark for
  formal olympiad-level mathematics,'' {\em arXiv preprint arXiv:2109.00110},
  2021.

\bibitem{mmlu}
D.~Hendrycks, C.~Burns, S.~Basart, A.~Zou, M.~Mazeika, D.~Song, and
  J.~Steinhardt, ``Measuring massive multitask language understanding,'' {\em
  arXiv preprint arXiv:2009.03300}, 2020.

\bibitem{bbh}
M.~Suzgun, N.~Scales, N.~Sch{\"a}rli, S.~Gehrmann, Y.~Tay, H.~W. Chung,
  A.~Chowdhery, Q.~V. Le, E.~H. Chi, D.~Zhou, {\em et~al.}, ``Challenging
  big-bench tasks and whether chain-of-thought can solve them,'' {\em arXiv
  preprint arXiv:2210.09261}, 2022.

\bibitem{codex}
M.~Chen, J.~Tworek, H.~Jun, Q.~Yuan, H.~P. de~Oliveira~Pinto, J.~Kaplan,
  H.~Edwards, Y.~Burda, N.~Joseph, G.~Brockman, A.~Ray, R.~Puri, G.~Krueger,
  M.~Petrov, H.~Khlaaf, G.~Sastry, P.~Mishkin, B.~Chan, S.~Gray, N.~Ryder,
  M.~Pavlov, A.~Power, L.~Kaiser, M.~Bavarian, C.~Winter, P.~Tillet, F.~P.
  Such, D.~Cummings, M.~Plappert, F.~Chantzis, E.~Barnes, A.~Herbert-Voss,
  W.~H. Guss, A.~Nichol, A.~Paino, N.~Tezak, J.~Tang, I.~Babuschkin, S.~Balaji,
  S.~Jain, W.~Saunders, C.~Hesse, A.~N. Carr, J.~Leike, J.~Achiam, V.~Misra,
  E.~Morikawa, A.~Radford, M.~Knight, M.~Brundage, M.~Murati, K.~Mayer,
  P.~Welinder, B.~McGrew, D.~Amodei, S.~McCandlish, I.~Sutskever, and
  W.~Zaremba, ``Evaluating large language models trained on code,'' {\em arXiv
  preprint arXiv:2107.03374}, 2021.

\bibitem{mbpp}
J.~Austin, A.~Odena, M.~Nye, M.~Bosma, H.~Michalewski, D.~Dohan, E.~Jiang,
  C.~Cai, M.~Terry, Q.~Le, {\em et~al.}, ``Program synthesis with large
  language models,'' {\em arXiv preprint arXiv:2108.07732}, 2021.

\bibitem{wizardmath}
H.~Luo, Q.~Sun, C.~Xu, P.~Zhao, J.~Lou, C.~Tao, X.~Geng, Q.~Lin, S.~Chen, and
  D.~Zhang, ``Wizardmath: Empowering mathematical reasoning for large language
  models via reinforced evol-instruct,'' {\em arXiv preprint arXiv:2308.09583},
  2023.

\bibitem{metamath}
L.~Yu, W.~Jiang, H.~Shi, J.~Yu, Z.~Liu, Y.~Zhang, J.~T. Kwok, Z.~Li, A.~Weller,
  and W.~Liu, ``Metamath: Bootstrap your own mathematical questions for large
  language models,'' {\em CoRR}, vol.~abs/2309.12284, 2023.

\bibitem{MathInstruct}
X.~Yue, X.~Qu, G.~Zhang, Y.~Fu, W.~Huang, H.~Sun, Y.~Su, and W.~Chen,
  ``Mammoth: Building math generalist models through hybrid instruction
  tuning,'' {\em CoRR}, vol.~abs/2309.05653, 2023.

\bibitem{berg-kirkpatrick-spokoyny-2020-empirical}
T.~Berg-Kirkpatrick and D.~Spokoyny, ``An empirical investigation of
  contextualized number prediction,'' in {\em EMNLP}, pp.~4754--4764, Nov.
  2020.

\bibitem{he2023socreval}
H.~He, H.~Zhang, and D.~Roth, ``Socreval: Large language models with the
  socratic method for reference-free reasoning evaluation,'' {\em arXiv}, 2023.

\bibitem{huang2018gamepad}
D.~Huang, P.~Dhariwal, D.~Song, and I.~Sutskever, ``Gamepad: A learning
  environment for theorem proving,'' in {\em ICLR}, 2019.

\bibitem{touvron2023llama2}
H.~Touvron, L.~Martin, K.~Stone, P.~Albert, A.~Almahairi, Y.~Babaei,
  N.~Bashlykov, S.~Batra, P.~Bhargava, S.~Bhosale, {\em et~al.}, ``Llama 2:
  Open foundation and fine-tuned chat models,'' {\em arXiv}, 2023.

\bibitem{wu2022timesnet}
H.~Wu, T.~Hu, Y.~Liu, H.~Zhou, J.~Wang, and M.~Long, ``{TimesNet: Temporal
  2D-Variation Modeling for General Time Series Analysis},'' in {\em ICLR},
  2022.

\bibitem{garza2023timegpt}
A.~Garza and M.~Mergenthaler-Canseco, ``{TimeGPT-1},'' {\em arXiv preprint
  arXiv:2310.03589}, 2023.

\bibitem{yu2023temporal}
X.~Yu, Z.~Chen, Y.~Ling, S.~Dong, Z.~Liu, and Y.~Lu, ``{Temporal Data Meets
  LLM--Explainable Financial Time Series Forecasting},'' {\em arXiv preprint
  arXiv:2306.11025}, 2023.

\bibitem{lopez2023can}
A.~Lopez-Lira and Y.~Tang, ``{Can ChatGPT Forecast Stock Price Movements?
  Return Predictability and Large Language Models},'' {\em arXiv preprint
  arXiv:2304.07619}, 2023.

\bibitem{li2023frozen}
J.~Li, C.~Liu, S.~Cheng, R.~Arcucci, and S.~Hong, ``{Frozen Language Model
  Helps ECG Zero-Shot Learning},'' in {\em MIDL}, 2023.

\bibitem{jiang2023health}
L.~Y. Jiang, X.~C. Liu, N.~P. Nejatian, M.~Nasir-Moin, D.~Wang, A.~Abidin, {\em
  et~al.}, ``{Health system-scale language models are all-purpose prediction
  engines},'' {\em Nature}, pp.~1--6, 2023.

\bibitem{liu2024spatialtemporal}
C.~Liu, S.~Yang, Q.~Xu, Z.~Li, C.~Long, Z.~Li, and R.~Zhao, ``{Spatial-Temporal
  Large Language Model for Traffic Prediction},'' 2024.

\bibitem{wang2023i}
X.~Wang, M.~Fang, Z.~Zeng, and T.~Cheng, ``{Where Would I Go Next? Large
  Language Models as Human Mobility Predictors},'' 2023.

\bibitem{kim2021reversible}
T.~Kim, J.~Kim, Y.~Tae, C.~Park, J.-H. Choi, and J.~Choo, ``{Reversible
  Instance Normalization for Accurate Time-Series Forecasting against
  Distribution Shift},'' in {\em ICLR}, 2021.

\bibitem{xue2023promptcast}
H.~Xue and F.~D. Salim, ``{PromptCast: A New Prompt-based Learning Paradigm for
  Time Series Forecasting},'' {\em IEEE TKDE}, 2023.

\bibitem{gruver2023llmtime}
N.~Gruver, M.~Finzi, S.~Qiu, and A.~G. Wilson, ``{Large Language Models Are
  Zero Shot Time Series Forecasters},'' in {\em NeurIPS}, 2023.

\bibitem{zhou2023onefitsall}
T.~Zhou, P.~Niu, X.~Wang, L.~Sun, and R.~Jin, ``{One Fits All: Power General
  Time Series Analysis by Pretrained LM},'' in {\em NeurIPS}, 2023.

\bibitem{chang2023llm4ts}
C.~Chang, W.-C. Peng, and T.-F. Chen, ``{LLM4TS: Aligning Pre-Trained LLMs as
  Data-Efficient Time-Series Forecasters},'' {\em arXiv preprint
  arXiv:2308.08469}, 2023.

\bibitem{cao2023tempo}
D.~Cao, F.~Jia, S.~O. Arik, T.~Pfister, Y.~Zheng, W.~Ye, and Y.~Liu, ``{TEMPO:
  Prompt-based Generative Pre-trained Transformer for Time Series
  Forecasting},'' in {\em ICLR}, 2024.

\bibitem{sun2023test}
C.~Sun, Y.~Li, H.~Li, and S.~Hong, ``{TEST: Text Prototype Aligned Embedding to
  Activate LLM's Ability for Time Series},'' in {\em ICLR}, 2024.

\bibitem{jin2023time}
M.~Jin, S.~Wang, L.~Ma, Z.~Chu, J.~Y. Zhang, X.~Shi, P.-Y. Chen, Y.~Liang,
  Y.-F. Li, S.~Pan, {\em et~al.}, ``{Time-LLM: Time Series Forecasting by
  Reprogramming Large Language Models},'' in {\em ICLR}, 2024.

\bibitem{hu2021lora}
E.~J. Hu, P.~Wallis, Z.~Allen-Zhu, Y.~Li, S.~Wang, L.~Wang, {\em et~al.},
  ``{LoRA: Low-Rank Adaptation of Large Language Models},'' in {\em ICLR},
  2021.

\bibitem{shi2023language}
X.~Shi, S.~Xue, K.~Wang, F.~Zhou, J.~Y. Zhang, J.~Zhou, C.~Tan, and H.~Mei,
  ``{Language Models Can Improve Event Prediction by Few-Shot Abductive
  Reasoning},'' in {\em NeurIPS}, 2023.

\bibitem{xu2023language}
H.~Xu, Y.~Gao, Z.~Hui, J.~Li, and X.~Gao, ``{Language Knowledge-Assisted
  Representation Learning for Skeleton-Based Action Recognition},'' {\em arXiv
  preprint arXiv:2305.12398}, 2023.

\bibitem{rasul2023lagllama}
K.~Rasul, A.~Ashok, A.~R. Williams, A.~Khorasani, G.~Adamopoulos,
  R.~Bhagwatkar, {\em et~al.}, ``{Lag-Llama: Towards Foundation Models for Time
  Series Forecasting},'' 2023.

\bibitem{mai2023enhanced}
C.~Mai, Y.~Chang, C.~Chen, and Z.~Zheng, ``{Enhanced Scalable Graph Neural
  Network via Knowledge Distillation},'' {\em IEEE TNNLS}, 2023.

\bibitem{xue2022leveraging}
H.~Xue, B.~P. Voutharoja, and F.~D. Salim, ``{Leveraging Language Foundation
  Models for Human Mobility Forecasting},'' in {\em SIGSPATIAL}, 2022.

\bibitem{chen2023gatgpt}
Y.~Chen, X.~Wang, and G.~Xu, ``{GATGPT: A Pre-trained Large Language Model with
  Graph Attention Network for Spatiotemporal Imputation},'' {\em arXiv preprint
  arXiv:2311.14332}, 2023.

\bibitem{achiam2023gpt}
J.~Achiam, S.~Adler, S.~Agarwal, L.~Ahmad, I.~Akkaya, F.~L. Aleman, D.~Almeida,
  J.~Altenschmidt, S.~Altman, S.~Anadkat, {\em et~al.}, ``Gpt-4 technical
  report,'' {\em arXiv preprint arXiv:2303.08774}, 2023.

\bibitem{dubey2024llama}
A.~Dubey, A.~Jauhri, A.~Pandey, A.~Kadian, A.~Al-Dahle, A.~Letman, A.~Mathur,
  A.~Schelten, A.~Yang, A.~Fan, {\em et~al.}, ``The llama 3 herd of models,''
  {\em arXiv preprint arXiv:2407.21783}, 2024.

\bibitem{guo2025deepseek}
D.~Guo, D.~Yang, H.~Zhang, J.~Song, R.~Zhang, R.~Xu, Q.~Zhu, S.~Ma, P.~Wang,
  X.~Bi, {\em et~al.}, ``Deepseek-r1: Incentivizing reasoning capability in
  llms via reinforcement learning,'' {\em arXiv preprint arXiv:2501.12948},
  2025.

\bibitem{jiang2024mixtral}
A.~Q. Jiang, A.~Sablayrolles, A.~Roux, A.~Mensch, B.~Savary, C.~Bamford, D.~S.
  Chaplot, D.~d.~l. Casas, E.~B. Hanna, F.~Bressand, {\em et~al.}, ``Mixtral of
  experts,'' {\em arXiv preprint arXiv:2401.04088}, 2024.

\bibitem{anthropic2024introducing}
A.~Anthropic, ``Introducing the next generation of claude,'' 2024.

\bibitem{team2024gemini}
G.~Team, P.~Georgiev, V.~I. Lei, R.~Burnell, L.~Bai, A.~Gulati, G.~Tanzer,
  D.~Vincent, Z.~Pan, S.~Wang, {\em et~al.}, ``Gemini 1.5: Unlocking multimodal
  understanding across millions of tokens of context,'' {\em arXiv preprint
  arXiv:2403.05530}, 2024.

\bibitem{harris2020array}
C.~R. Harris, K.~J. Millman, S.~J. Van Der~Walt, R.~Gommers, P.~Virtanen,
  D.~Cournapeau, E.~Wieser, J.~Taylor, S.~Berg, N.~J. Smith, {\em et~al.},
  ``Array programming with numpy,'' {\em Nature}, vol.~585, no.~7825,
  pp.~357--362, 2020.

\bibitem{mckinney2010data}
W.~McKinney {\em et~al.}, ``Data structures for statistical computing in
  python.,'' {\em SciPy}, vol.~445, no.~1, pp.~51--56, 2010.

\bibitem{hunter2007matplotlib}
J.~D. Hunter, ``Matplotlib: A 2d graphics environment,'' {\em Computing in
  science \& engineering}, vol.~9, no.~03, pp.~90--95, 2007.

\bibitem{pedregosa2011scikit}
F.~Pedregosa, G.~Varoquaux, A.~Gramfort, V.~Michel, B.~Thirion, O.~Grisel,
  M.~Blondel, P.~Prettenhofer, R.~Weiss, V.~Dubourg, {\em et~al.},
  ``Scikit-learn: Machine learning in python,'' {\em the Journal of machine
  Learning research}, vol.~12, pp.~2825--2830, 2011.

\bibitem{met2010cartopy}
U.~Met~Office, ``Cartopy: A cartographic python library with a matplotlib
  interface,'' {\em Exeter, Devon}, 2010.

\bibitem{alemany2018diseno}
S.~Alemany~Ibor, {\em Dise{\~n}o e implementaci{\'o}n de un simulador basado en
  agentes estilo JGOMAS en Python}.
\newblock PhD thesis, Universitat Polit{\`e}cnica de Val{\`e}ncia, 2018.

\bibitem{mitchell2011pulp}
S.~Mitchell, M.~OSullivan, and I.~Dunning, ``Pulp: a linear programming toolkit
  for python,'' {\em The University of Auckland, Auckland, New Zealand},
  vol.~65, p.~25, 2011.

\bibitem{wu2023autogen}
Q.~Wu, G.~Bansal, J.~Zhang, Y.~Wu, S.~Zhang, E.~Zhu, B.~Li, L.~Jiang, X.~Zhang,
  and C.~Wang, ``Autogen: Enabling next-gen llm applications via multi-agent
  conversation framework,'' {\em arXiv preprint arXiv:2308.08155}, 2023.

\bibitem{forootani2024climate}
A.~Forootani, D.~E. Aliabadi, and D.~Thraen, ``Climate aware deep neural
  networks (cadnn) for wind power simulation,'' {\em arXiv preprint
  arXiv:2412.12160}, 2024.

\bibitem{zhangoffline}
S.~Zhang, J.~Zhang, J.~Liu, L.~Song, C.~Wang, R.~Krishna, and Q.~Wu, ``Offline
  training of language model agents with functions as learnable weights,'' in
  {\em Forty-first International Conference on Machine Learning}.

\bibitem{rawte2023survey}
V.~Rawte, A.~Sheth, and A.~Das, ``A survey of hallucination in large foundation
  models,'' {\em arXiv}, 2023.

\bibitem{he_rethinking_2022}
H.~He, H.~Zhang, and D.~Roth, ``Rethinking with retrieval: Faithful large
  language model inference,'' {\em arXiv}, 2022.

\bibitem{zhou2023solving}
A.~Zhou, K.~Wang, Z.~Lu, W.~Shi, S.~Luo, Z.~Qin, S.~Lu, A.~Jia, L.~Song,
  M.~Zhan, {\em et~al.}, ``Solving challenging math word problems using gpt-4
  code interpreter with code-based self-verification,'' {\em arXiv}, 2023.

\bibitem{li2023making}
Y.~Li, Z.~Lin, S.~Zhang, Q.~Fu, B.~Chen, J.-G. Lou, and W.~Chen, ``Making
  language models better reasoners with step-aware verifier,'' in {\em ACL},
  pp.~5315--5333, 2023.

\bibitem{zhao2023verify}
R.~Zhao, X.~Li, S.~Joty, C.~Qin, and L.~Bing, ``Verify-and-edit: A
  knowledge-enhanced chain-of-thought framework,'' {\em arXiv}, 2023.

\bibitem{shridhar2023screws}
K.~Shridhar, H.~Jhamtani, H.~Fang, B.~Van~Durme, J.~Eisner, and P.~Xia,
  ``Screws: A modular framework for reasoning with revisions,'' {\em arXiv},
  2023.

\bibitem{gelernter1960empirical}
H.~Gelernter, J.~R. Hansen, and D.~W. Loveland, ``Empirical explorations of the
  geometry theorem machine,'' in {\em western joint IRE-AIEE-ACM computer
  conference}, pp.~143--149, 1960.

\bibitem{lu2021iconqa}
P.~Lu, L.~Qiu, J.~Chen, T.~Xia, Y.~Zhao, W.~Zhang, Z.~Yu, X.~Liang, and S.-C.
  Zhu, ``Iconqa: A new benchmark for abstract diagram understanding and visual
  language reasoning,'' in {\em NeurIPS Datasets and Benchmarks Track}, 2021.

\bibitem{antol2015vqa}
S.~Antol, A.~Agrawal, J.~Lu, M.~Mitchell, D.~Batra, C.~L. Zitnick, and
  D.~Parikh, ``Vqa: Visual question answering,'' in {\em ICCV}, pp.~2425--2433,
  2015.

\bibitem{hossain2019comprehensive}
M.~Z. Hossain, F.~Sohel, M.~F. Shiratuddin, and H.~Laga, ``A comprehensive
  survey of deep learning for image captioning,'' {\em ACM Computing Surveys
  (CsUR)}, vol.~51, no.~6, pp.~1--36, 2019.

\bibitem{gawlikowski2023survey}
J.~Gawlikowski, C.~R.~N. Tassi, M.~Ali, J.~Lee, M.~Humt, J.~Feng, A.~Kruspe,
  R.~Triebel, P.~Jung, R.~Roscher, {\em et~al.}, ``A survey of uncertainty in
  deep neural networks,'' {\em Artificial Intelligence Review}, vol.~56,
  no.~Suppl 1, pp.~1513--1589, 2023.

\bibitem{duan2023shifting}
J.~Duan, H.~Cheng, S.~Wang, C.~Wang, A.~Zavalny, R.~Xu, B.~Kailkhura, and
  K.~Xu, ``Shifting attention to relevance: Towards the uncertainty estimation
  of large language models,'' {\em arXiv}, 2023.

\bibitem{DBLP:conf/icml/ZhouJWCS23}
W.~Zhou, Y.~E. Jiang, E.~Wilcox, R.~Cotterell, and M.~Sachan, ``Controlled text
  generation with natural language instructions,'' in {\em ICML}, vol.~202,
  pp.~42602--42613, 2023.

\bibitem{wang2021gpt}
B.~Wang and A.~Komatsuzaki, ``Gpt-j-6b: A 6 billion parameter autoregressive
  language model,'' 2021.

\bibitem{chung2022scaling}
H.~W. Chung, L.~Hou, S.~Longpre, B.~Zoph, Y.~Tay, W.~Fedus, Y.~Li, X.~Wang,
  M.~Dehghani, S.~Brahma, {\em et~al.}, ``Scaling instruction-finetuned
  language models,'' {\em arXiv}, 2022.

\bibitem{thoppilan2022lamda}
R.~Thoppilan, D.~De~Freitas, J.~Hall, N.~Shazeer, A.~Kulshreshtha, H.-T. Cheng,
  A.~Jin, T.~Bos, L.~Baker, Y.~Du, {\em et~al.}, ``Lamda: Language models for
  dialog applications,'' {\em arXiv}, 2022.

\bibitem{davies2021advancing}
A.~Davies, P.~Veli{\v{c}}kovi{\'c}, L.~Buesing, S.~Blackwell, D.~Zheng,
  N.~Toma{\v{s}}ev, R.~Tanburn, P.~Battaglia, C.~Blundell, A.~Juh{\'a}sz, {\em
  et~al.}, ``Advancing mathematics by guiding human intuition with ai,'' {\em
  Nature}, vol.~600, no.~7887, pp.~70--74, 2021.

\bibitem{bernardino2023mathematical}
A.~N. Bernardino Romera-Paredes, Mohammadamin~Barekatain {\em et~al.},
  ``Mathematical discoveries from program search with large language models,''
  {\em Nature}, 2023.

\bibitem{shridhar2022automatic}
K.~Shridhar, J.~Macina, M.~El-Assady, T.~Sinha, M.~Kapur, and M.~Sachan,
  ``Automatic generation of socratic subquestions for teaching math word
  problems,'' {\em arXiv}, 2022.

\bibitem{ding_enhancing_2023}
N.~Ding, Y.~Chen, B.~Xu, Y.~Qin, S.~Hu, Z.~Liu, M.~Sun, and B.~Zhou,
  ``Enhancing chat language models by scaling high-quality instructional
  conversations,'' in {\em EMNLP}, pp.~3029--3051, 2023.

\end{thebibliography}
